\documentclass[10pt,twocolumn,letterpaper]{article}

\usepackage{cvpr}              %

\usepackage[dvipsnames]{xcolor}

\usepackage{times}
\usepackage{epsfig}
\usepackage{graphicx}
\usepackage{amsmath}
\usepackage{amssymb}

\usepackage{lipsum}
\usepackage{enumitem}
\usepackage{mathtools}
\usepackage[dvipsnames]{xcolor}
\usepackage{tabularray}
\UseTblrLibrary{booktabs}
\usepackage[export]{adjustbox}
\usepackage{capt-of}
\usepackage{xparse}
\usepackage{cancel}
\usepackage{makecell}
\usepackage{rotating}
\usepackage{subcaption}
\usepackage{balance}
\usepackage{float}

\newcommand{\PAR}[1]{\vskip4pt \noindent{\bf #1~}}

\DeclareMathAlphabet\mathbfcal{OMS}{cmsy}{b}{n}

\DeclarePairedDelimiter{\Edel}{\{}{\}}
\newcommand{\E}{\operatorname{\mathbb{E}}\Edel}
\DeclarePairedDelimiter{\norm}{\lVert}{\rVert}

\DeclareDocumentCommand{\gradp}{e{_^}}{%
\nabla
\IfValueT{#1}{_{\!#1}}%
\IfValueT{#2}{^{#2}}%
}

\newcommand{\twodots}{\mathinner{\ldotp\ldotp}}

\newcommand{\evalat}[2][\mathbf{0}]{\left. #2 \right|_{#1}}

\newcommand{\bI}{\mathbf{I}}
\newcommand{\bS}{\mathbf{S}}
\newcommand{\bx}{\mathbf{x}}
\newcommand{\bp}{\mathbf{p}}
\newcommand{\bt}{\mathbf{t}}
\newcommand{\by}{\mathbf{y}}
\newcommand{\bd}{\mathbf{d}}
\newcommand{\bs}{\mathbf{s}}

\newcommand{\bC}{\mathbf{C}}

\newcommand{\bW}{\mathbf{W}}

\newcommand{\bzero}{\mathbf{0}}
\newcommand{\dbx}{\delta\mathbf{x}}

\newcommand{\eps}{\varepsilon}
\newcommand{\pt}{\partial}

\newcommand{\calW}{\mathcal{W}}
\newcommand{\bcalI}{\mathbfcal{I}}

\newcommand{\bxi}{\boldsymbol{\xi}}
\newcommand{\bSigma}{\boldsymbol{\Sigma}}

\newcommand{\dby}{\Delta{\by}}
\newcommand{\bR}{\mathbf{R}}
\newcommand{\bH}{\mathbf{H}}
\newcommand{\bJ}{\mathbf{J}}
\newcommand{\be}{\mathbf{e}}
\newcommand{\bK}{\mathbf{K}}
\newcommand{\bM}{\mathbf{M}}

\newcommand{\br}{\mathbf{r}}
\newcommand{\bc}{\mathbf{c}}

\newcommand{\bxhat}{\hat{\mathbf{x}}}

\newcommand{\bzeta}{\boldsymbol{\zeta}}

\newcommand{\bLambda}{\boldsymbol{\Lambda}} 
\newcommand{\bnu}{\boldsymbol{\nu}}

\newcommand{\bw}{\mathbf{w}}

\newcommand{\bb}[1]{\mathbb{#1}}

\newcommand{\rand}[1]{\underbar{$#1$}}
\newcommand{\ceqq}{\coloneqq}
\newcommand{\homo}[1]{\tilde{#1}}
\newcommand{\sel}[1]{^{(#1)}}

\newcommand{\homobx}{\homo{\bx}}

\newcommand{\qualitativeComparison}[1][htp!]{
\begin{figure*}[#1]
    \centering
    \begin{tblr}{
            width=1.0\linewidth,
            vspan=even,
            colspec={*{4}c},
            rowsep=2pt,
            column{2}={colsep=0pt},
            hline{2} = {1pt,solid},
            hline{5,8,11} = {1,4}{1pt,dashed,gray3},
            cell{2,5,8,11}{1-3} = {r=3}{c},
            cell{2,5,8,11}{1} = {r=3}{c, cmd=\rotatebox{90}},
        }
        & Point-wise (Sec. \ref{subsec:pointwise}) & 
        Structure Tensor (Sec. \ref{subsec:tensor})
        & Zoom\\
        Key.Net \cite{barroso2022key} &
        \includegraphics[width=6.1cm,  valign=m]{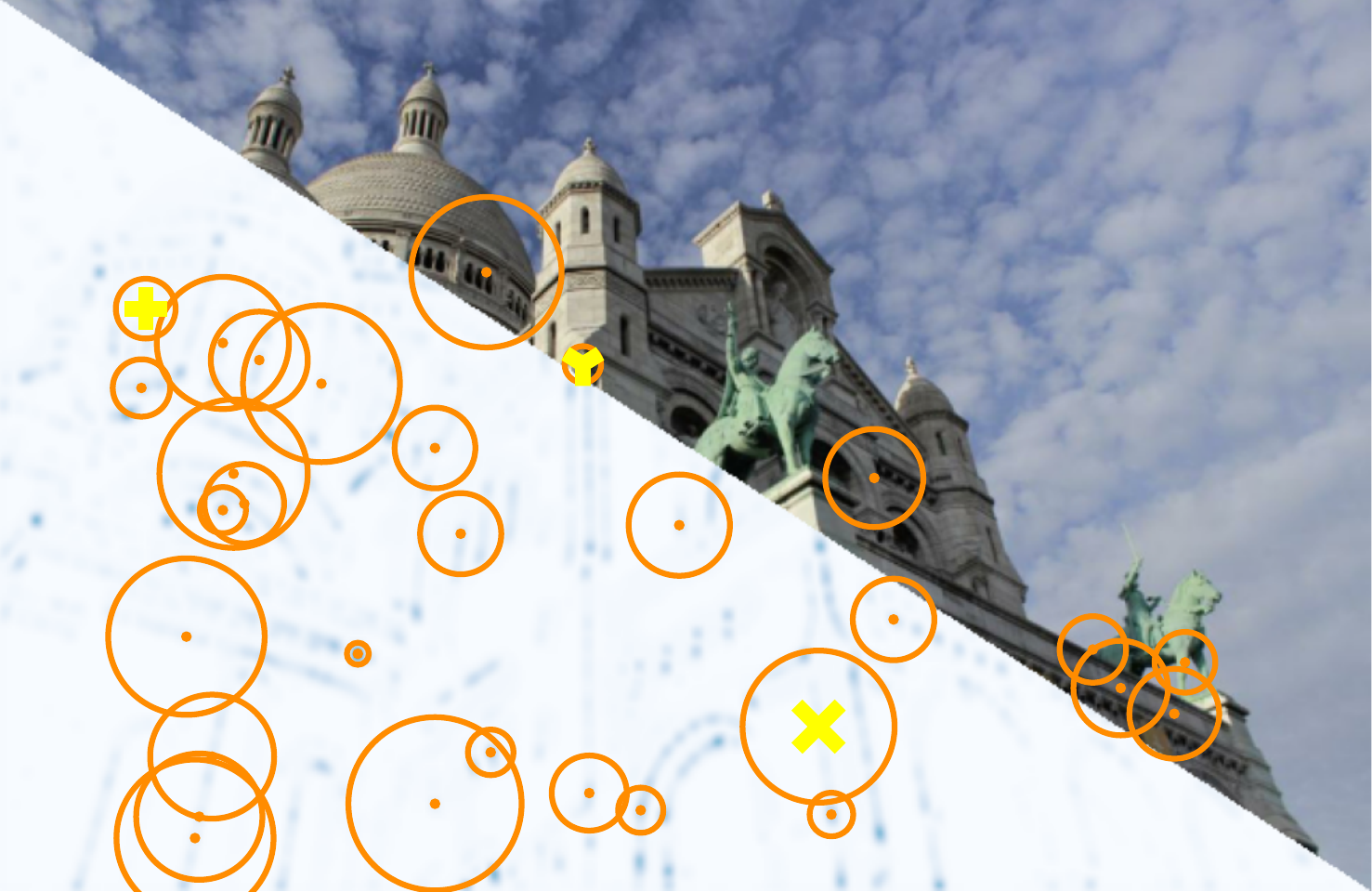} & 
        \includegraphics[width=6.1cm,  valign=m]{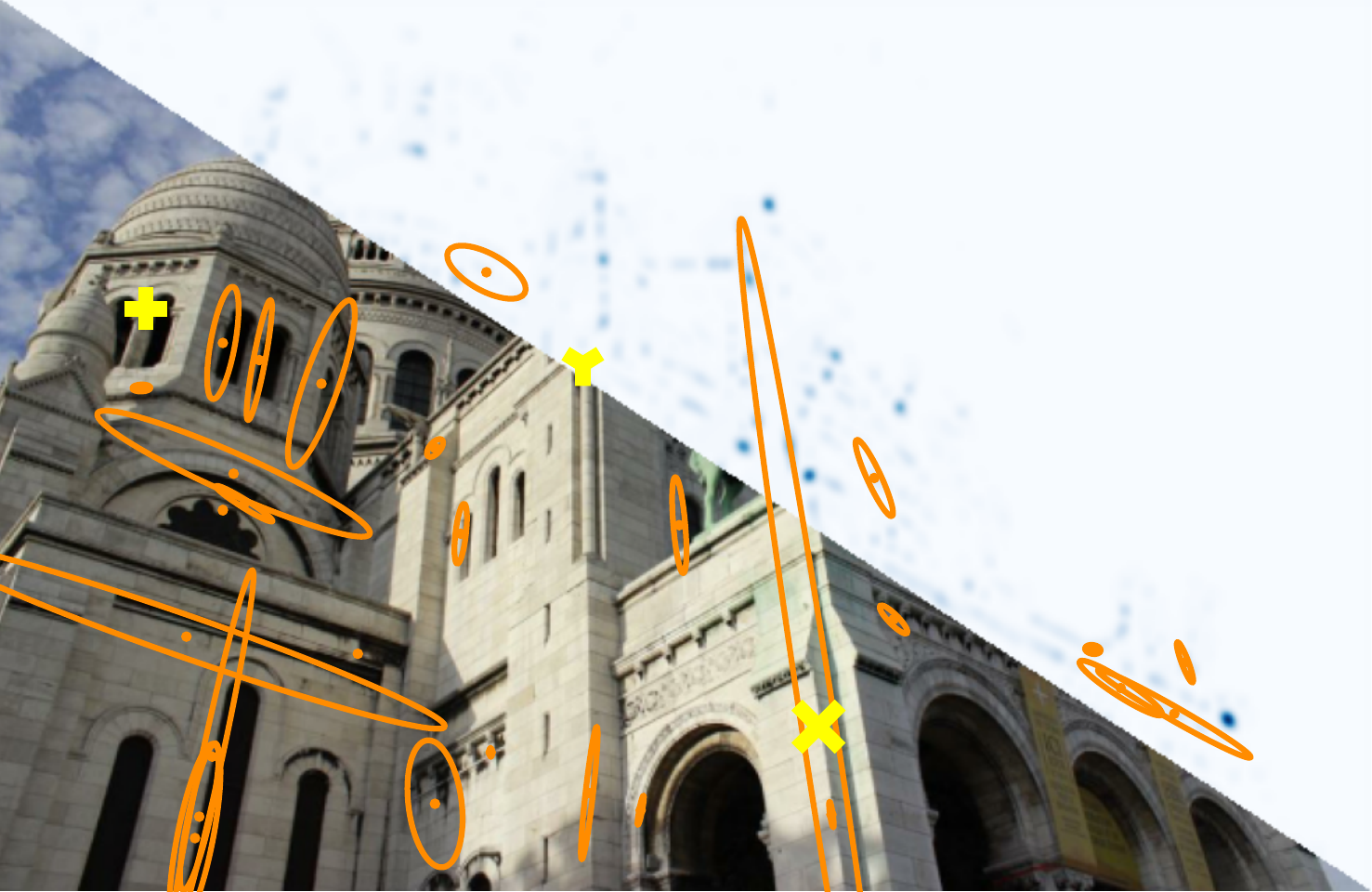} &
        \includegraphics[height=1.5cm, valign=m]{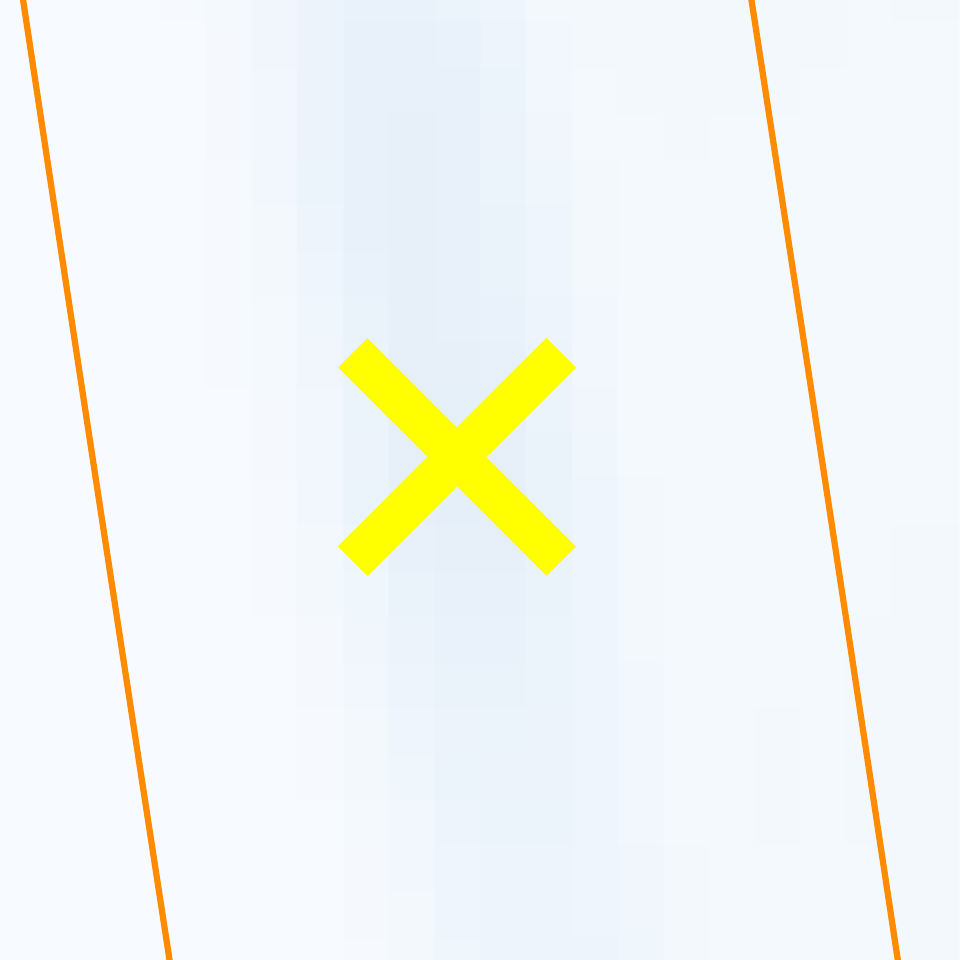}\\
        - & - & - &
        \includegraphics[height=1.5cm, valign=m]{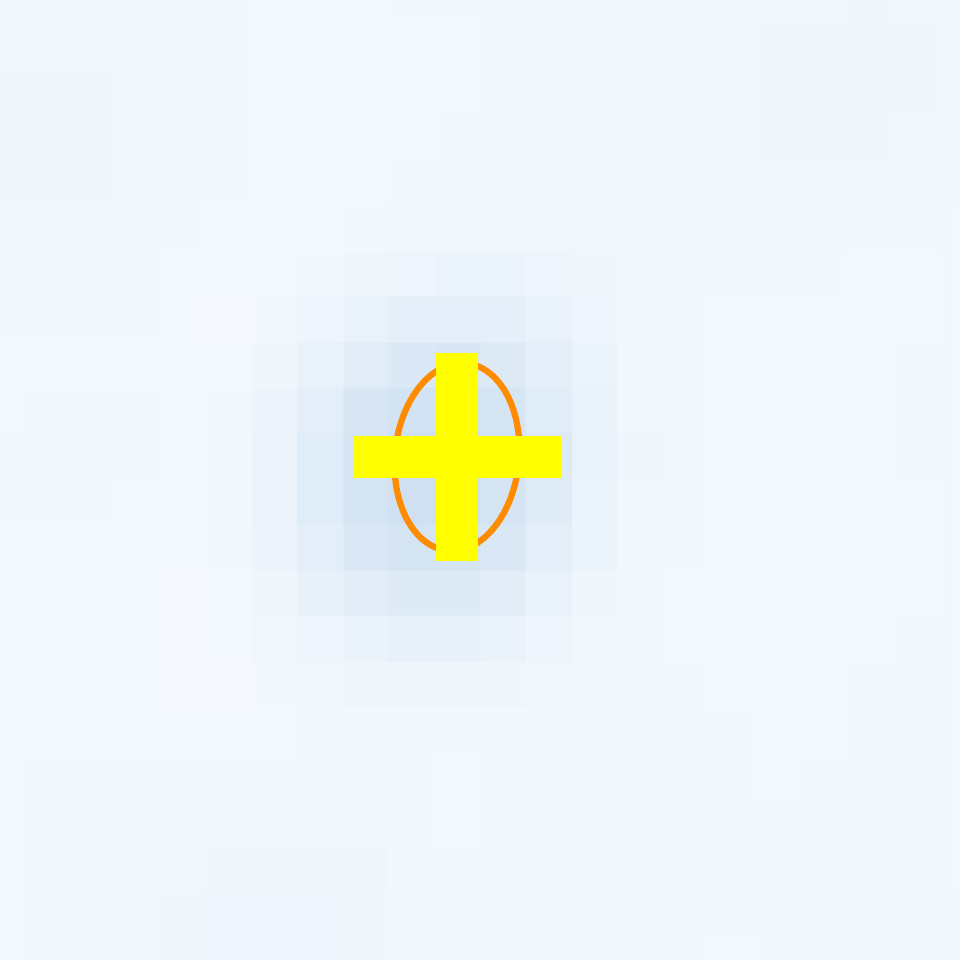}\\
        - & - & - &
        \includegraphics[height=1.5cm, valign=m]{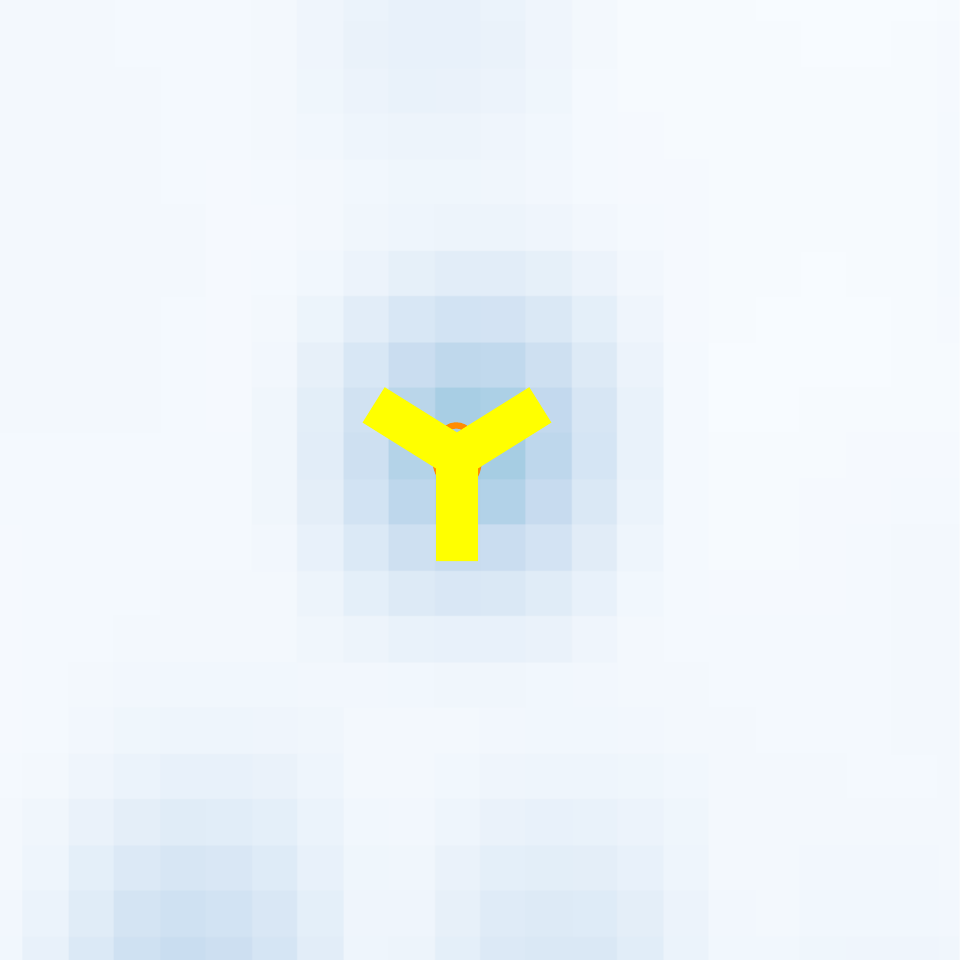}\\
        Superpoint \cite{detone2018superpoint} &
        \includegraphics[width=6.1cm,  valign=m]{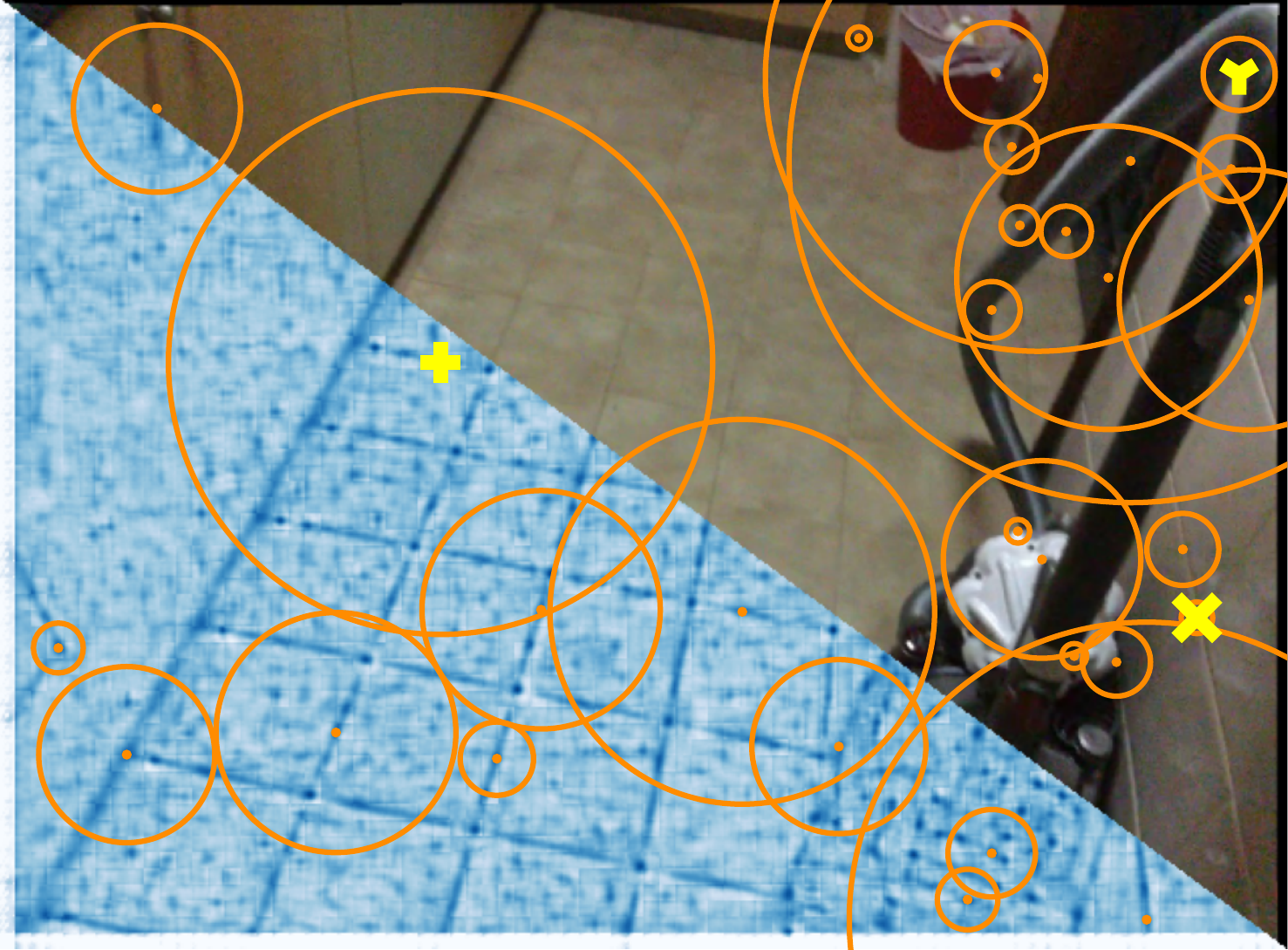} & 
        \includegraphics[width=6.1cm,  valign=m]{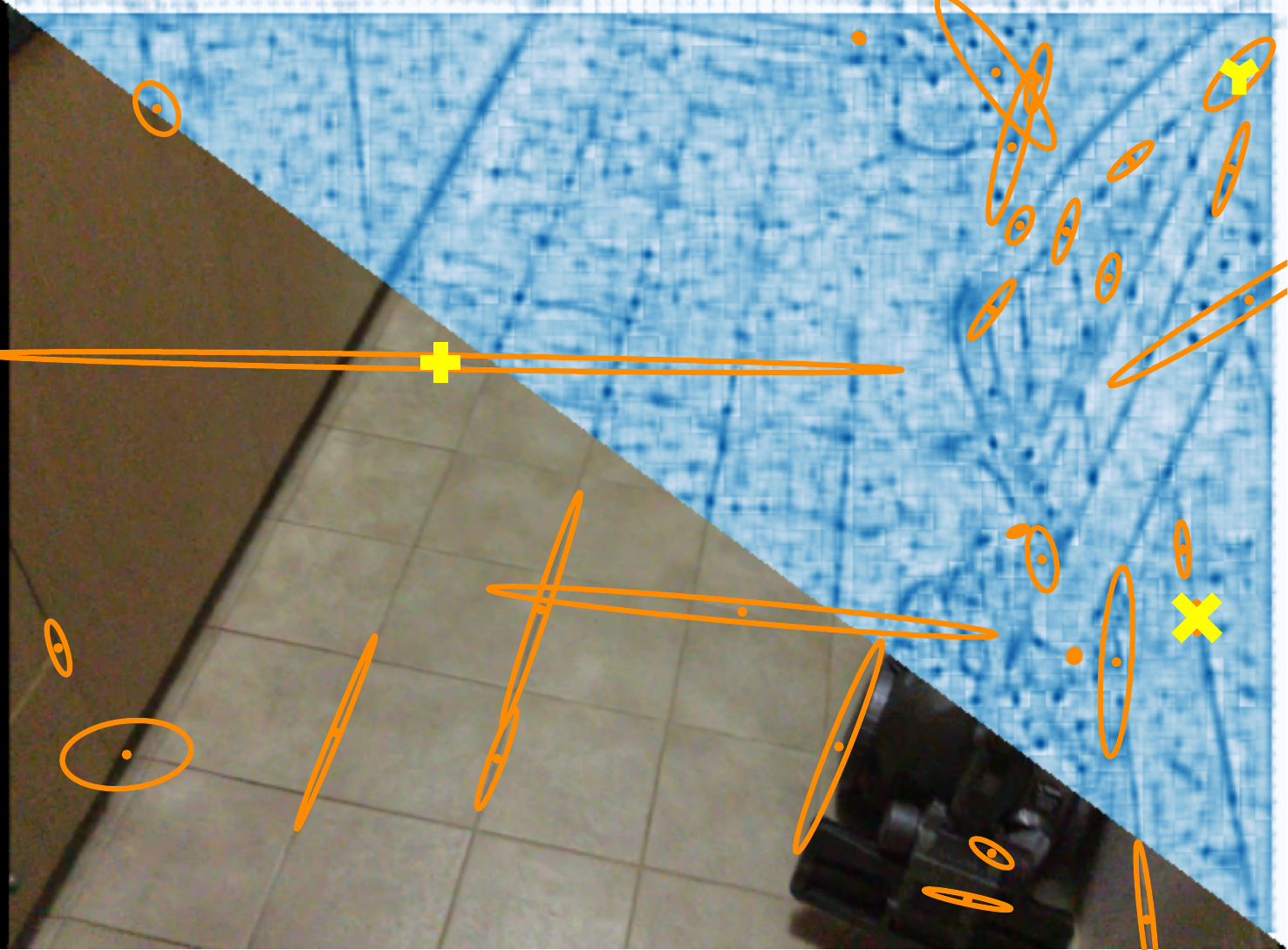} &
        \includegraphics[height=1.5cm, valign=m]{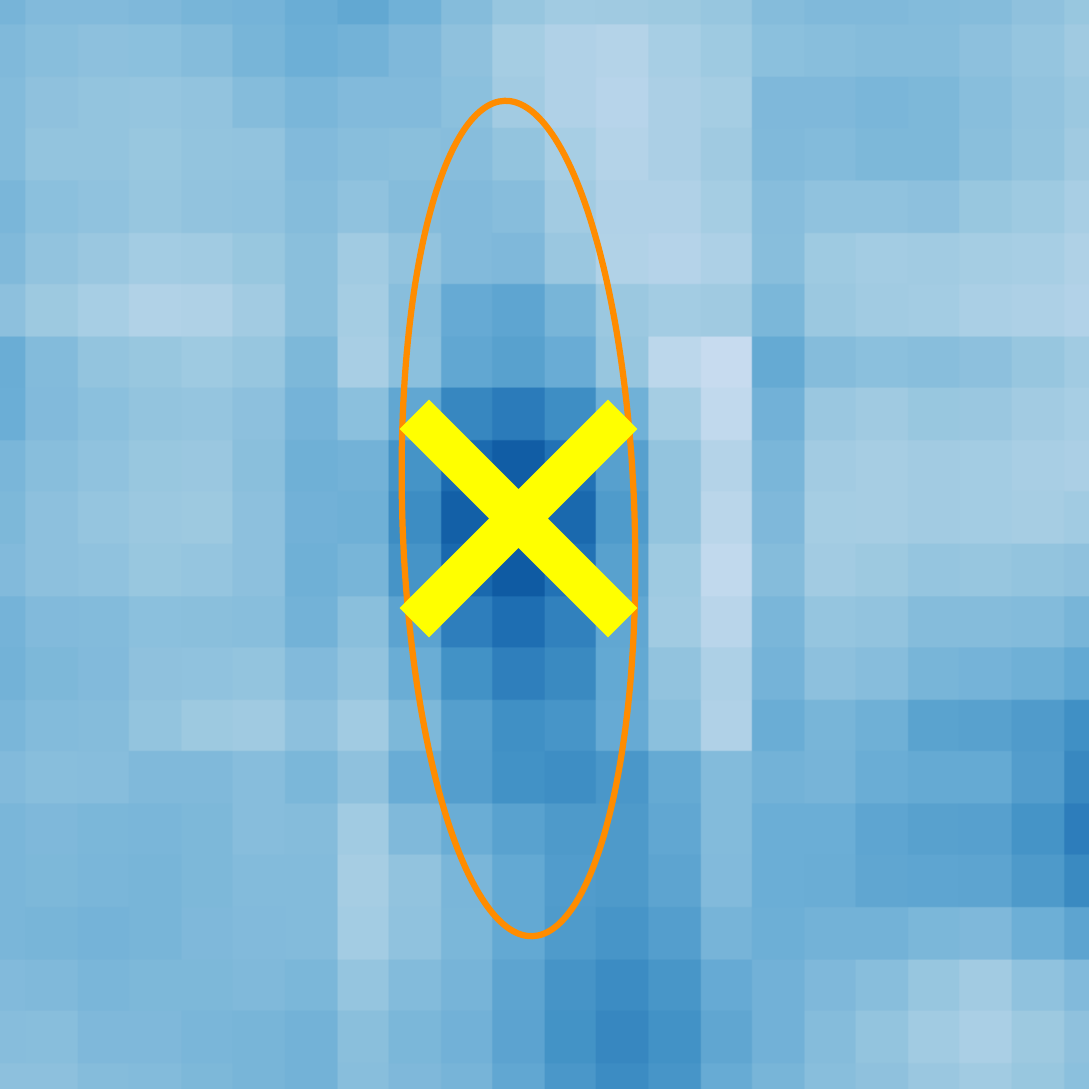}\\
        - & - & - &
        \includegraphics[height=1.5cm, valign=m]{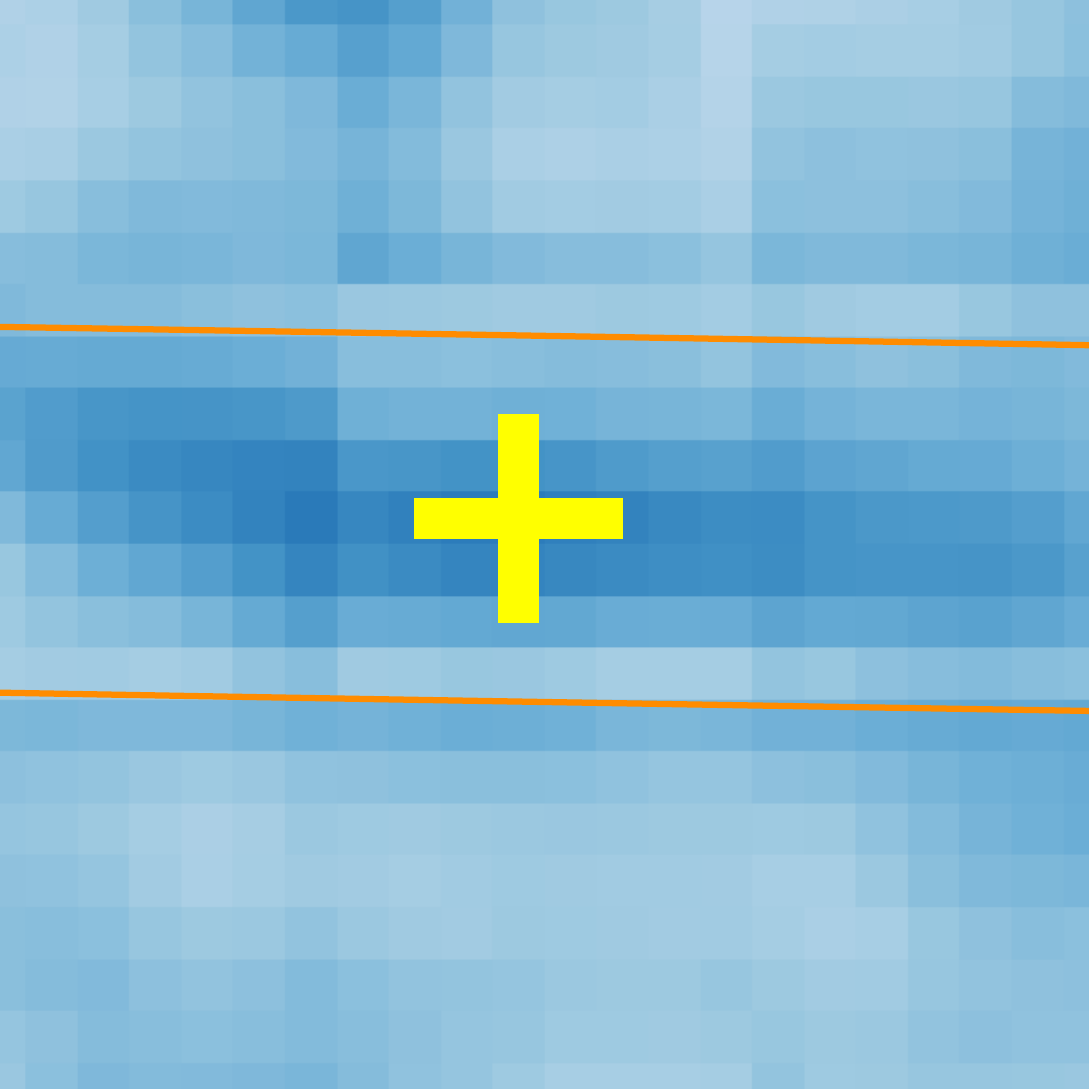}\\
        - & - & - &
        \includegraphics[height=1.5cm, valign=m]{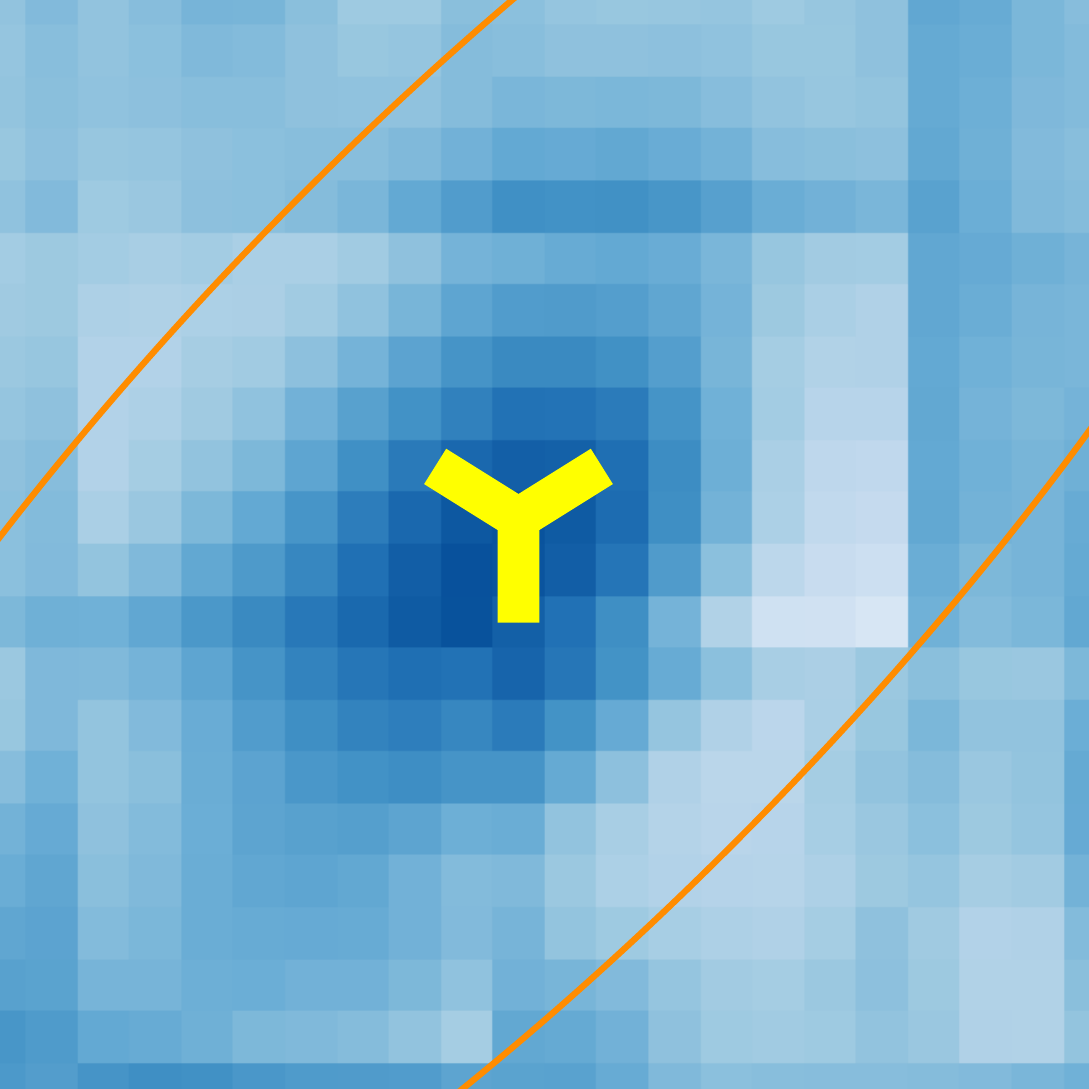}\\
        D2Net \cite{dusmanu2019d2} &
        \includegraphics[width=6.1cm,  valign=m]{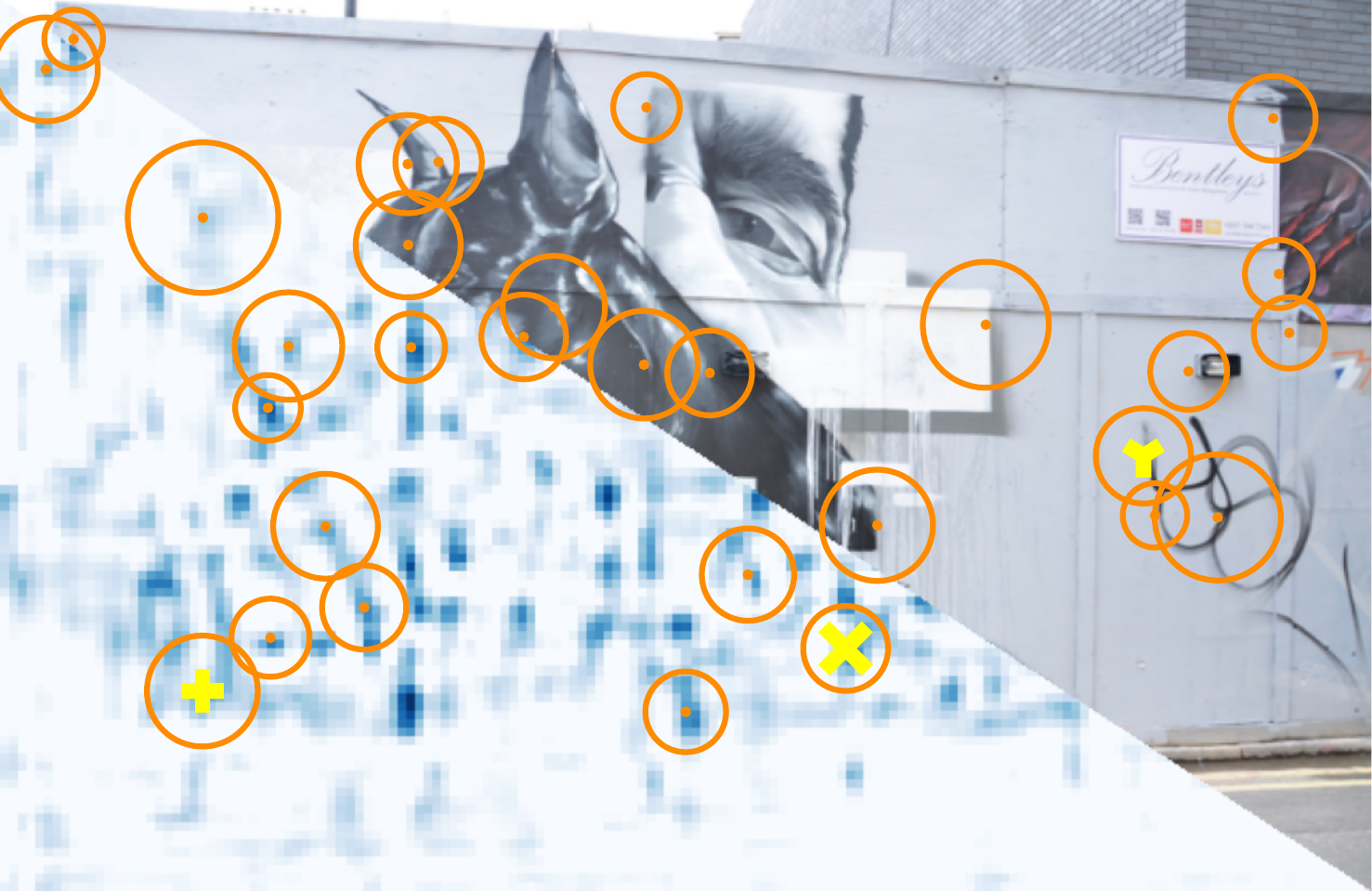} & 
        \includegraphics[width=6.1cm,  valign=m]{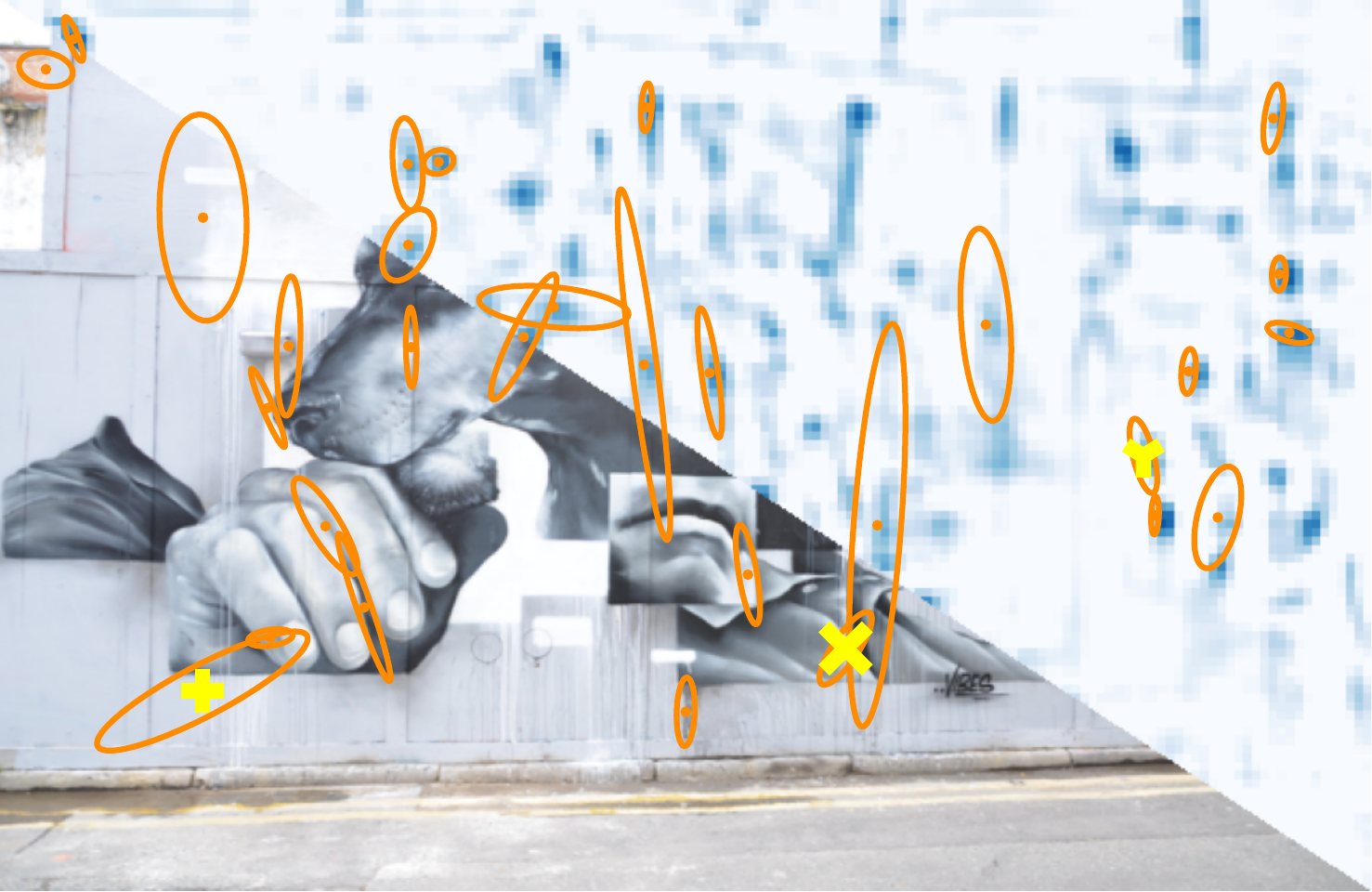} &
        \includegraphics[height=1.5cm, valign=m]{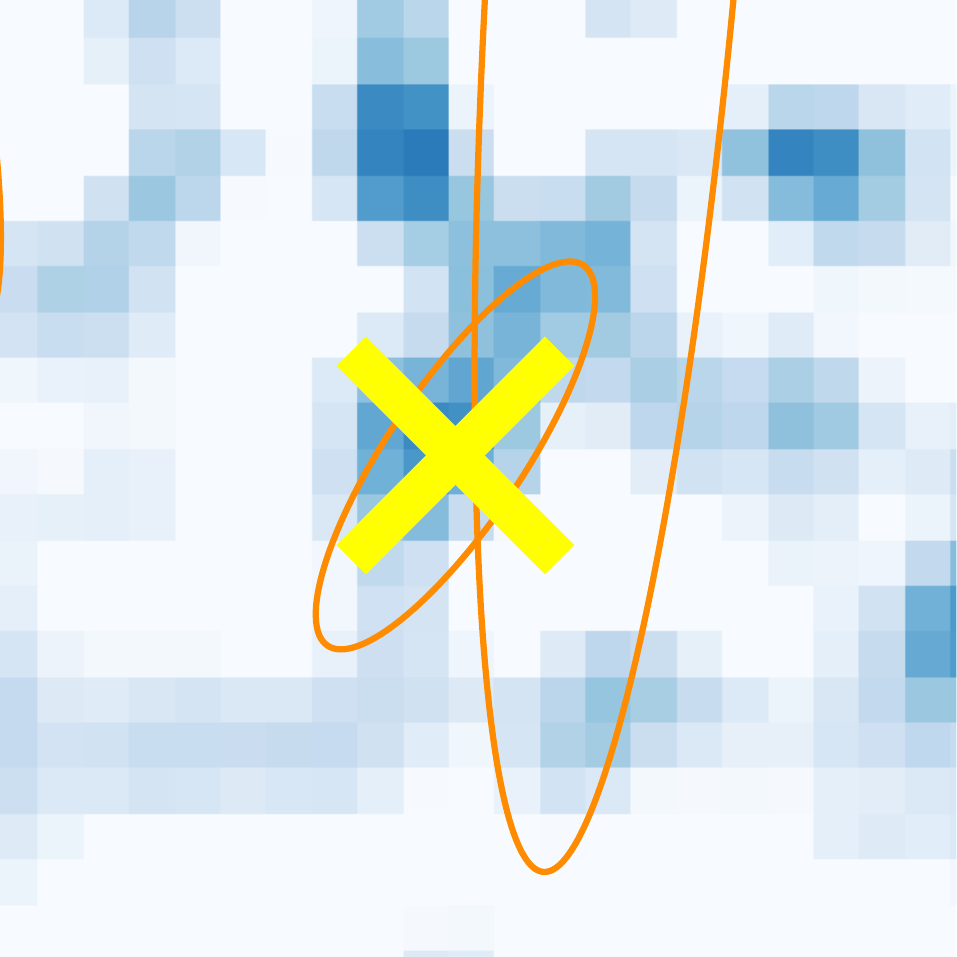}\\
        - & - & - &
        \includegraphics[height=1.5cm, valign=m]{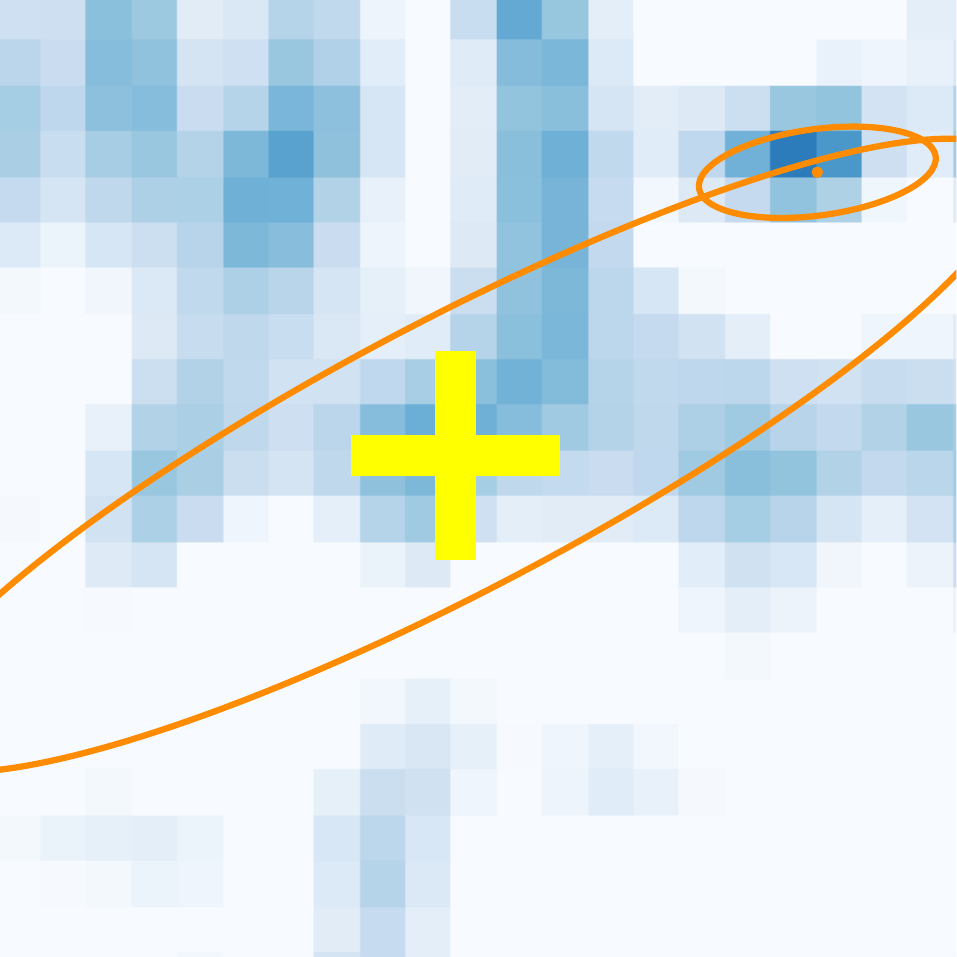}\\
        - & - & - &
        \includegraphics[height=1.5cm, valign=m]{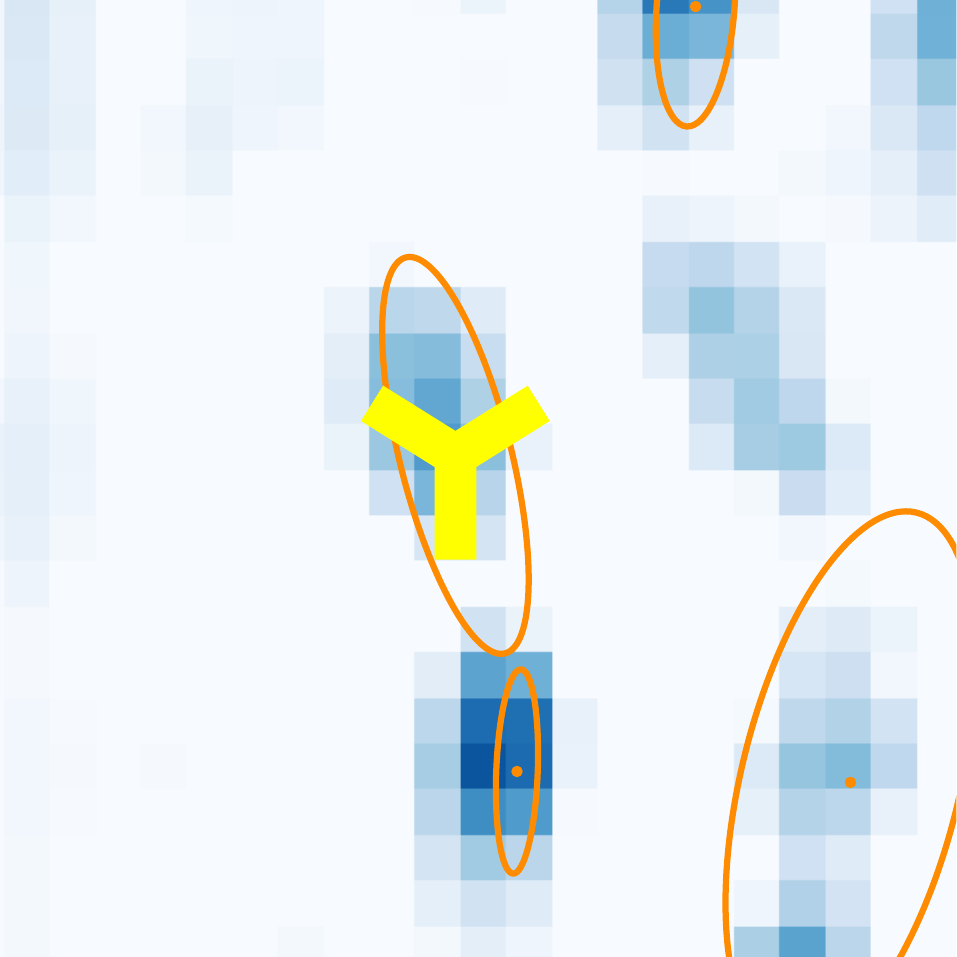}\\
        R2D2 \cite{revaud2019r2d2} &
        \includegraphics[width=6.1cm,  valign=m]{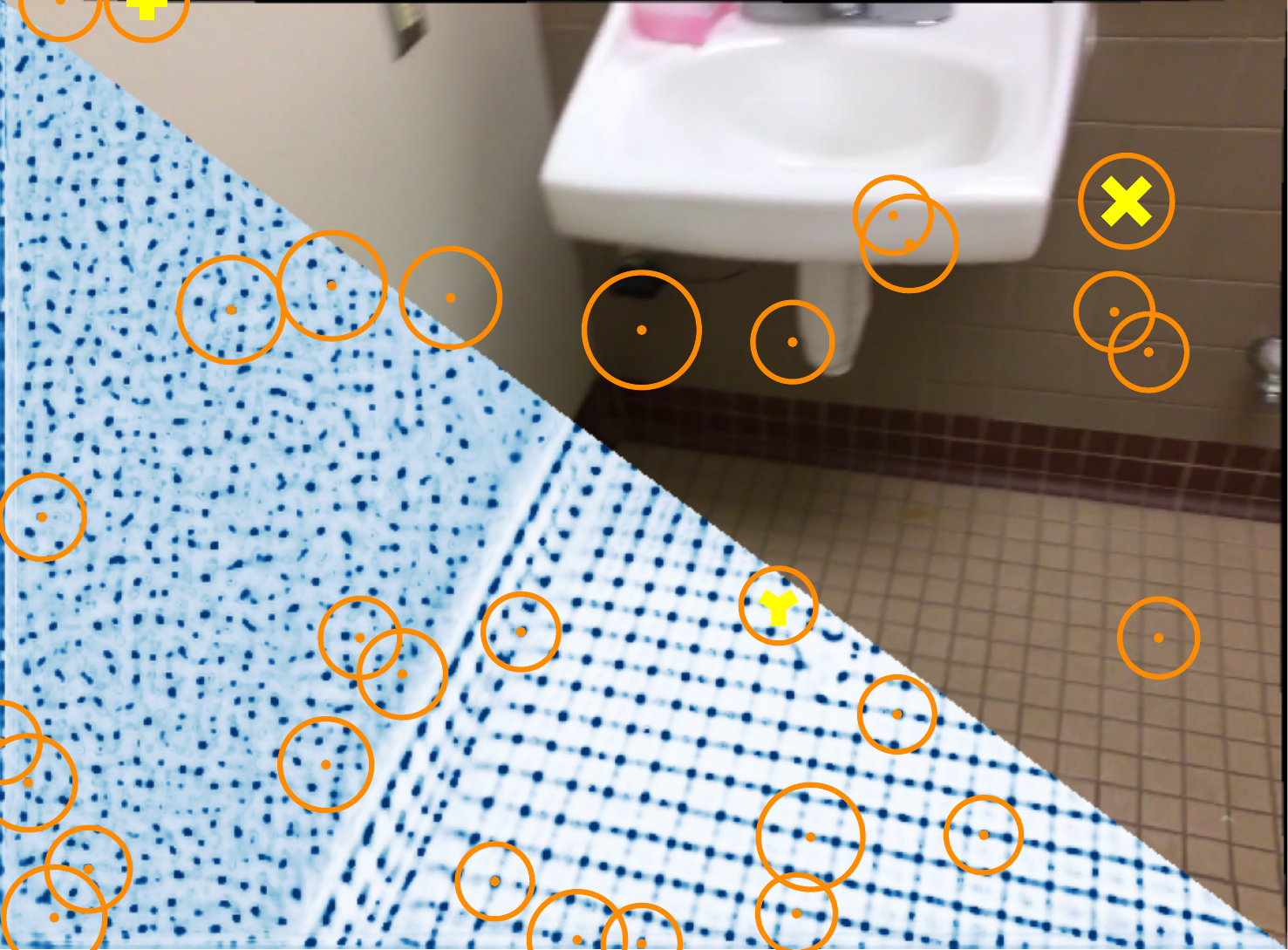} & 
        \includegraphics[width=6.1cm,  valign=m]{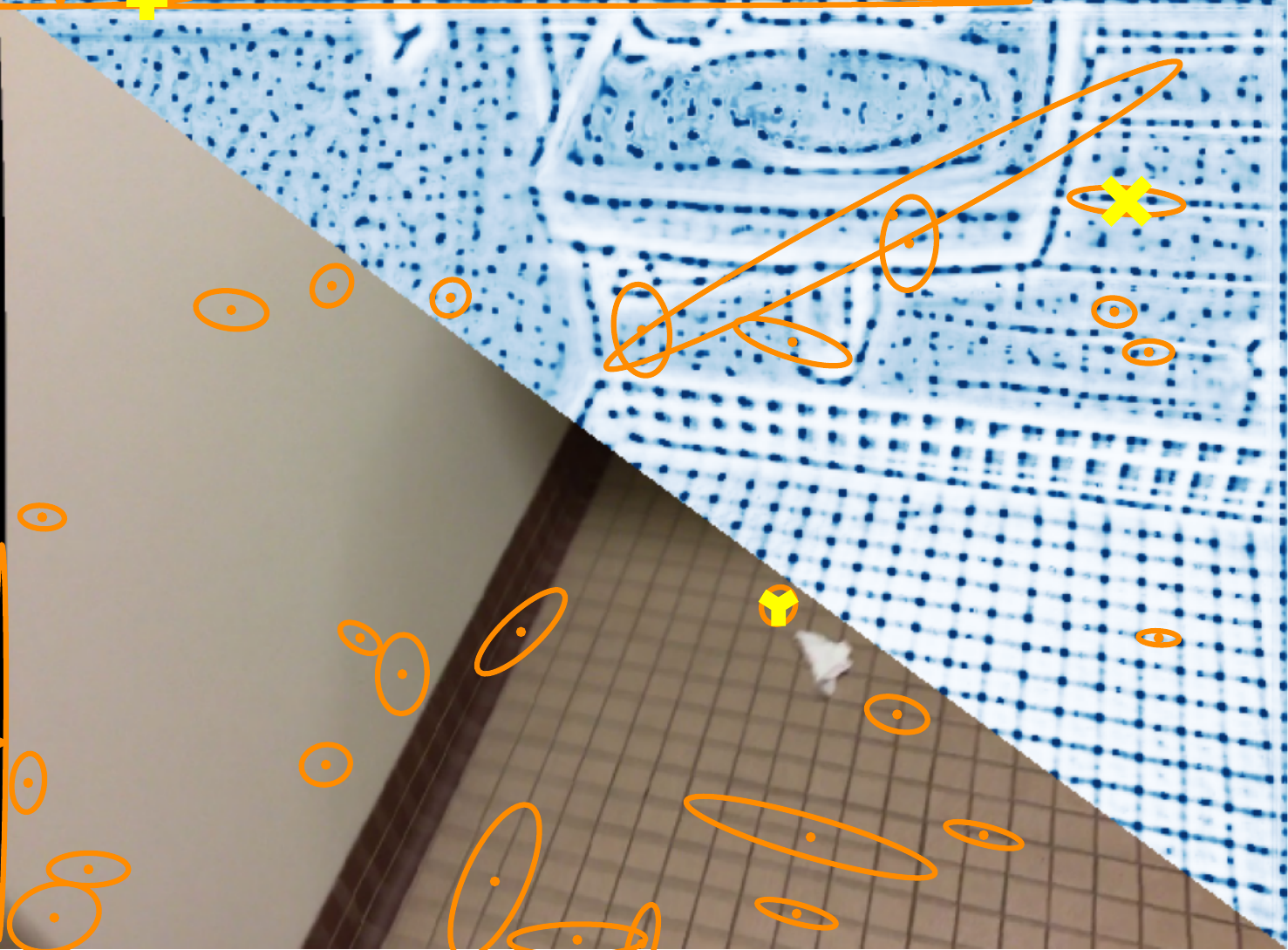} &
        \includegraphics[height=1.5cm, valign=m]{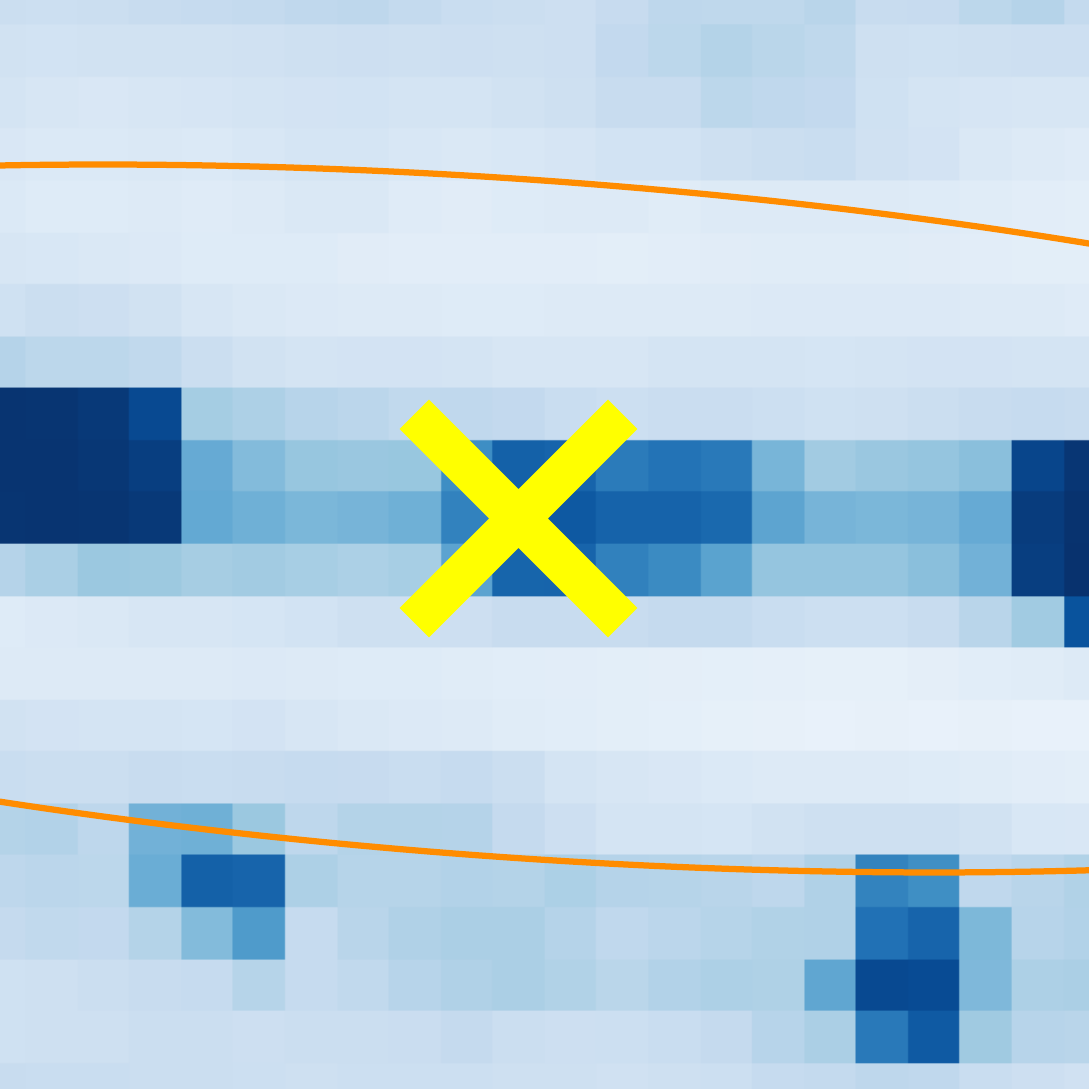}\\
        - & - & - &
        \includegraphics[height=1.5cm, valign=m]{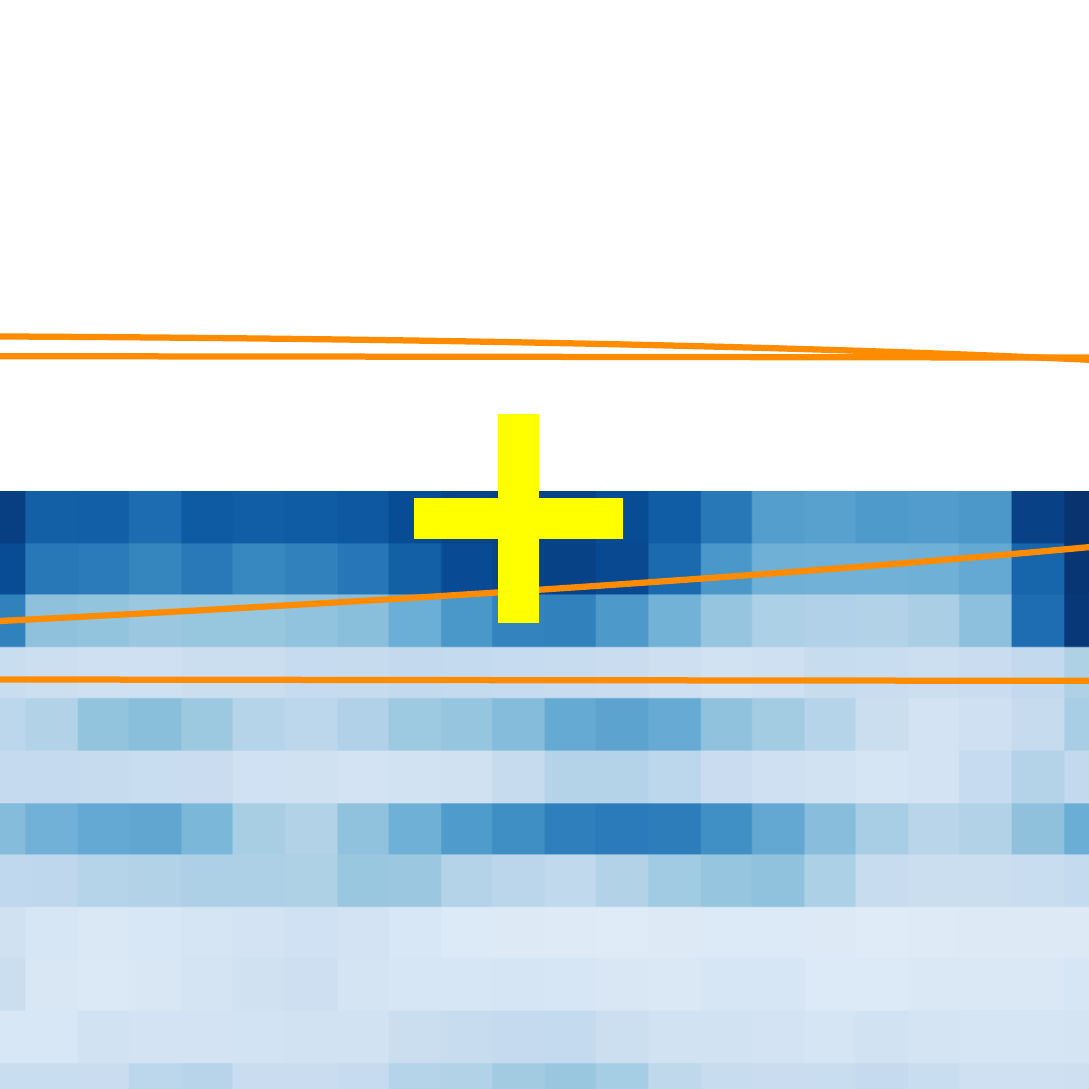}\\
        - & - & - &
        \includegraphics[height=1.5cm, valign=m]{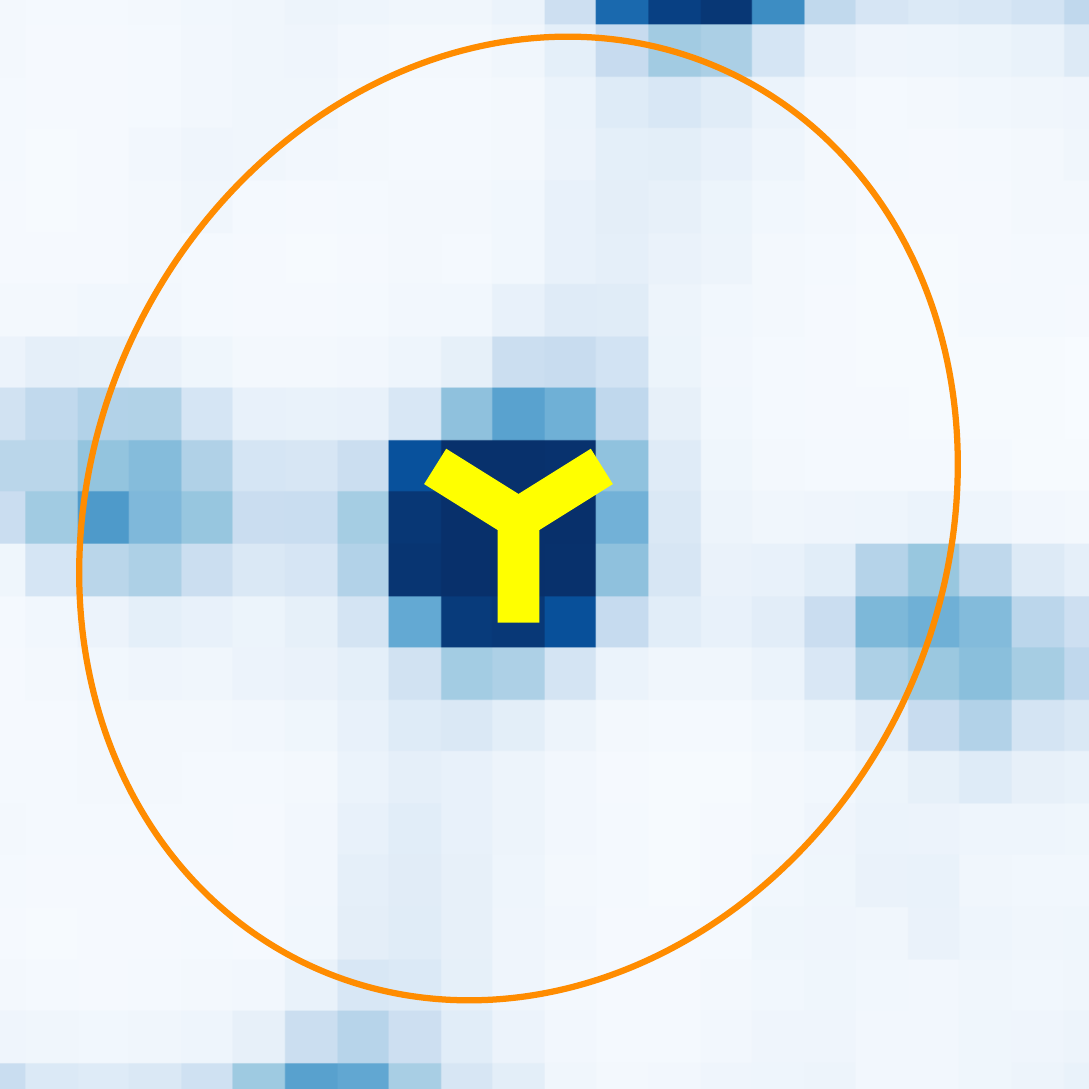}\\
    \end{tblr}
    \caption{\textbf{Qualitative comparison}. Our estimated covariance matrices for each detected local feature are illustrated with uncertainty ellipses for four state-of-the-art pretrained detectors. Our pointwise covariance estimator yields isotropic uncertainty predictions (equal in all directions). Our proposed local structure tensor estimates the directionality of the spatial uncertainty, resulting in anisotropic predictions. For instance, Superpoint (which is trained to detect corners) has high uncertainty patterns corresponding to lines. For visualization purposes, a uniform sample of local feature detections is shown.}
    \label{fig:comparison}
\end{figure*}
}
\newcommand{\tumkittiAll}[1][htb]{%
\begin{table*}[#1]
\scriptsize
    \newcommand{\er}{e$_\text{rot}$}
    \newcommand{\et}{e$_\text{t}$}
    
    \centering
    \begin{tblr}{
            colspec={ll*{10}{r}},
            hline{1,3}={3-Z}{solid},
            vline{3},
            vline{5}={gray!70, dashed},
            rowsep=0pt,
            hline{19,Z},
            hline{7,11,15,23,27,31}={gray!70, dashed},
            cell{1}{odd}={c=2}{c},
            row{2}={c},
            cell{3,7,11,15,19,23,27,31}{1}={r=4}{c, cmd=\rotatebox{90}},
            hline{3}={2}{-}{solid},
            hline{19}={2}{-}{solid},
            vline{1}={3-Z}{solid},
            vline{Z},
        }
       &          & w/o unc. & 1-3 & 2D-full (ours) & 1-5 & 2D-iso (ours) & 1-7 & 2D-3D-full (ours) & 1-9 & 2D-3D-iso (ours) \\
       &          & \er & \et & \er & \et & \er & \et & \er & \et & \er & \et \\
       00         & D2Net   & 8.47 & $>$1e3 & 1.39 & 121.88 & \textbf{1.35} & \textbf{118.64} & 1.53 & 139.06 & 1.49 & 134.75 \\
       --         & Key.Net & 31.44 & $>$1e3 & 0.96 & 79.81 & \textbf{0.78} & \textbf{66.85} & 0.94 & 89.71 & 0.91 & 87.13 \\
       --         & R2D2    & 16.72 & $>$1e3 & 0.88 & 76.72 & \textbf{0.88} & \textbf{76.17} & 1.33 & 118.36 & 2.42 & 175.33 \\
       --         & SP      & 10.44 & $>$1e3 & 0.80 & 72.00 & \textbf{0.80} & \textbf{71.21} & 0.94 & 88.46 & 0.91 & 83.91 \\
       01         & D2Net   & 114.01 & $>$1e3 & 1.16 & 196.71 & \textbf{1.07} & \textbf{187.38} & 4.75 & 844.34 & 4.43 & 627.20  \\
       --         & Key.Net & 314.77 & $>$1e3 & 1.32 & 187.74 & \textbf{0.47} & \textbf{71.26} & 6.14 & $>$1e3 & 3.75 & 623.79  \\
       --         & R2D2    & 162.70 & $>$1e3 & 0.56 & 87.16 & \textbf{0.53} & \textbf{82.78} & 24.90 & $>$1e3 & 12.37 & $>$1e3  \\
       --         & SP      & 141.78 & $>$1e3 & 0.48 & 77.65 & \textbf{0.48} & \textbf{76.96} & 0.59 & 101.21 & 5.65 & $>$1e3 \\
       02         & D2Net   & 33.66 & $>$1e3 & 1.29 & 50.81 & \textbf{1.22} & \textbf{48.43} & 1.41 & 57.08 & 1.41 & 55.31  \\
       --         & Key.Net & 85.56 & $>$1e3 & 1.13 & \textbf{42.42} & 1.35 & 53.68 & \textbf{1.10} & 45.29 & 1.48 & 62.15  \\
       --         & R2D2    & 68.08 & $>$1e3 & 1.39 & 43.36 & \textbf{1.00} & \textbf{40.27} & 3.59 & $>$1e3 & 1.38 & 269.12  \\
       --         & SP      & 30.22 & $>$1e3 & \textbf{0.88} & \textbf{35.62} & 0.88 & 35.69 & 1.58 & 69.81 & 1.44 & 532.93  \\
       all        & D2Net   & 30.84 & $>$1e3 & 1.32 & 98.54 & \textbf{1.27} & \textbf{95.03} & 1.82 & 178.20 & 1.77 & 152.29  \\
       --         & Key.Net & 85.53 & $>$1e3 & 1.07 & 74.85 & \textbf{1.00} & \textbf{61.51} & 1.57 & 175.41 & 1.46 & 133.40  \\
       --         & R2D2    & 54.94 & $>$1e3 & 1.07 & 63.13 & \textbf{0.89} & \textbf{61.05} & 4.84 & $>$1e3 & 3.02 & $>$1e3  \\
       --         & SP      & 33.18 & $>$1e3 & 0.80 & 56.57 & \textbf{0.80} & \textbf{56.16} & 1.19 & 81.60 & 1.65 & $>$1e3  \\
       fr1\_xyz   & D2Net   & 20.73 & $>$1e3 & 4.17 & 1.00 & 4.11 & \textbf{0.99} & 4.22 & 1.02 & \textbf{4.06} & 1.00 \\
       --         & Key.Net & 29.98 & $>$1e3 & 4.20 & 1.03 & \textbf{4.09} & \textbf{1.00} & 4.55 & 1.11 & 4.14 & 1.03 \\
       --         & R2D2    & 45.83 & $>$1e3 & 3.56 & 0.89 & 3.57 & 0.89 & \textbf{3.53} & \textbf{0.88} & 3.55 & 0.89 \\
       --         & SP      & 14.67 & $>$1e3 & \textbf{4.15} & \textbf{1.02} & 4.16 & 1.02 & 4.19 & 1.03 & 4.15 & 1.03 \\
       fr1\_rpy   & D2Net   & 129.62 & $>$1e3 & 5.99 & 1.49 & \textbf{5.97} & \textbf{1.48} & 9.10 & 2.40 & 9.09 & 12.16  \\
       --         & Key.Net & 180.33 & $>$1e3 & 8.53 & 1.83 & \textbf{8.03} & \textbf{1.69} & 11.17 & 15.41 & 10.39 & 113.43  \\
       --         & R2D2    & 320.41 & $>$1e3 & \textbf{5.05} & \textbf{1.26} & 5.08 & 1.28 & 5.24 & 1.59 & 5.46 & 2.02  \\
       --         & SP      & 145.67 & $>$1e3 & 11.56 & \textbf{2.80} & \textbf{11.47} & 2.82 & 11.87 & 2.92 & 14.57 & 20.82  \\
       fr1\_360   & D2Net   & 17.90 & $>$1e3 & 5.51 & 1.84 & \textbf{5.48} & \textbf{1.81} & 5.91 & 2.05 & 5.98 & 2.05 \\
       --         & Key.Net & 27.00 & $>$1e3 & 5.91 & 1.99 & \textbf{5.69} & \textbf{1.86} & 10.77 & 154.25 & 6.51 & 2.41 \\ 
       --         & R2D2    & 72.02 & $>$1e3 & \textbf{5.06} & 1.68 & 5.09 & \textbf{1.68} & 5.11 & 1.81 & 5.14 & 1.85 \\
       --         & SP      & 31.01 & $>$1e3 & 6.34 & 2.08 & \textbf{6.19} & \textbf{2.03} & 6.73 & 2.25 & 6.83 & 2.42 \\
       all        & D2Net   & 52.73 & $>$1e3 & 5.18 & 1.44 & \textbf{5.14} & \textbf{1.42} & 6.28 & 1.79 & 6.24 & 4.74  \\
       --         & Key.Net & 75.10 & $>$1e3 & 6.11 & 1.60 & \textbf{5.84} & \textbf{1.51} & 8.70 & 57.70 & 6.86 & 35.99 \\
       --         & R2D2    & 130.76 & $>$1e3 & \textbf{4.51} & \textbf{1.27} & 4.53 & 1.28 & 4.56 & 1.41 & 4.64 & 1.54 \\
       --         & SP      & 60.19 & $>$1e3 & 7.16 & 1.92 & \textbf{7.08} & \textbf{1.91} & 7.39 & 2.02 & 8.24 & 7.54 \\
    \end{tblr}
    \caption{\textbf{Motion estimation from 2D-3D correspondences on KITTI \cite{geiger2013vision} and TUM-RGBD \cite{sturm2012benchmark}}. Sequences are specified at the left-most column (`all' is their aggregation). e$_\text{rot}$ and e$_\text{t}$ are the \textbf{mean} absolute rotation (in $0.1\times$deg.) and translation (in cm.) errors. We compare estimations without using uncertainty (first two columns) and with 2D and 3D uncertainties via the proposed \textit{full} and \textit{iso}tropic covariances. Errors are consistently reduced when using uncertainty estimates. The best result for each sequence-detector is in \textbf{bold}.}
    \label{tab:tumkitti_all}
\end{table*}
}

\definecolor{cvprblue}{rgb}{0.21,0.49,0.74}
\usepackage[pagebackref,breaklinks,colorlinks,citecolor=cvprblue]{hyperref}

\def\paperID{***} %
\def\confName{3DV\xspace}
\def\confYear{2024\xspace}

\title{DAC: Detector-Agnostic Spatial Covariances for Deep Local Features}

\author{Javier Tirado-Garín\\
I3A, University of Zaragoza\\
{\tt\small j.tiradog@unizar.es}
\and
Frederik Warburg\\
Technical University of Denmark\\
{\tt\small frwa@dtu.dk}
\and
Javier Civera\\
I3A, University of Zaragoza\\
{\tt\small jcivera@unizar.es}
}

\begin{document}
\twocolumn[{%
\renewcommand\twocolumn[1][]{#1}%
\maketitle
\begin{center}
    \newcommand{\topBase}{figures/comparison/desktop\_superpoint}
    \newcommand{\bottomBase}{figures/comparison/pont\_d2net}
    \newcommand{\wbig}{5.0cm}
    \newcommand{\wsml}{0.9cm}
    \begin{tblr}{
            colspec={*{5}c},
            width=1.0\linewidth,
            vspan=even,
            colsep=2pt,
            cell{1,4}{1-4}={r=3}{c},
            cell{1,4}{1}={c, cmd={\rotatebox[origin=c]{90}}},
        }
        Superpoint & 
        \includegraphics[width=\wbig, valign=m]{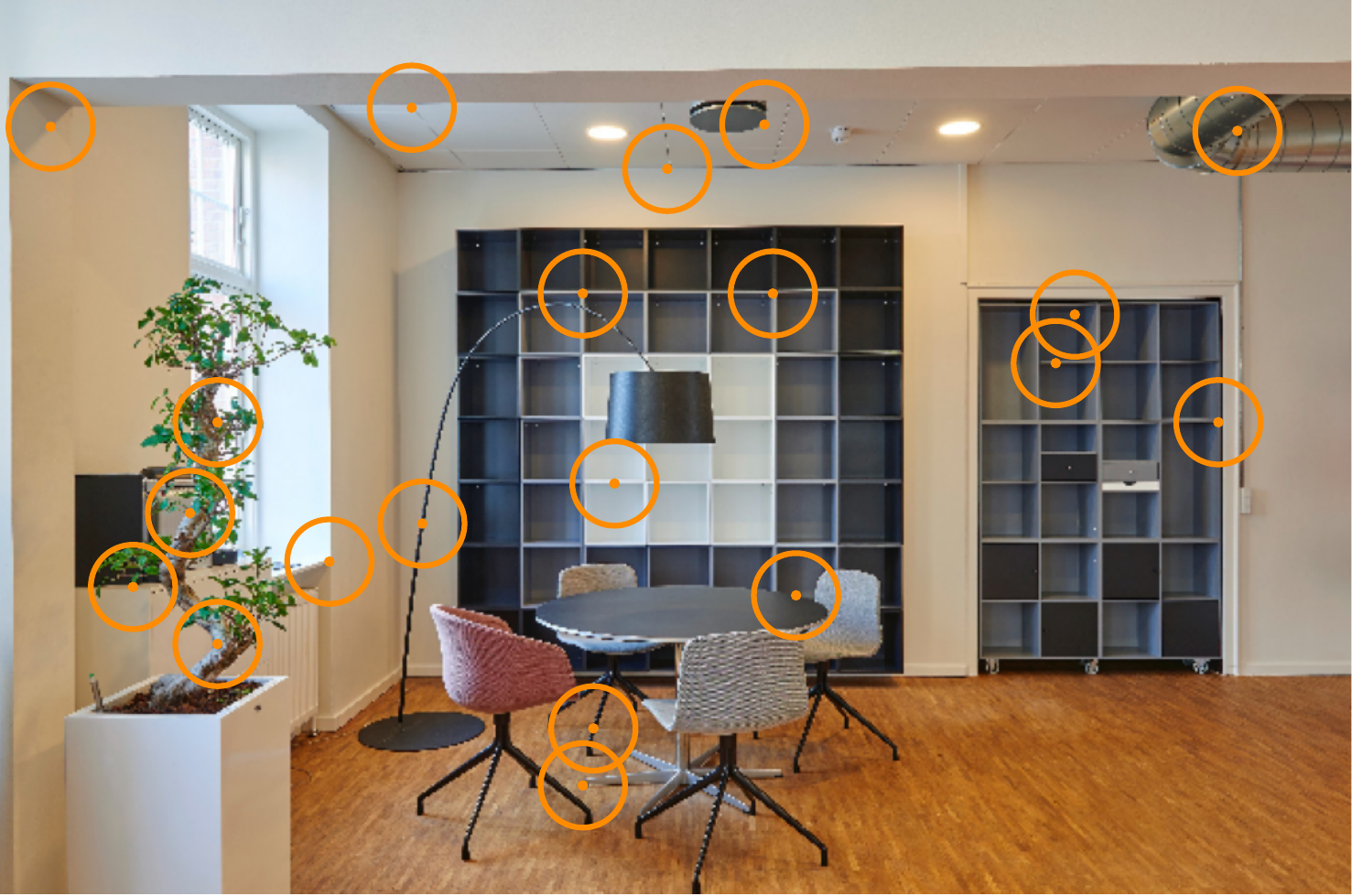} &
        \includegraphics[width=\wbig, valign=m]{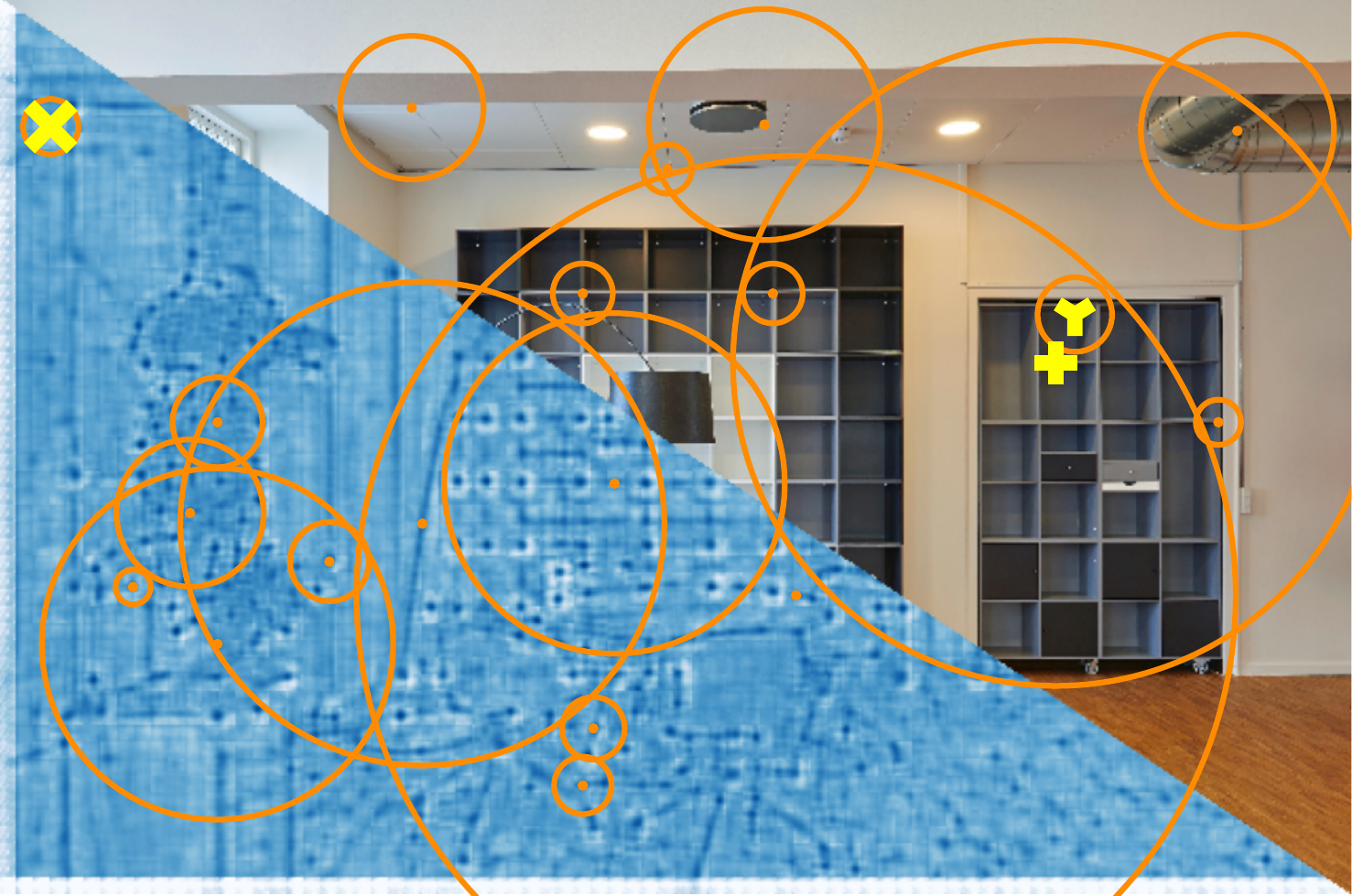} &
        \includegraphics[width=\wbig, valign=m]{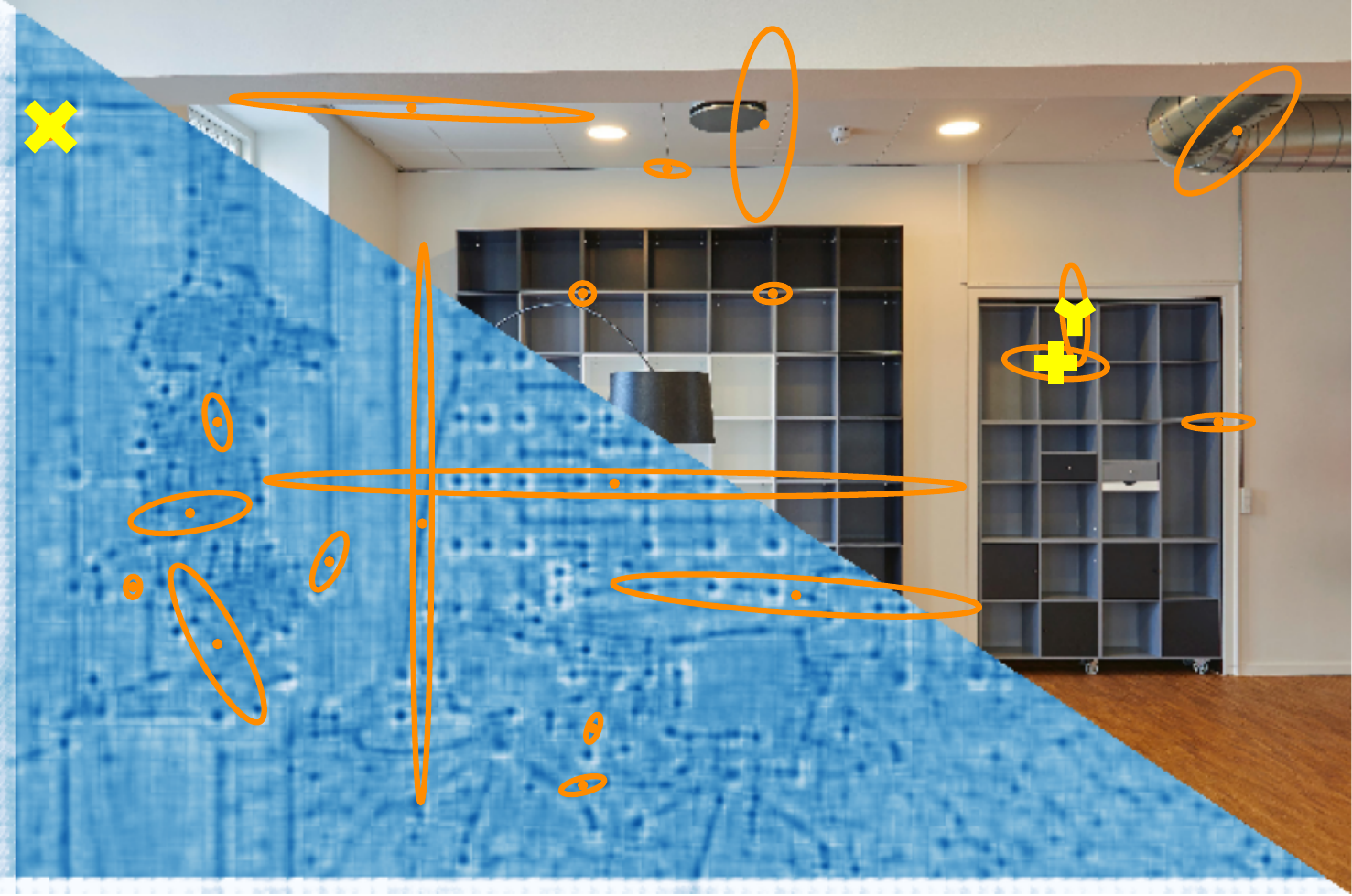} & 
        \includegraphics[width=\wsml, valign=m]{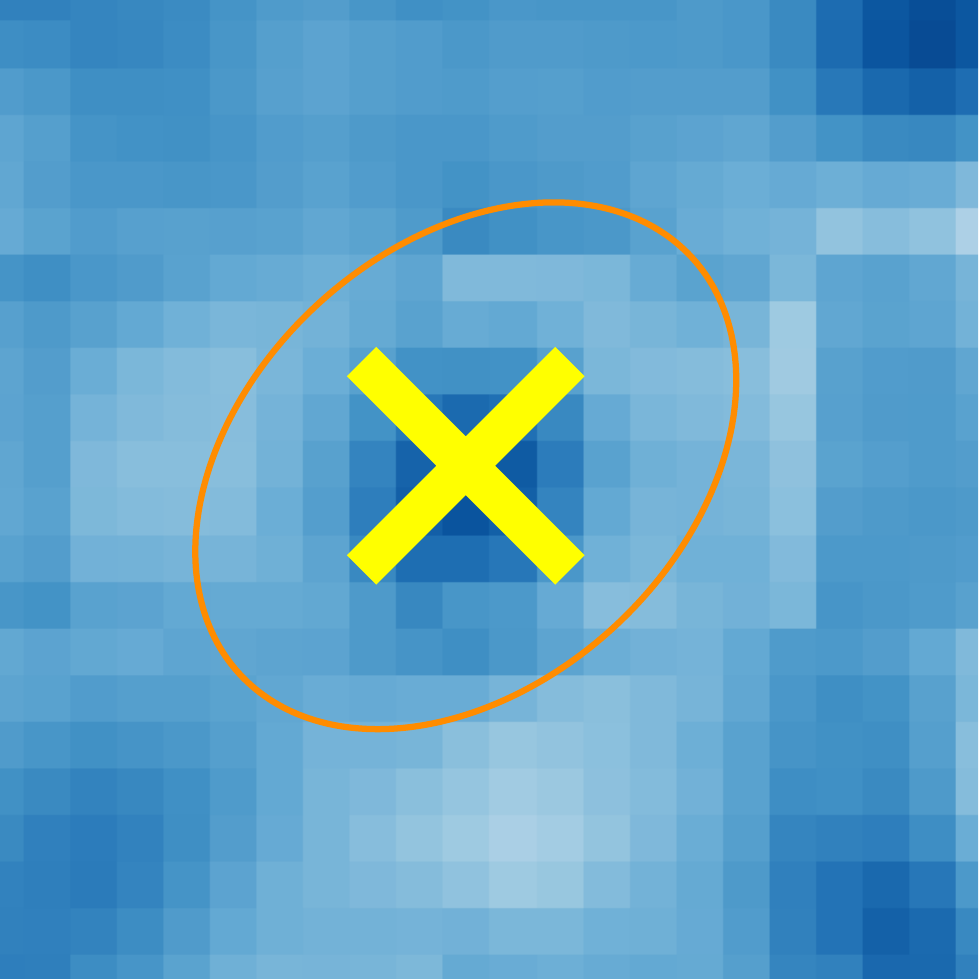}\\
        - & - & - & - &
        \includegraphics[width=\wsml, valign=m]{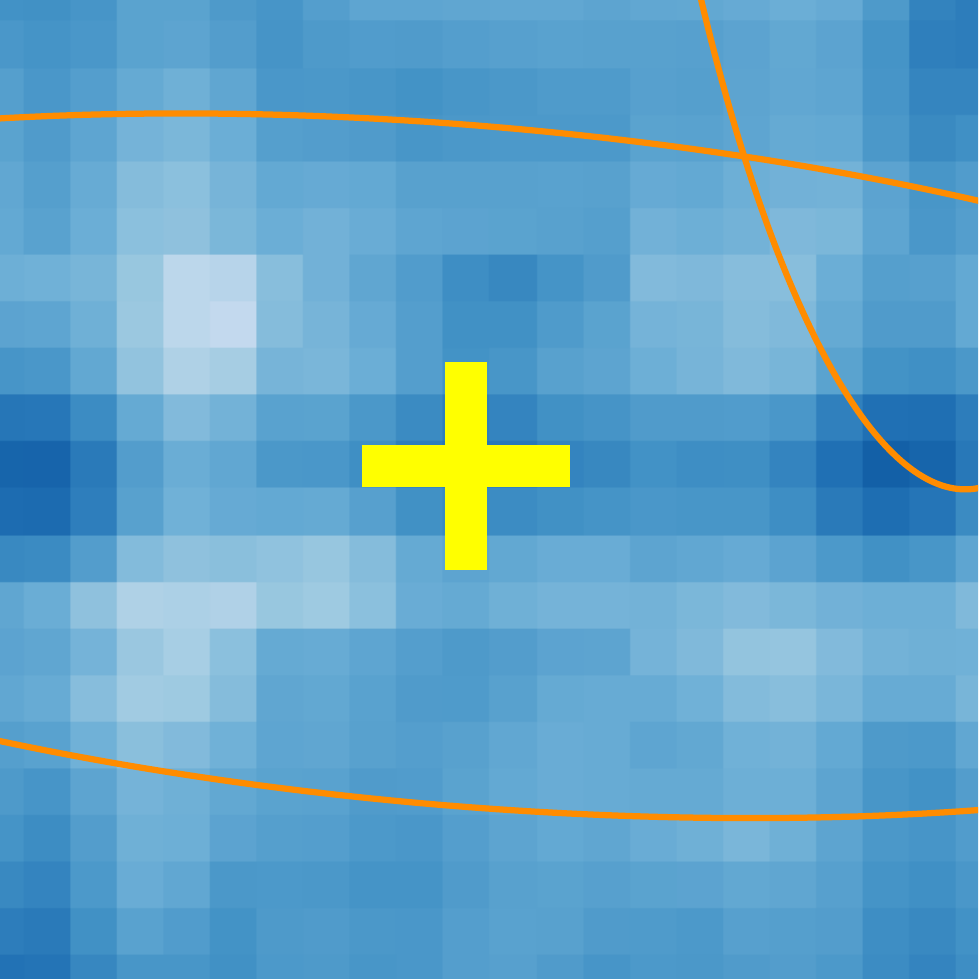}\\
        - & - & - & - &
        \includegraphics[width=\wsml, valign=m]{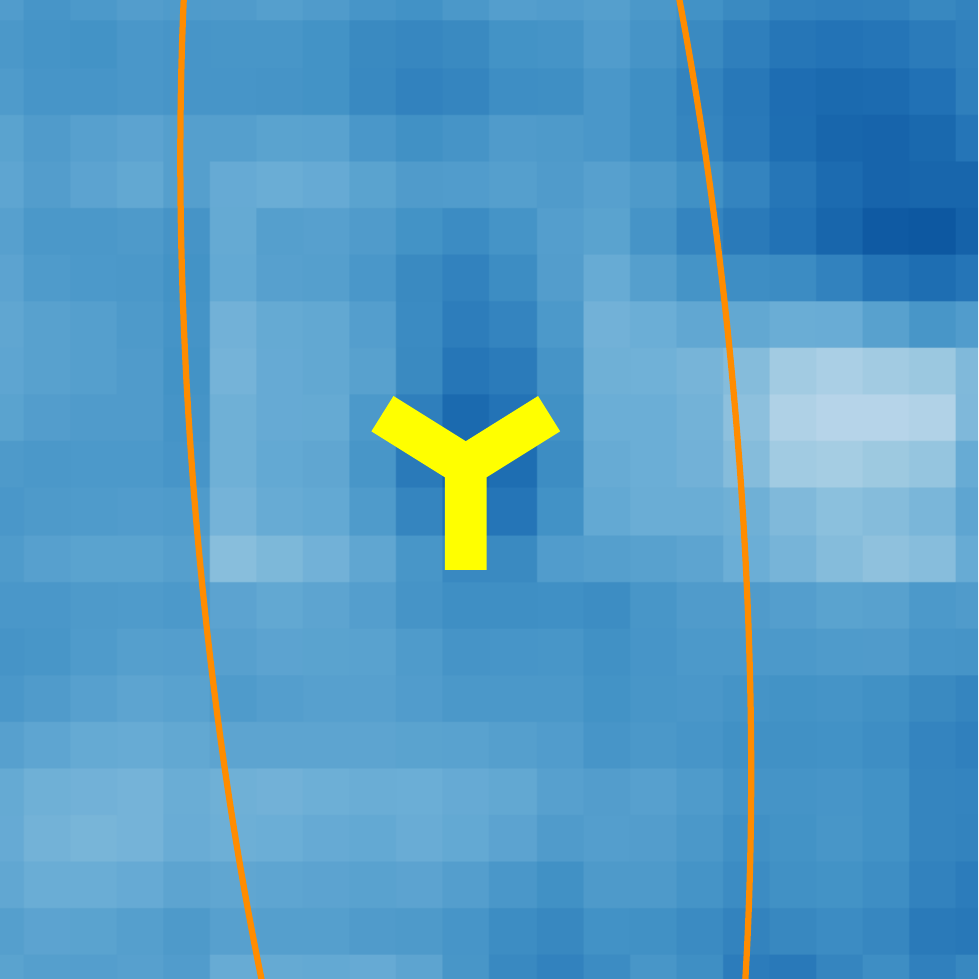}\\
        D2Net &
        \includegraphics[width=\wbig, valign=m]{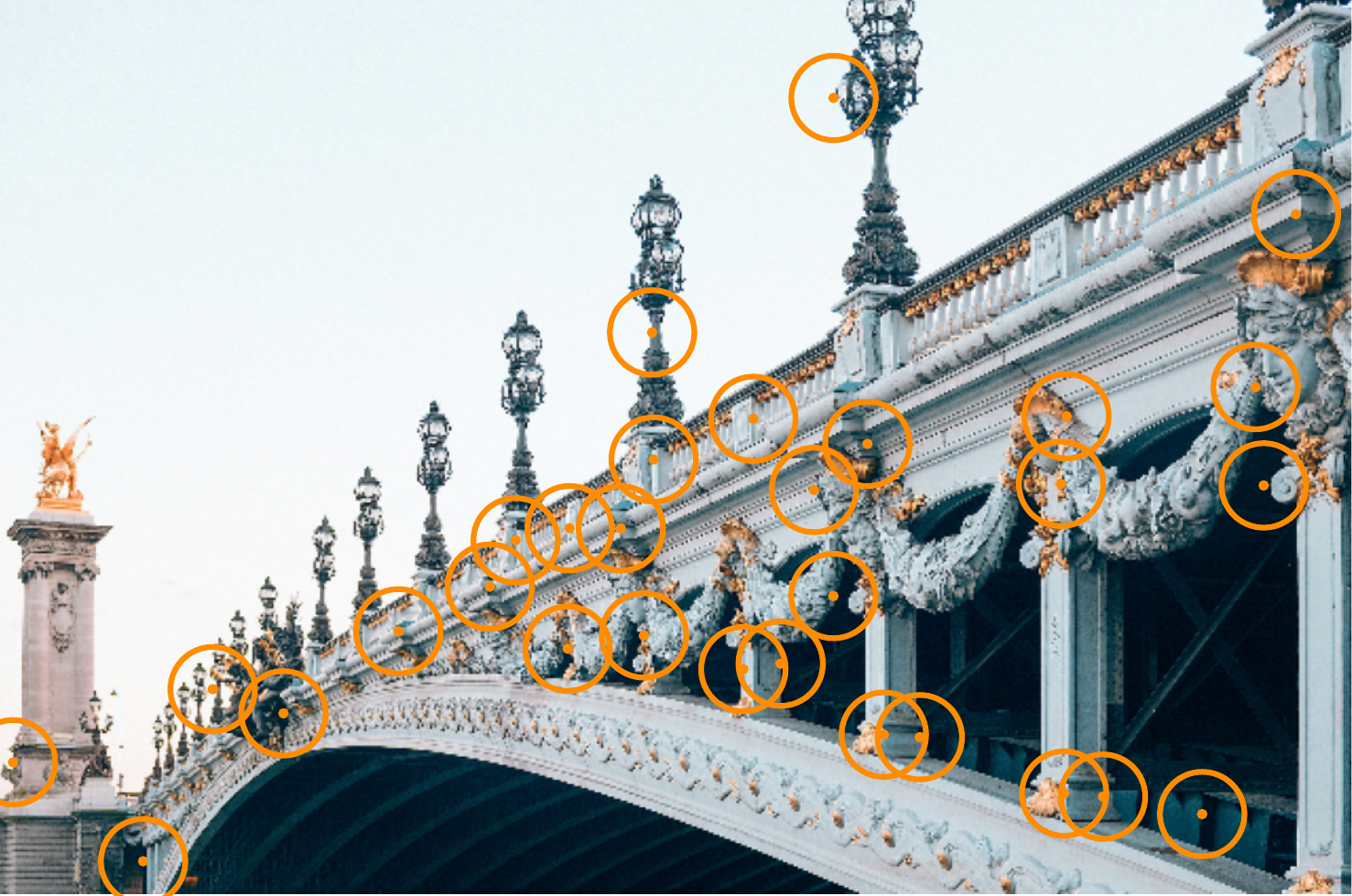} &
        \includegraphics[width=\wbig, valign=m]{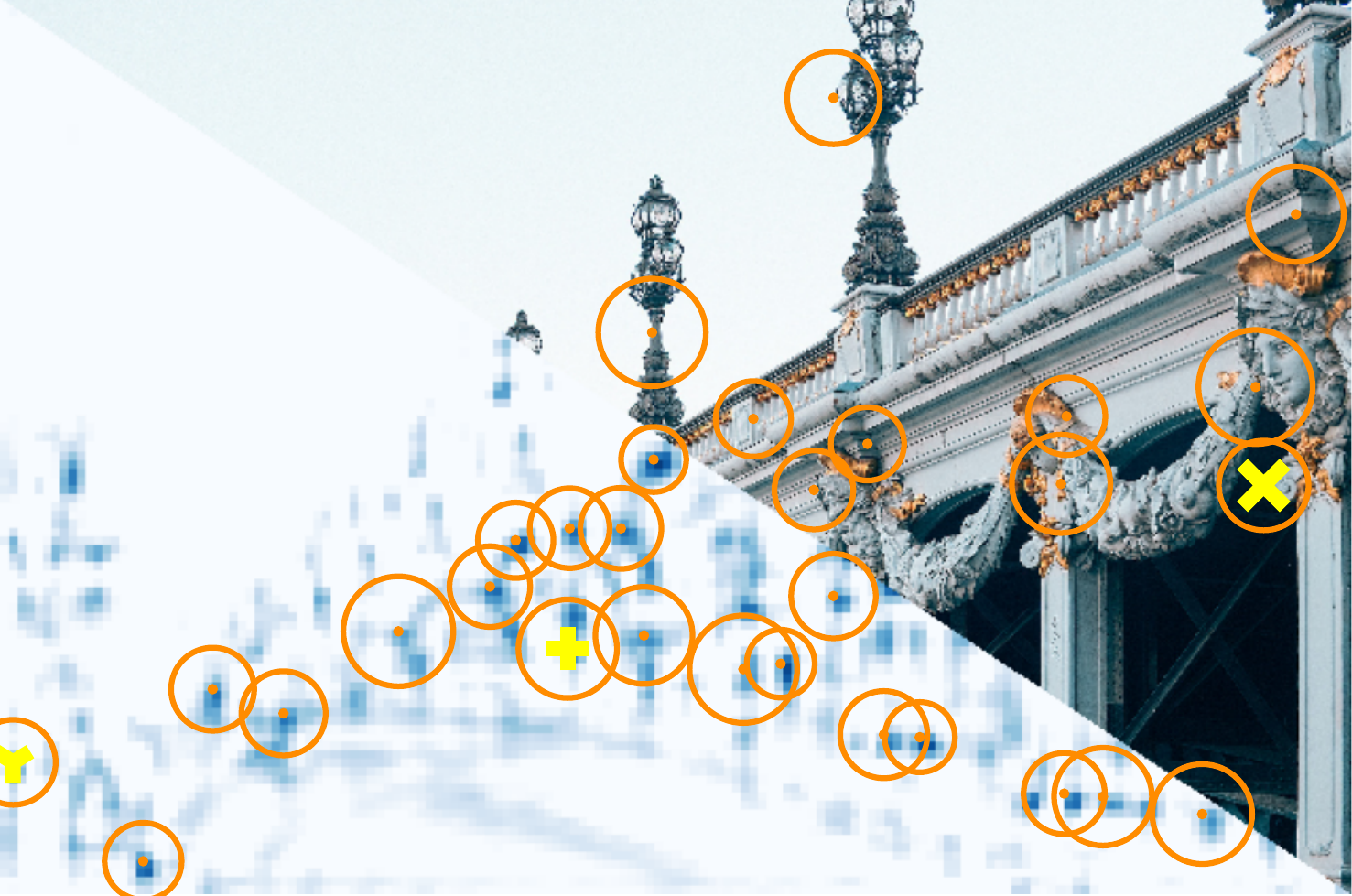} &
        \includegraphics[width=\wbig, valign=m]{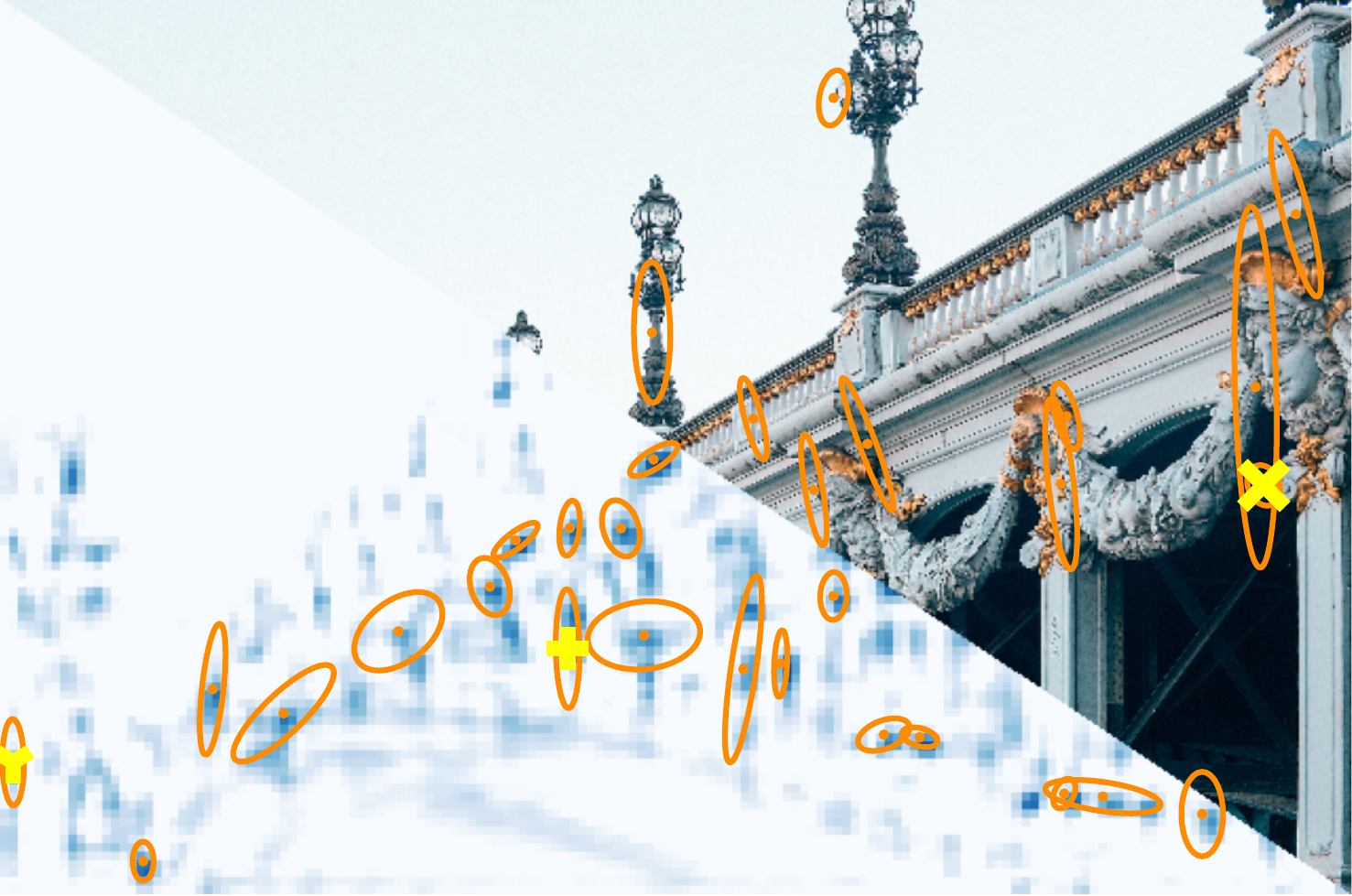} &
        \includegraphics[width=\wsml, valign=m]{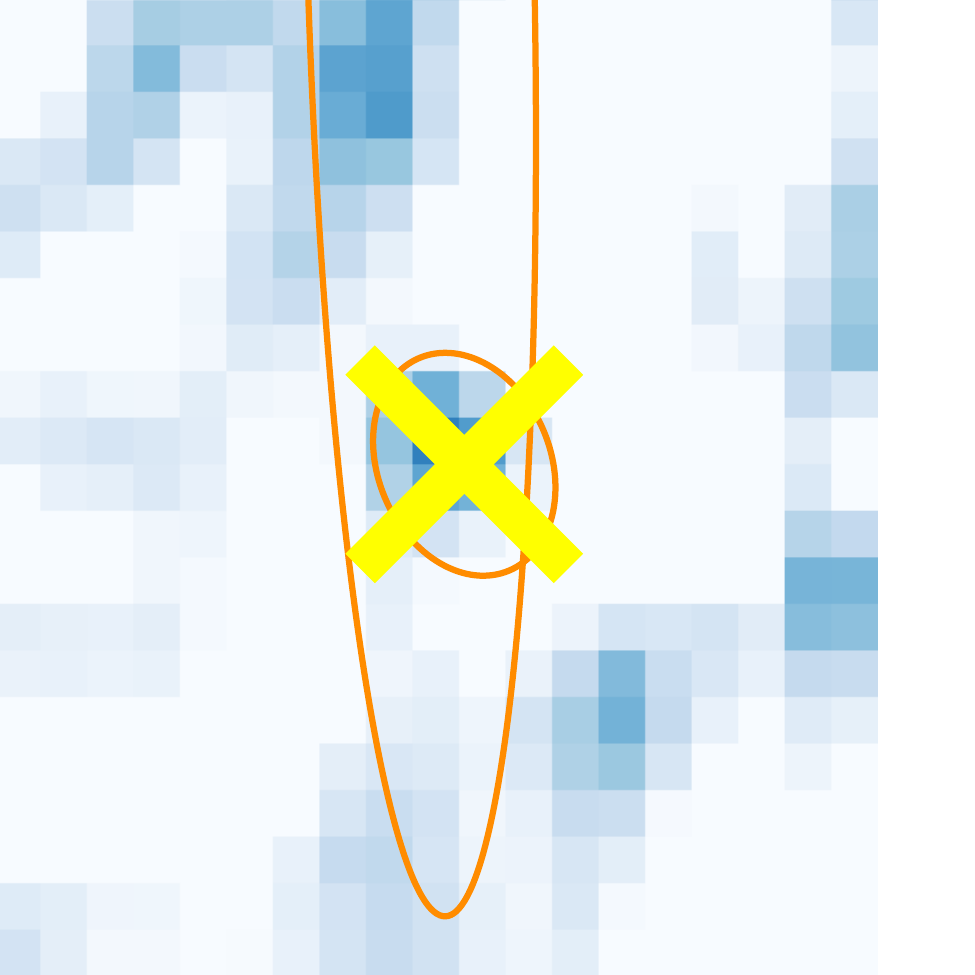}\\
        - & - & - & - &
        \includegraphics[width=\wsml, valign=m]{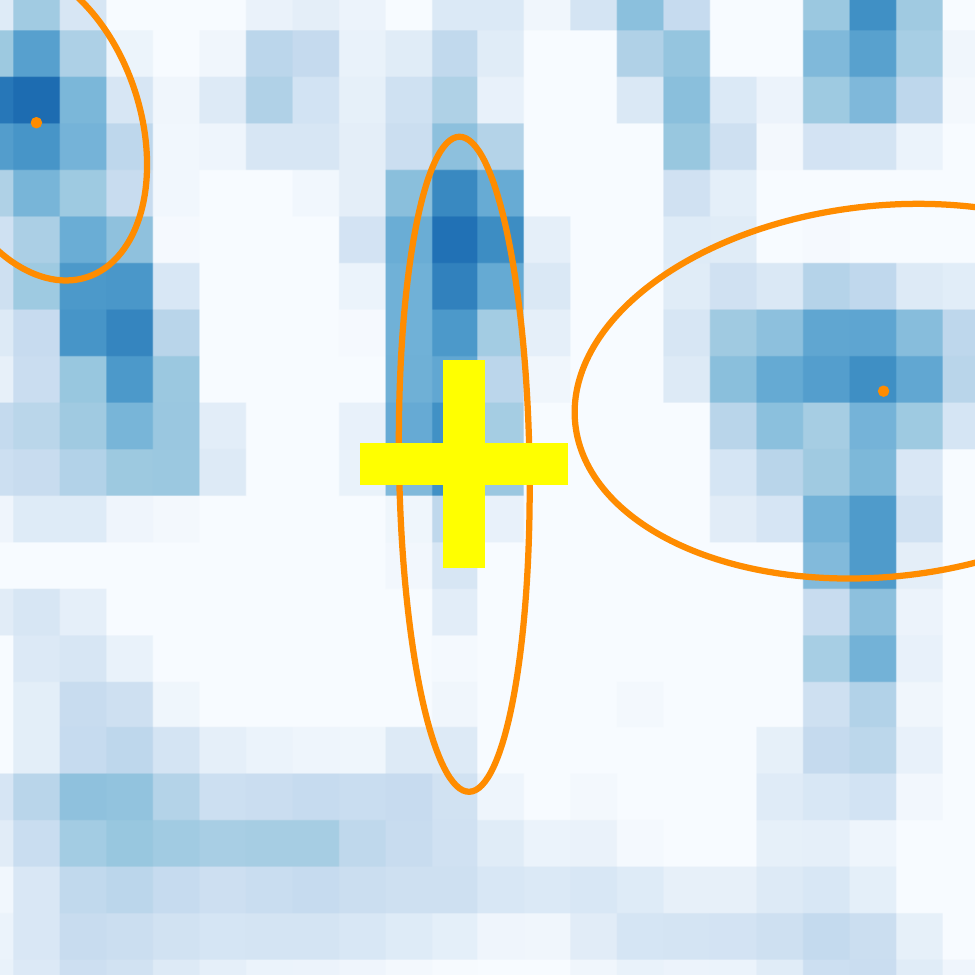}\\
        - & - & - & - &
        \includegraphics[width=\wsml, valign=m]{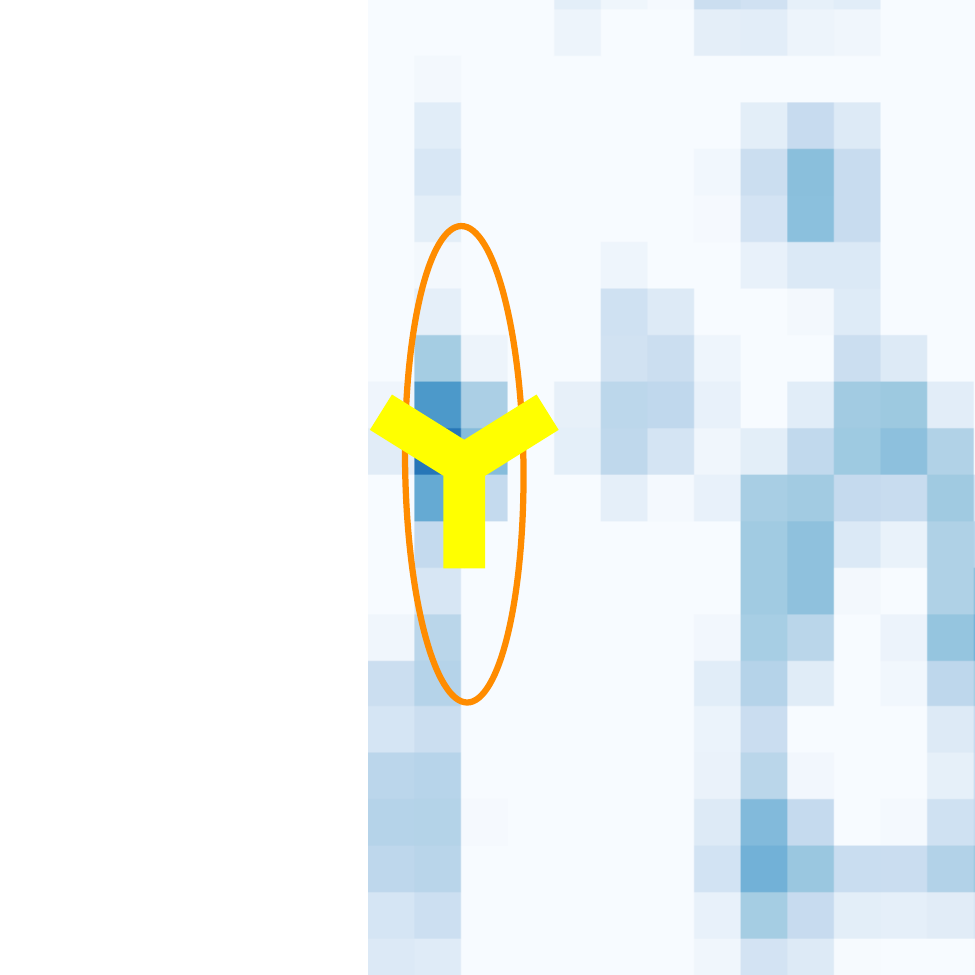}\\
        & {\footnotesize(a) Constant Covariance} & {\footnotesize(b) Isotropic Covariance} & {\footnotesize(c) Full Covariance} &
    \end{tblr}
    \captionof{figure}{\textbf{Detector-agnostic, post-hoc uncertainty for learned local-feature detectors.} State-of-the-art deep feature detectors, such as Superpoint \cite{detone2018superpoint} and D2Net \cite{dusmanu2019d2}, do not estimate the spatial uncertainty of their detections. This corresponds to assuming constant covariances (see (a)), which lead to suboptimal performance. We propose two methods to model local features' uncertainty: (b) a point-wise isotropic covariance and (c) a structure-based full covariance estimate. We demonstrate that modeling uncertainties leads to improved performance in downstream tasks such as solving the perspective-n-point problem and nonlinear optimizations.}
    \label{fig:teaser}
\end{center}%
}]     
\begin{abstract}
Current deep visual local feature detectors do not model the spatial uncertainty of detected features, producing suboptimal results in downstream applications. In this work, we propose two post-hoc covariance estimates that can be plugged into any pretrained deep feature detector: a simple, isotropic 
\iftoggle{cvprfinal}{}{covariance}
estimate that uses the predicted score at a given pixel location, and a full 
\iftoggle{cvprfinal}{}{covariance}
estimate via the local structure tensor of the learned score maps. Both methods are easy to implement and can be applied to any deep feature detector. We show that these covariances are directly related to errors in feature matching, leading to improvements in downstream tasks, including solving the perspective-n-point problem and motion-only bundle adjustment.
\iftoggle{cvprfinal}{Code is available at \url{https://github.com/javrtg/DAC}.}{}
\end{abstract}    
\section{Introduction}\label{sec:introduction}

Estimating a map along with camera poses from a collection of images is a long-standing and challenging problem in computer vision 
\cite{schonberger2016structure, mur2017orb, campos2021orb} with relevant applications in domains such as autonomous driving and augmented reality. Recently, deep visual detectors and descriptors have shown increased resilience to extreme viewpoint and appearance changes \cite{sarlin2019coarse, barbed2022superpoint, sun2022onepose, zhou2022geometry, xu2022airvo}. However, a common limitation among all these deep descriptors and detectors is the lack of a probabilistic formulation for detection noise. Consequently, downstream pose estimation relies on the assumption of constant spatial covariances, as shown in Figure \ref{fig:teaser} (a), leading to suboptimal results.

We propose to model the spatial covariance of detected keypoints in deep detectors. We find that recent detectors share a common design, where a deep convolutional backbone is used to predict a score map that assigns a ``probability'' to a pixel being a point of interest. We exploit this common design space, to propose two post-hoc methods that estimate a covariance matrix for each keypoint detected in any pretrained feature detector. Our simplest method uses the score at a given pixel to initialize an isotropic covariance matrix. We illustrate the learned score maps (lower triangle) overlaid on exemplar images, along with our proposed isotropic covariances in Figure \ref{fig:teaser} (b) for two state-of-the-art deep feature detectors, Superpoint \cite{detone2018superpoint} and D2Net \cite{dusmanu2019d2}. We also suggest a theoretically-motivated method to estimate the full spatial covariance matrix of detected keypoints using the local structure tensor. The structure tensor models the local saliency of the detections in the score map.
Figure \ref{fig:teaser} (c) shows the deduced full covariances. These covariances capture the larger uncertainty along edges on the learned score map. To the best of our knowledge, we are the first to model spatial uncertainties of deep detectors.

We show in a series of experiments that our proposed methods for modeling the spatial covariances of detected features are directly related with the errors in matching. Accounting for them allows us to improve performance in tasks such as solving the perspective-n-point (PnP) problem and nonlinear optimizations. 

\section{Related Work}\label{sec:rel_work}
We propose to model the spatial covariances of \emph{learned} local features. Modeling the spatial uncertainty of features is not a new idea, and has been studied for \emph{hand-crafted} detectors. In this section, we first review hand-crafted methods and how uncertainty estimation has been proposed for them. We then describe recent progress in learned detectors.

\PAR{Handcrafted local features.}
Harris \etal \cite{harris1988combined} pioneeres rotationally-invariant corner detection by using a heuristic over the eigenvalues of the local structure tensor, while Shi et al. \cite{shi1994good} proposes its smallest eigenvalue for detection. Mikolajczyk and Schmid \cite{mikolajczyk2004scale} robustifies it against scale and affine transformations. On the other hand, SIFT (DoG) \cite{lowe2004distinctive} popularizes detection (and description) of \emph{blobs}. SURF \cite{bay2008speeded} reduces its execution time by using integral images and (A-)KAZE \cite{alcantarilla2012kaze, Alcantarilla2013FastED} proposes non-linear diffusion to improve the invariance to changes in scale. Lastly, FAST~\cite{rosten2008faster} and its extensions \cite{rublee2011orb, leutenegger2011brisk} stand out for achieving the lowest execution times, thanks to only requiring intensity comparisons between the neighboring pixels of an image patch. 

\PAR{Uncertainty quantification in handcrafted detectors.} Inclusion of spatial uncertainty of local features for estimation of geometry has been extensively studied \cite{kanatani1996statistical, triggs2000bundle, Hartley2004, forstner2016photogrammetric}. However, its \emph{quantification} on \emph{classical} local features is still recognized as an open problem in the literature \cite{kanatani2004geometric, kanatani2008statistical, ferraz2014leveraging} and which has not been addressed yet with \emph{learned} detections. Several works have shown the benefits of a precise quantification of uncertainty. For this purpose, they adapt the quantifications to specific classical detectors \cite{forstner1987fast, brooks2001value, kanazawa2001we, zeisl2009estimation, vakhitov2021uncertainty, muhle2022probabilistic}, assume planar surfaces \cite{ferraz2014leveraging, peng2019calibration} and require an accurate offline calibration \cite{fontan2022model, fontan2023sid}. 
Only recently \cite{Muhle_2023_CVPR} proposes a learning-based approach of spatial covariances, trained per detector, in order to weigh normalized epipolar errors \cite{longuet1981computer, lee2020geometric}.
In contrast to previous works, we propose a general formulation for their quantification, directly applicable on state-of-the-art learned detectors, seamlessly fitting them by leveraging their common characteristics.

\PAR{Learned local features and lack of uncertainty.} The dominant approach \cite{verdie2015tilde, yi2016lift, detone2018superpoint, dusmanu2019d2, revaud2019r2d2, tyszkiewicz2020disk, luo2020aslfeat, barroso2022key} consists on training CNNs to regress a score map over which detections are extracted via non-maximum-suppresion (NMS). 
Superpoint \cite{detone2018superpoint} is an efficient detector-descriptor of corners, robust to noise and image transformations thanks to a synthetic pre-training followed by \emph{homographic adaptation} on real images. D2Net \cite{dusmanu2019d2} shows the applicability, to the problem of local feature detection and description, of classification networks \cite{simonyan2015a}. R2D2 \cite{revaud2019r2d2} proposes a \emph{reliability} measure, used to discard unmatchable points. Similarly, DISK~\cite{tyszkiewicz2020disk} bases its learning on matches, and KeyNet \cite{barroso2022key}, inspired by classical systems,  proposes to learn from spatial image gradients. This work experiments with Superpoint, D2Net, R2D2, and KeyNet, but our proposal is applicable to the rest of the systems. Finally, recent works \cite{sun2021loftr, jiang2021cotr, tang2022quadtree} exploit \emph{attention} mechanisms to match pixels without explicitly using detectors. Although their inclusion in geometric estimation pipelines can be engineered \cite{shen2022semi}, they suffer from lack of repeatability. Because of this, in this work, we focus on the quantification of spatial uncertainty for learned \emph{detectors}.

\begin{figure}[t]
    \centering
    \includegraphics[width=1.0\linewidth]{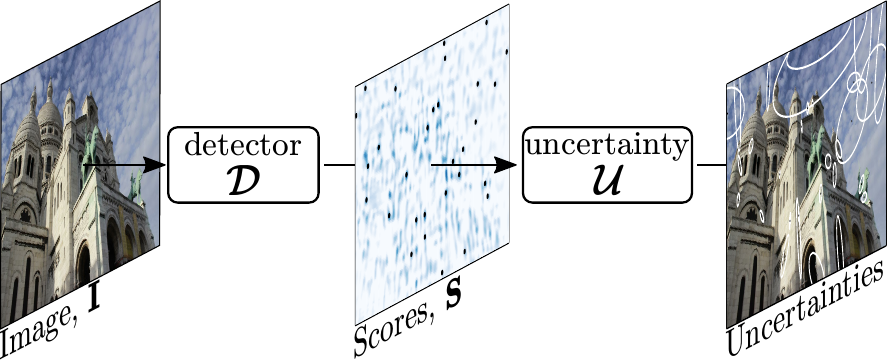}
    \caption{%
        \textbf{Method overview}. By exploiting the dominant approach in \emph{learned} local-feature detectors (represented by $\mathcal{D}$)  of regressing, for each input image $\bI$, a map of scores $\bS\ceqq\mathcal{D}(\bI)$, over which detections are done, we propose to quantify the uncertainty of each detected location $\bx_i$ by a mapping $\mathcal{U}(\bS, \bx_i)$ agnostic to $\mathcal{D}$.}
    \label{fig:model_overview}
\end{figure}
\section{Method}\label{sec:method}
Our framework takes as input a RGB image $\bI$ of spatial dimensions $H\times W$, and outputs local features together with their \emph{spatial} uncertainties. It is composed of two main components as shown in Fig. \ref{fig:model_overview}. (1) A pretrained feature detector and (2) our novel and detector-agnostic uncertainty \mbox{module}. Our methods work with \emph{any} pretrained state-of-the-art detectors, taking their learned score maps as input, and outputting reliable covariance estimates for each detected keypoint. We show that the estimated covariances are well-calibrated and improve downstream tasks such as PnP and motion-only bundle adjustment. %

\subsection{Overview}

\paragraph{Pre-trained local-feature detector.}%
Our framework is directly applicable to the vast majority of learned detectors \cite{verdie2015tilde, yi2016lift, detone2018superpoint, revaud2019r2d2, dusmanu2019d2, luo2020aslfeat, tyszkiewicz2020disk, barroso2022key}. These detectors share a common architectural design, using convolutional backbones to predict a score map $\bS\in\bb{R}^{kH\times kW}$ with $k\in\bb{R}^{+}$, followed by non-maximum-suppression (NMS) to extract a sparse set of features. We leverage this standard design to make a detector-agnostic, post-hoc uncertainty module that takes the estimated score maps as input and estimates the spatial uncertainty based on the \emph{peakiness} in a local region around the detected point. Our approach can be applied to these detectors without any training or fine-tuning.

\PAR{Detector-Agnostic Feature Uncertainties.}  
Instead of considering the 2D position of detections as deterministic locations, we propose to model their spatial uncertainties.
More formally, we consider that the spatial location $\bx_i\in\bb{R}^2$ of a local feature $i$, detected in $\bI$, stems from perturbing its \emph{true} location $\bx_{i, \mathrm{true}}\in\bb{R}^{2}$, with random noise $\bxi_i\in\bb{R}^2$
\begin{equation}\label{eq:base_model}
    \bx_i = \bx_{i,\mathrm{true}} + \bxi_i~,
\end{equation}
whose probability distribution we want to estimate. We follow the dominant model in computer vision \cite{kanatani1996statistical, triggs2000bundle, Hartley2004}, which uses \emph{second order statistics} to describe the spatial uncertainty of each location
\begin{align}\label{eq:unbiased_location}
    \E{\bxi_i} = \bzero~, 
    \quad
    \bSigma_{i} \ceqq \text{Cov}(\bxi_i) = \E{\bxi_i\bxi_i^\top}~,
    \quad \forall i~.
\end{align}

\begin{figure}[tp!]
    \centering
    \def\svgwidth{0.98\linewidth}
    \begingroup%
  \makeatletter%
  \providecommand\color[2][]{%
    \errmessage{(Inkscape) Color is used for the text in Inkscape, but the package 'color.sty' is not loaded}%
    \renewcommand\color[2][]{}%
  }%
  \providecommand\transparent[1]{%
    \errmessage{(Inkscape) Transparency is used (non-zero) for the text in Inkscape, but the package 'transparent.sty' is not loaded}%
    \renewcommand\transparent[1]{}%
  }%
  \providecommand\rotatebox[2]{#2}%
  \newcommand*\fsize{\dimexpr\f@size pt\relax}%
  \newcommand*\lineheight[1]{\fontsize{\fsize}{#1\fsize}\selectfont}%
  \ifx\svgwidth\undefined%
    \setlength{\unitlength}{469.44629407bp}%
    \ifx\svgscale\undefined%
      \relax%
    \else%
      \setlength{\unitlength}{\unitlength * \real{\svgscale}}%
    \fi%
  \else%
    \setlength{\unitlength}{\svgwidth}%
  \fi%
  \global\let\svgwidth\undefined%
  \global\let\svgscale\undefined%
  \makeatother%
  \begin{picture}(1,0.55334129)%
    \lineheight{1}%
    \setlength\tabcolsep{0pt}%
    \put(0,0){\includegraphics[width=\unitlength,page=1]{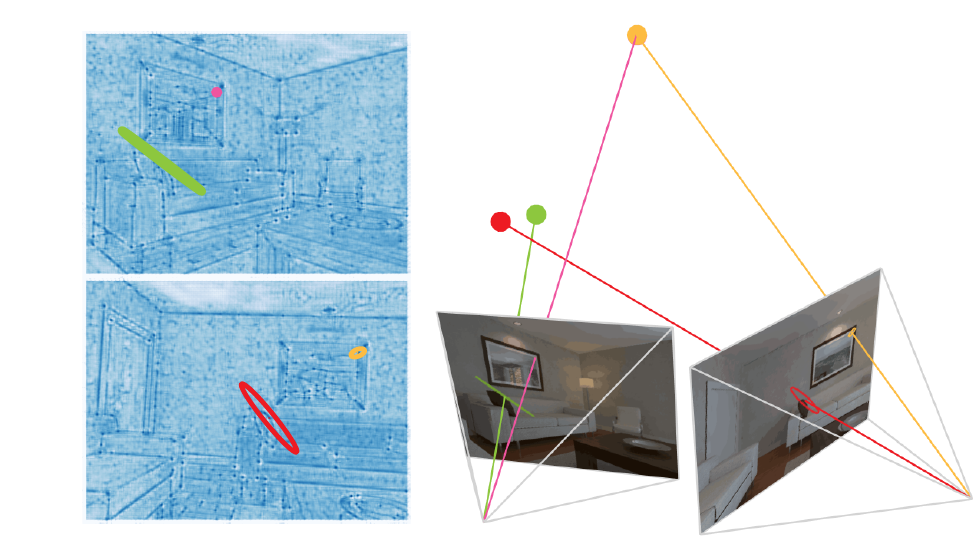}}%
    \put(0.02280363,0.27633242){\color[rgb]{0,0,0}\rotatebox{90}{\makebox(0,0)[t]{\lineheight{1.25}\smash{\begin{tabular}[t]{c}local feat. scores\end{tabular}}}}}%
    \put(0,0){\includegraphics[width=\unitlength,page=2]{figures/heatmap_ells_blues.pdf}}%
    \put(0.05132668,-0.00249055){\color[rgb]{0,0,0}\makebox(0,0)[t]{\lineheight{1.25}\smash{\begin{tabular}[t]{c}$-$\end{tabular}}}}%
    \put(0.05132668,0.51349435){\color[rgb]{0,0,0}\makebox(0,0)[t]{\lineheight{1.25}\smash{\begin{tabular}[t]{c}$+$\end{tabular}}}}%
    \put(0,0){\includegraphics[width=\unitlength,page=3]{figures/heatmap_ells_blues.pdf}}%
  \end{picture}%
\endgroup%

    \caption{\textbf{Influence of 2D uncertainties in 3D geometry.} The learned score maps for two images and matched local features with highest and lowest estimated uncertainty ---represented w. ellipses--- for two points detected in two views. More uncertain 2D points present significantly higher 3D reprojection errors. %
    }
    \label{fig:simple_eg}
\end{figure}

Thereby our goal is to quantify each \emph{covariance matrix} $\bSigma_{i}$. To this end, we propose two methods: (1) A point-wise estimator of \emph{isotropic} covariances (Sec.~\ref{subsec:pointwise}) that uses the inverse of the scores at the detected point, and (2) a \emph{full} covariance estimator based on the local structure tensor (Sec.~\ref{subsec:tensor}) which models the local saliency %
to estimate the uncertainty in all directions. We show that both approaches lead to reliable uncertainty estimates, benefiting downstream tasks. Intuitively, a \emph{peaky} score map at the location of a detected feature will yield low spatial uncertainty, whereas a flatter score map will yield larger spatial uncertainty. Figure~\ref{fig:simple_eg} qualitatively shows how our deduced 2D uncertainties relate to 3D uncertainties.
In the Supplemental we explore the agnostic behavior of our covariances.

\subsection{Point-wise Estimation}\label{subsec:pointwise}

As the simplest estimator of the spatial uncertainty, we propose to use  the regressed score of each local feature to create an \emph{isotropic} covariance matrix
\begin{equation}\label{eq:met_puntual}
    \bSigma_i \ceqq \frac{1}{\bS(\bx_i)}~\bI_{2\times2} 
    = \begin{bmatrix}
        1/\bS(\bx_i) & 0 \\
        0 & 1/\bS(\bx_i)
    \end{bmatrix}~.
\end{equation}
Fig~\ref{fig:teaser} shows that this estimator yields isotropic predictions of uncertainty (equal in all directions), so it only quantifies the relative scale regardless of the  learned local structure.

\subsection{Structure Tensor}\label{subsec:tensor}

\begin{figure}
    \centering
    \def\svgwidth{0.98\linewidth}
    \begingroup%
  \definecolor{myGreen}{rgb}{0.216, 0.902, 0.216}
  \definecolor{myRed}{rgb}{0.820, 0.0, 0.0}
  \definecolor{myBlue}{rgb}{0.0, 0.0, 0.820}
  \makeatletter%
  \providecommand\color[2][]{%
    \errmessage{(Inkscape) Color is used for the text in Inkscape, but the package 'color.sty' is not loaded}%
    \renewcommand\color[2][]{}%
  }%
  \providecommand\transparent[1]{%
    \errmessage{(Inkscape) Transparency is used (non-zero) for the text in Inkscape, but the package 'transparent.sty' is not loaded}%
    \renewcommand\transparent[1]{}%
  }%
  \providecommand\rotatebox[2]{#2}%
  \newcommand*\fsize{\dimexpr\f@size pt\relax}%
  \newcommand*\lineheight[1]{\fontsize{\fsize}{#1\fsize}\selectfont}%
  \ifx\svgwidth\undefined%
    \setlength{\unitlength}{485.37497791bp}%
    \ifx\svgscale\undefined%
      \relax%
    \else%
      \setlength{\unitlength}{\unitlength * \real{\svgscale}}%
    \fi%
  \else%
    \setlength{\unitlength}{\svgwidth}%
  \fi%
  \global\let\svgwidth\undefined%
  \global\let\svgscale\undefined%
  \makeatother%
  \begin{picture}(1,0.4615267)%
    \lineheight{1}%
    \setlength\tabcolsep{0pt}%
    \put(0.50059755,0.42708974){\color[rgb]{0,0,0}\makebox(0,0)[t]{\lineheight{1.25}\smash{\begin{tabular}[t]{c}$%
	\phi([\textcolor{myRed}{\gradp_xS}, \textcolor{myGreen}{\gradp_yS}])=[\textcolor{myRed}{(\gradp_xS)^2}, \textcolor{myGreen}{(\gradp_yS)^2}, \textcolor{myBlue}{\gradp_xS\gradp_yS}]$\end{tabular}}}}%
    \put(0,0){\includegraphics[width=\unitlength,page=1]{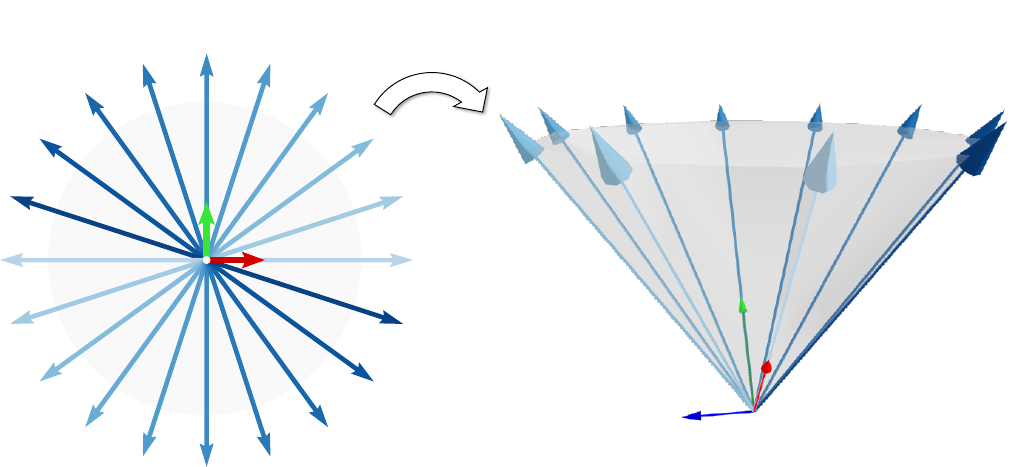}}%
  \end{picture}%
\endgroup%

    \caption{\textbf{Structure tensor as kernel.} The direction of local gradients $[\gradp_xS, \gradp_yS]\ceqq\pt\bS/\pt\bx$ can be averaged without loss of information with the local structure tensor. It acts as a kernel $\phi: \bb{R}^2 \mapsto \bb{R}^3$, mapping vectors with $180^\circ$ of difference (opposite directions) to the same point in $\bb{R}^3$. }%
    \label{fig:structure_tensor}
\end{figure}

\qualitativeComparison[htp!]

Quantification of local saliency motivates the use of the local structure tensor, $\bC_i\in\bb{R}^{2\times2}$ \cite{harris1988combined}. Defining $[\gradp_xS_i, \gradp_yS_i] \coloneqq \partial \bS / \partial \bx |_{\bx_i}$ as the spatial gradient of $\bS$ evaluated at $\bx_i$, $\bC_i$ in its local neighborhood $\calW_i$ (window of size $u\times v$) is given by
\begin{equation}\label{eq:tensor}
\resizebox{0.9\linewidth}{!}{%
    ${\displaystyle%
    \bC_i \ceqq 
    \sum_{j\in\calW_i} w_j 
    \left. \frac{\pt\bS}{\pt\bx}\right|_{\bx_j}^\top
    \left. \frac{\pt\bS}{\pt\bx}\right|_{\bx_j} =
    \sum_{j\in\mathcal{W}_i} w_j \begin{bmatrix}
        (\gradp_xS_j)^2 & \gradp_xS_j\gradp_yS_j \\[0.4em]
        \gradp_yS_j\gradp_xS_j & (\gradp_yS_j)^2
    \end{bmatrix}~,%
    }$%
}%
\end{equation}

with $w_j\in\bb{R}^{+}$ being the weight of pixel $j$, preferably defined by a Gaussian centered at $\bx_i$ \cite{sanchez2018analysis}. As such, $\bC_i$ is a positive semidefinite (PSD) matrix, resulting from averaging the directionality of gradients and hence not canceling opposite ones (see Fig.~\ref{fig:structure_tensor}).

The reason why $\bC_i$ captures the local saliency lies in the auto-correlation function, $c: \bb{R}^{u\times v}\mapsto \bb{R}$, which averages local changes in $\bS$ given small displacements $\delta\bx$ \cite{harris1988combined}:
\begin{equation}
    c_i =  \sum_{j\in\calW_i} w_j (\bS(\bx_j) - \bS(\bx_j+\dbx))^2~,
\end{equation}
with $\bS(\bx_j)$ indicating the score at $\bx_j$. Linearly approximating $\bS(\bx_j+\dbx)\approx\bS(\bx_j)+\partial \bS / \partial \bx |_{\bx_j}\dbx$, yields
\begin{align}
    c_i 
    & \approx \sum_{j\in\calW_i} 
    w_j ( \bS(\bx_j) - \bS(\bx_j) - 
    \left.\frac{\pt\bS}{\pt\bx} \right|_{\bx_j} \dbx )^2~,\\
    & = \sum_{j\in\calW_i} w_j 
    \dbx^\top 
    \left.\frac{\pt\bS}{\pt\bx} \right|_{\bx_j}^\top  
    \left.\frac{\pt\bS}{\pt\bx} \right|_{\bx_j} \dbx 
    = \dbx^\top \bC_i \dbx~.
\end{align}
Thus, extreme saliency directions are obtained by solving:
\begin{equation}
    \bigg\{
    \begin{array}{l}
        \displaystyle\max_{\dbx}\\
        \displaystyle\min_{\dbx}
    \end{array}
    \quad
    \dbx^\top\bC_i \dbx, \qquad \mathrm{s.t.} \quad \norm{\dbx}=1~,
\end{equation}
where the constraint $\norm{\dbx}=1$ ensures non-degenerate directions, which can be obtained with Lagrange multipliers \cite{rockafellar1993lagrange} \ie by defining the Lagrangian
\begin{equation}\label{eq:lagrangian}
    \mathcal{L}(\dbx, \lambda) \ceqq \dbx^\top\bC_i\dbx - \lambda(\dbx^\top\dbx-1)~,
\end{equation}
differentiating w.r.t. $\dbx$ and setting it to $\bzero$:
\begin{equation}\label{eq:dlagrangian}
    2\dbx^\top\bC_i - 2\lambda\dbx^\top = \mathbf{0} \Rightarrow
    \bC_i\dbx = \lambda\dbx~,%
\end{equation}
we conclude that directions of extreme saliency correspond to the eigenvectors of $\bC_i$. Since inverting a matrix, does the same to its eigenvalues without affecting its eigenvectors\footnote{Let $\mathbf{A}\in\bb{R}^{m\times m}$ and $\det\mathbf{A}\neq0$, then $\lambda^{-1}\mathbf{v}=\mathbf{A}^{-1}\mathbf{v}$, with $\lambda$ and $\mathbf{v}$ being the eigenvalues and eigenvectors of $\mathbf{A}$.}, $\bSigma_i\ceqq\bC_i^{-1}$ results in a proper covariance matrix (PSD) assigning greater uncertainty in the direction of less saliency and vice-versa. %

\PAR{Statistical interpretation.}
Under a Gaussian model of aleatoric uncertainty, common in deep learning \cite{kendall2017uncertainties, gustafsson2020evaluating} and motivated by the principle of maximum entropy \cite{meidow2009reasoning} of the local features' scores, distributed independently by
\begin{equation}
    \mathcal{N}(\bS(\bx_i+\bt_i),~\sigma^2)~,
\end{equation}
inspired by \cite{kanazawa2001we}, we can set up a parametric optimization of $\bt\in\bb{R}^2$ which maximizes the \emph{likelihood}, $\mathcal{L}(\bt_i\mid\mathcal{S})$ with $\mathcal{S}\coloneqq \{\rand{\bS}(\bx_j)\mid\bx_j\in\mathcal{W}_i\}$, of the observations:
\begin{equation}\label{eq:uncertainty_model}
    \rand{\bS}(\bx_i) = \bS(\bx_i + \bt_i) + \eps_i~, 
    \quad \mathrm{with} \quad 
    \eps_i\sim\mathcal{N}(0,~\sigma^2)~,%
\end{equation}
where $\rand{\bS}(\bx_i)$ is the observed score, perturbed by the random noise $\eps_i$, independently affecting the rest of observations. Thus, the optimization is formulated as follows
\begin{equation}
 \resizebox{0.88\linewidth}{!}{%
 {$\displaystyle%
    \hat{\bt}_i = \arg\max_{\bt_i} 
    \prod_{j\in\mathcal{W}_i} 
    \frac{1}{\sigma\sqrt{2\pi}} 
    \exp \left(-\frac{(\rand{\bS}(\bx_j)-\bS(\bx_j+\bt_i))^2}{2\sigma^2}\right)~.\label{eq:like}%
$}}
\end{equation}

Its solution, or \emph{maximum-likelihood-estimation} (MLE), is known \emph{a priori}: $\hat{\bt}_i=\bzero$, which is \emph{unbiased} given our statistical model (Eq. \ref{eq:uncertainty_model}), and coherent with Equation \ref{eq:unbiased_location}.

\begin{figure*}[ht]
    \centering
    \begin{tblr}{
            width=1.0\linewidth,
            colspec={*{5}c},
            column{1}={colsep=0pt},
            vspan=even,
        }
        {$\overline{\mathrm{MMA}}$\\overall} &
        \includegraphics[width=3.7cm, valign=c]{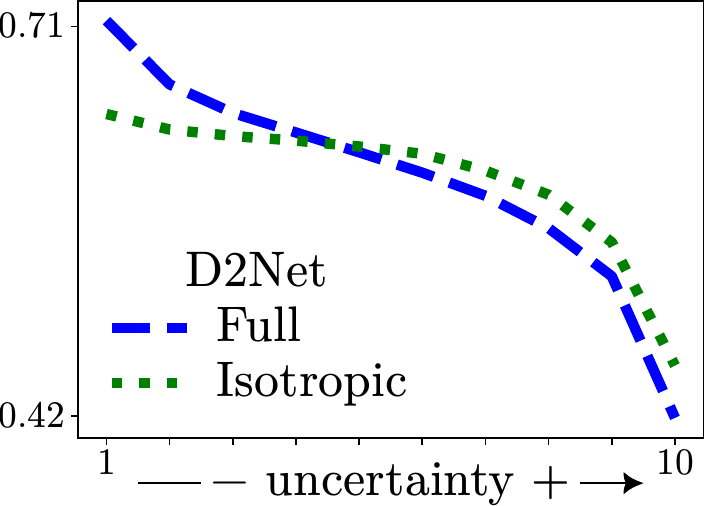} &
        \includegraphics[width=3.7cm, valign=c]{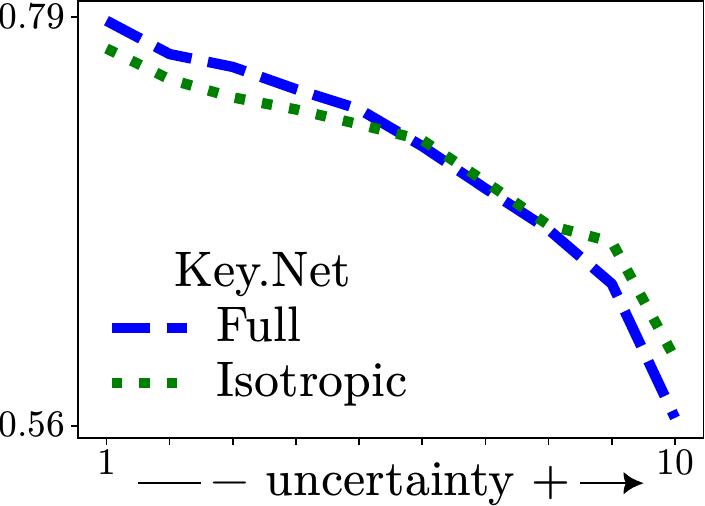} &
        \includegraphics[width=3.7cm, valign=c]{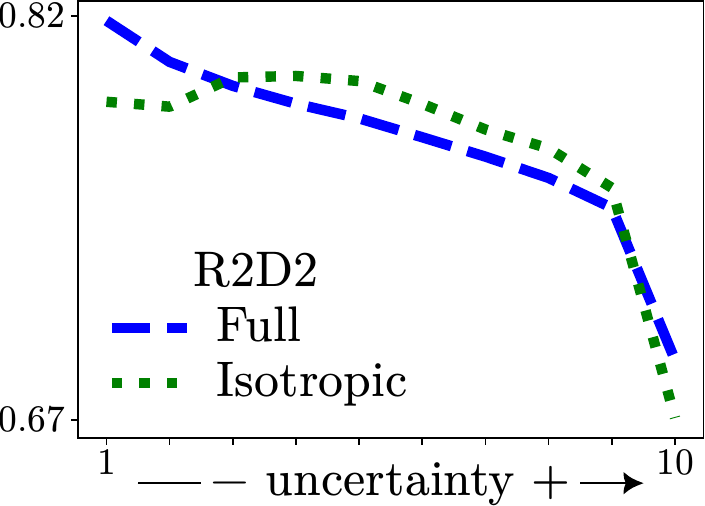} &
        \includegraphics[width=3.7cm, valign=c]{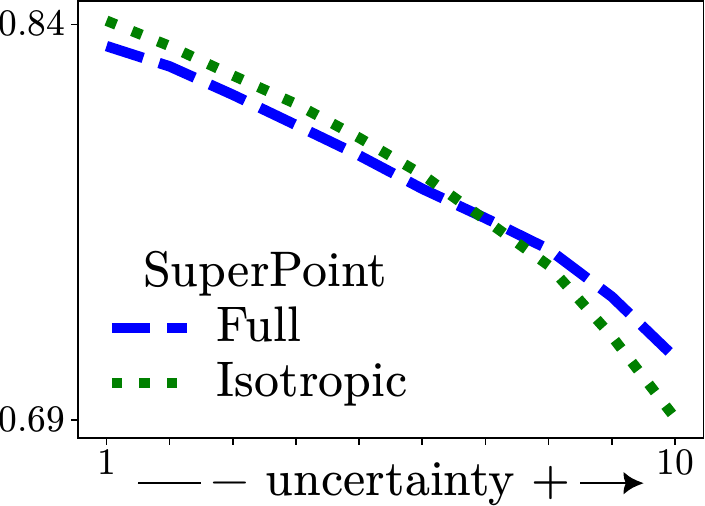}\\
    \end{tblr}
    \caption{\textbf{Uncertainty and matching accuracy on HPatches \cite{balntas2017hpatches}.} We show $\overline{\mathrm{MMA}}$ for each detector. We organize the results into 10 uncertainty ranges, ranging from lowest to highest (x-axis), each containing an equal number of matches. Our covariances accurately model less accurate matches, with the \emph{full}-based approach (structure tensor) showing higher sensitivity in D2Net, Key.Net and R2D2.}
    \label{fig:hpatches}
\end{figure*}

With these conditions, the inverse of the \emph{Fisher information matrix}, $\bcalI(\hat{\bt})$, evaluated at the MLE, imposes a lower bound in the covariance matrix of the estimator $\hat{\bt}_i=\bzero$, known as \emph{Cramer-Rao Lower Bound} (CRLB) \cite{hardle2007applied}:
\begin{equation}\label{eq:crlb}
    \operatorname{Var}(\hat{\bt}_i) \geq \bcalI(\hat{\bt}_i)^{-1}~.
\end{equation}
$\bcalI(\hat{\bt})$ is defined as $\E{\bs_i^\top\bs_i}$, being the variance of the log-likelihood derivative (known as \emph{score}) $\bs_i\ceqq\pt\log\mathcal{L}(\bt_i\mid\mathcal{S})/\pt\bt_i$, since $\E{\bs_i}=\bzero$ \cite{hardle2007applied}. In our case, it is given by
\begin{equation}\label{eq:scores}
\resizebox{0.88\linewidth}{!}{%
${\displaystyle
    \bs_i =
    \sum_{\bx_j\in\mathcal{W}_i} 
    \left(
    \bd_j \ceqq
    \frac{\rand{\bS}(\bx_j) - \bS(\bx_j+\bt_i)}{\sigma^2} 
    \frac{\pt \bS(\bx_j+\bt_i)}{\pt\bt_i}\right)~.%
}$}%
\end{equation}
Due to the linearity of expectation and the independence of observations, $\E{\bd_j^\top\bd_k}=\bzero,~\forall j\neq k$. Thereby
\begin{equation}\label{eq:fim_derivation}
    \bcalI(\hat{\bt}_i) = 
    \sum_{j\in\mathcal{W}_i} 
    \left.
    \E{\bd_j^\top\bd_j}
    \right|_{\hat{\bt}_i}~.
\end{equation}
Since derivatives of Eq. \ref{eq:scores} are applied on our deterministic model, they can go out of the expectation, and evaluating them on $\hat{\bt}_i=\bzero$ lead to
\begin{equation}
\begin{split}
    \bcalI(\hat{\bt}_i) = 
    \sum_{\bx_j\in\mathcal{W}_i}
    \frac{1}{\sigma^4} &
    \left.
    \frac{\pt\bS(\bx_j)}{\pt \bx}
    \right|_{\bx_j}^\top
    \left.
    \frac{\pt\bS(\bx_j)}{\pt \bx}
    \right|_{\bx_j} \\
    & \left.\E{%
    (\rand{\bS}(\bx_j) - \bS(\bx_j+\bt_i))^2 
    }\right|_{\hat{\bt}_i}~.%
\end{split}
\end{equation}
Lastly, $\left.\E{(\rand{\bS}(\bx_j) - \bS(\bx_j+\bt_i))^2}\right|_{\hat{\bt}_i} = \E{\eps_i^2} = \sigma^2$ since $\E{\eps_i} = 0$, implying that our Fisher information matrix is
\begin{equation}
    \bcalI(\hat{\bt}_i) = 
    \frac{1}{\sigma^2}
    \sum_{\bx_j\in\mathcal{W}_i}
    \left.
    \frac{\pt\bS(\bx_j)}{\pt \bx}
    \right|_{\bx_j}^\top
    \left.
    \frac{\pt\bS(\bx_j)}{\pt \bx}
    \right|_{\bx_j}~,
\end{equation}
at the MLE. It matches the local structure tensor (Eq. \ref{eq:tensor}) up to a scale factor $\operatorname{Var}(\eps_i)^{-1}=\sigma^{-2}$, unknown \emph{a priori}. Recalling the CRLB (Eq. \ref{eq:crlb}), although achievable only asymptotically \cite{triggs2000bundle, hardle2007applied}, it motivates $\bSigma_i\ceqq\bC_i^{-1}$ as an up-to-scale covariance matrix of each location $\bx_i$.

\section{Experiments}
\paragraph{Implementation details.}%
In our experiments, we compute the structure tensor (and relative uncertainty) independently of the learned detector: for each local feature $i$, spatial differentiation at $\bS(\bx_j),~\forall j\in\calW_i$, is done with Sobel filters. Integration in $\calW_i$ is done with a $7\times7$ window, the result of using an isotropic Gaussian filter with $\sigma=1$ of cutoff frequency $3\sigma$. Throughout all the experiments, we evaluate and extract our covariances using the detectors of state-of-the-art learned systems: Key.Net~\cite{barroso2022key}, Superpoint~\cite{detone2018superpoint}, D2Net~\cite{dusmanu2019d2} and R2D2~\cite{revaud2019r2d2}. The score map used with Superpoint is the one prior to the channel-wise softmax to avoid the alteration of learned patterns crossing the boundaries of each grid. For the rest of the systems, we directly use their regressed score map. Diverse qualitative examples are shown in Figure~\ref{fig:comparison}.

\begin{figure}[ht!]
    \centering
    \includegraphics[width=1.0\linewidth]{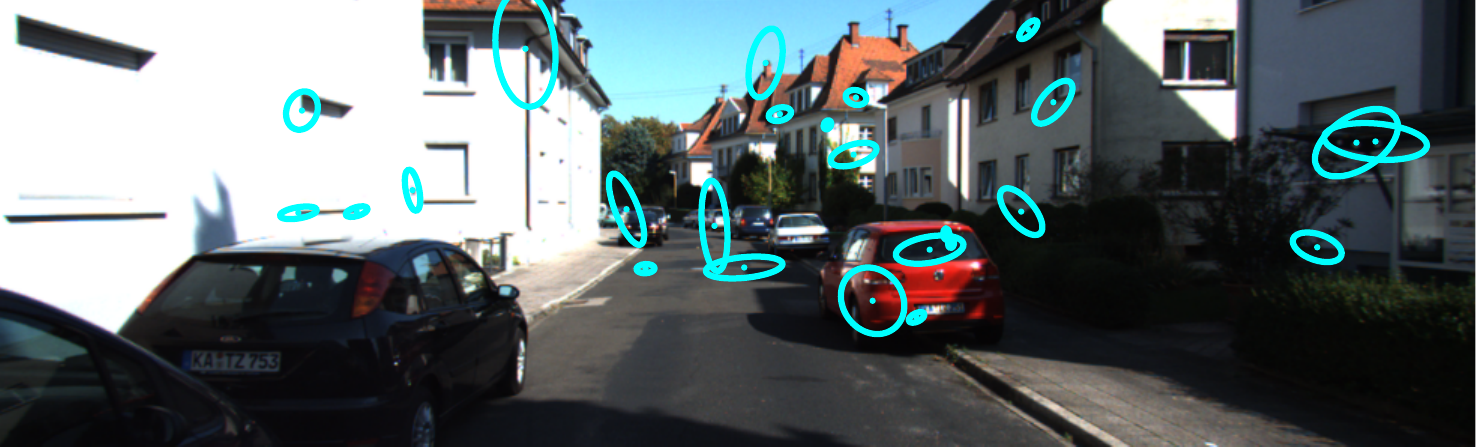}
    \caption{\textbf{Qualitative results on KITTI \cite{geiger2013vision}.} Our covariances correctly model detections with unclear spatial locations, such as faraway corners or those located at edges.}
    \label{fig:qualitative_kitti}
\end{figure}

\begin{figure}[ht!]
    \centering
    \includegraphics[width=1\linewidth]{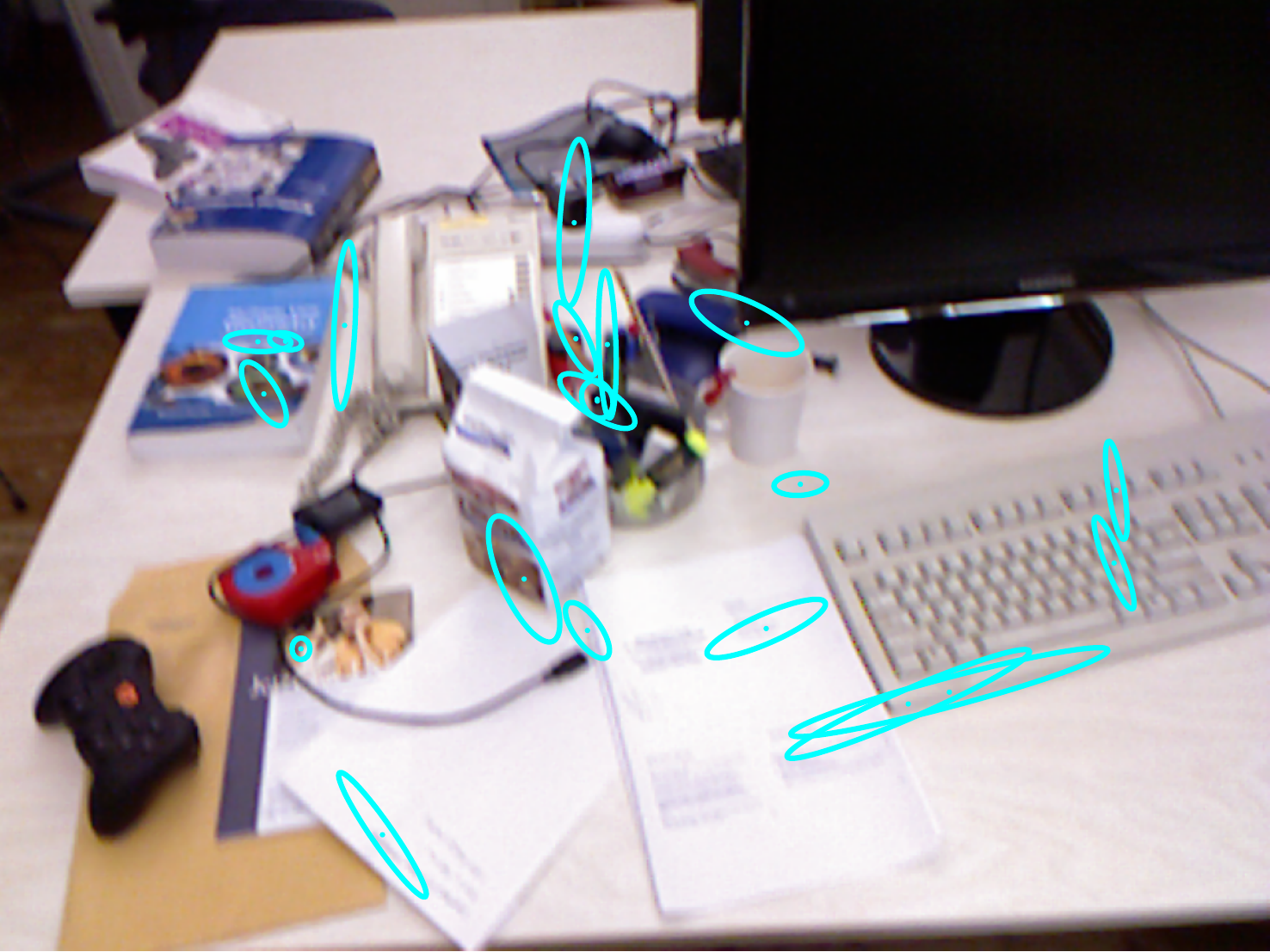}
    \caption{\textbf{Qualitative results on TUM-RGBD \cite{sturm2012benchmark}.} We assign more spatial uncertainty at \emph{a priori} less distinctive regions, such as papers and edges of the keyboard.}
    \label{fig:qualitative_tum}
\end{figure}

\subsection{Matching accuracy.}
We first test the relation between our estimated covariances with the accuracy of local-feature matching. Intuitively, local features detected with higher uncertainty should relate to less accurate matches, and vice versa. For this purpose, we consider the widely adopted HPatches dataset \cite{balntas2017hpatches}. HPatches contains 116 sequences of 6 images each. 57 sequences exhibit significant illumination changes while the other 59 sequences undergo viewpoint changes.

\begin{figure*}[h!]
    \centering
    \begin{tblr}{
            width=1.0\textwidth,
            colspec={*{3}{c}},
            rowsep=0pt,
            cell{1,3}{1}={r=2}{c, cmd={\rotatebox[origin=c]{90}}},
        }
        TUM `freibug\_1' & 
        \includegraphics[width=7.9cm, valign=c]{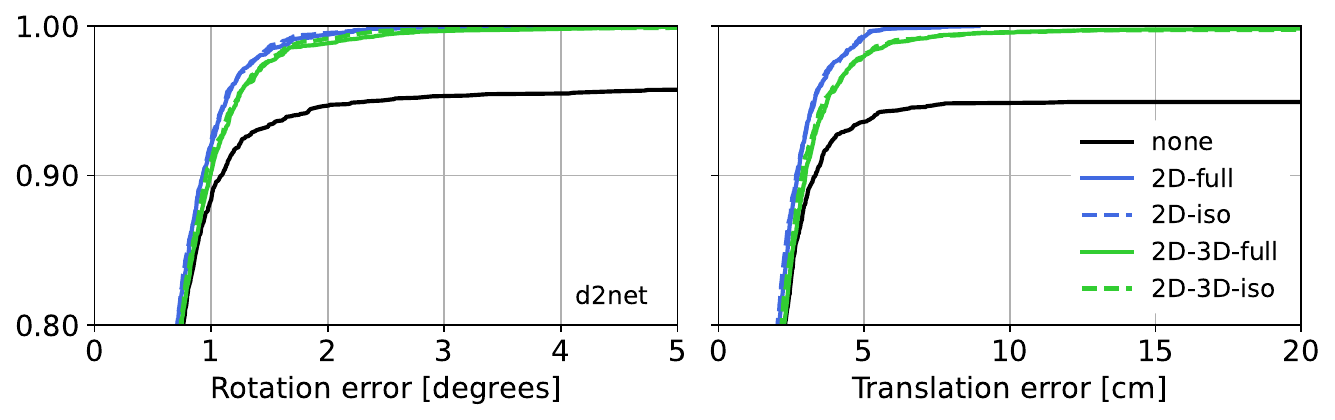} &
        \includegraphics[width=7.9cm, valign=c]{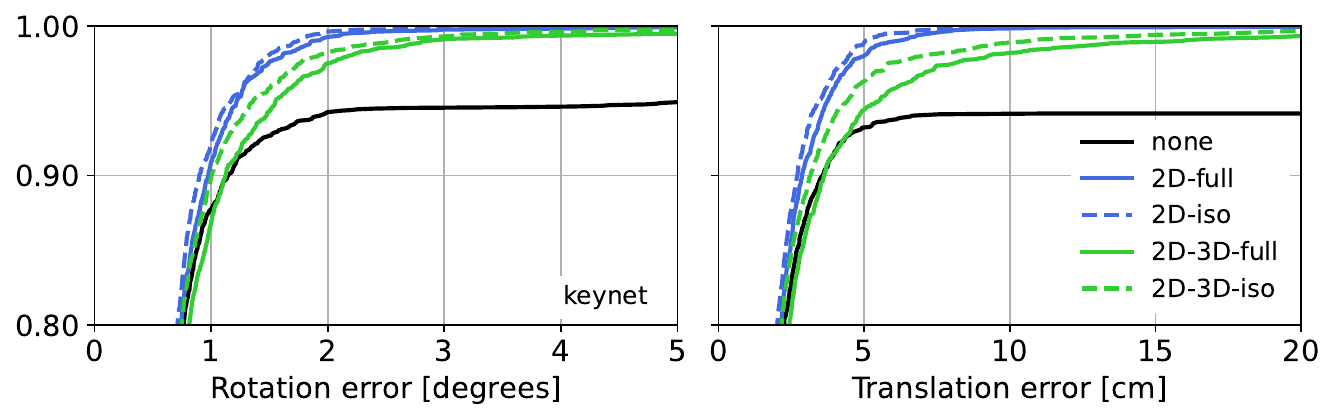} \\
        - &
        \includegraphics[width=7.9cm, valign=c]{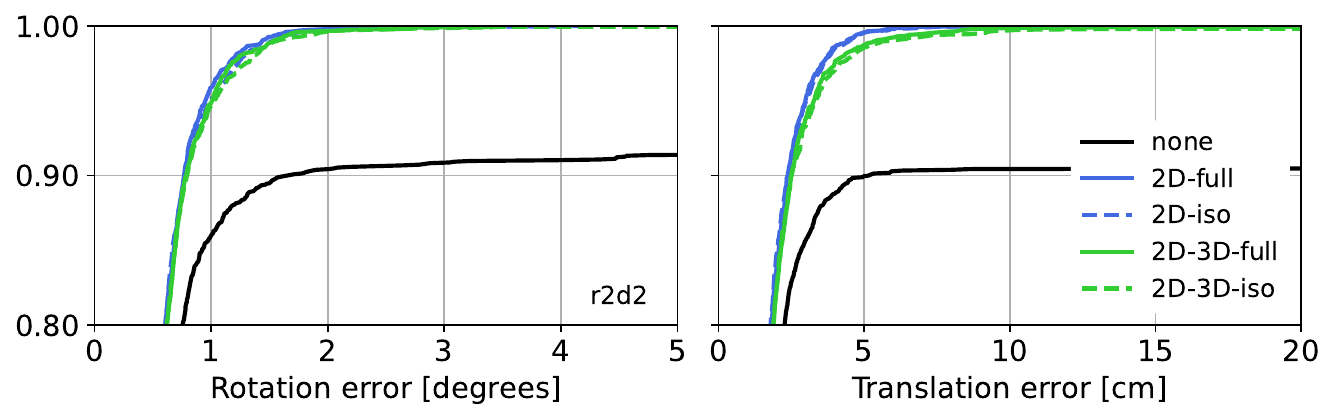} &
        \includegraphics[width=7.9cm, valign=c]{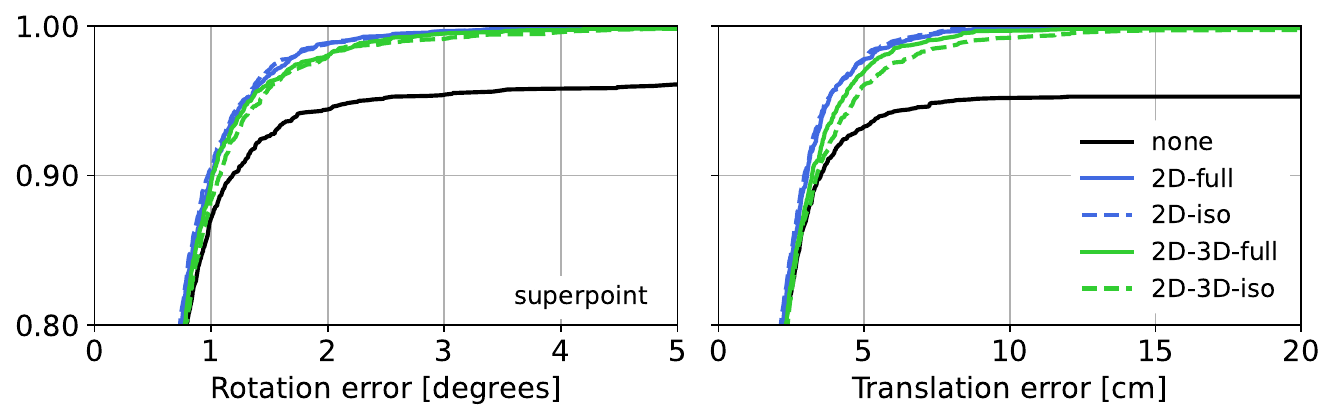} \\
        KITTI 00-02 &
        \includegraphics[width=7.9cm, valign=c]{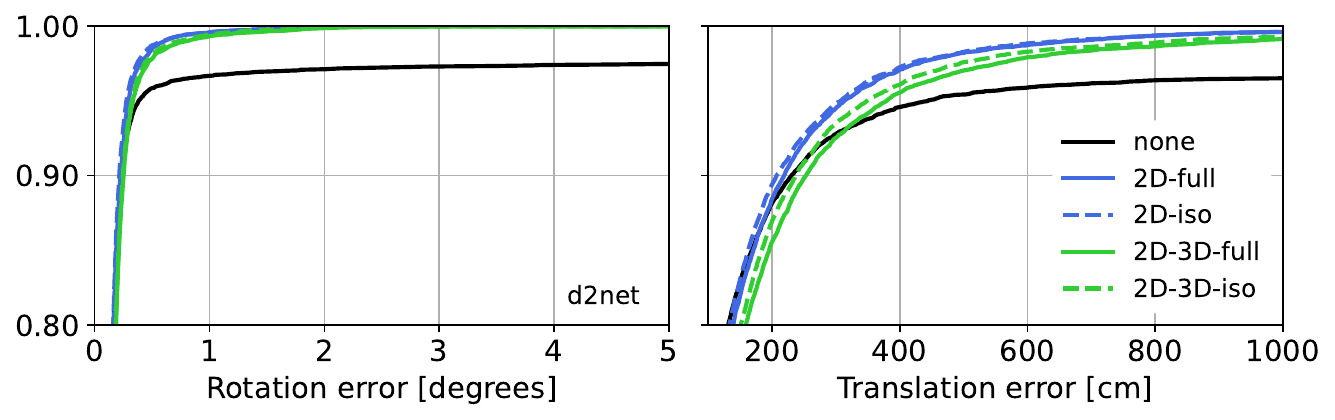} &
        \includegraphics[width=7.9cm, valign=c]{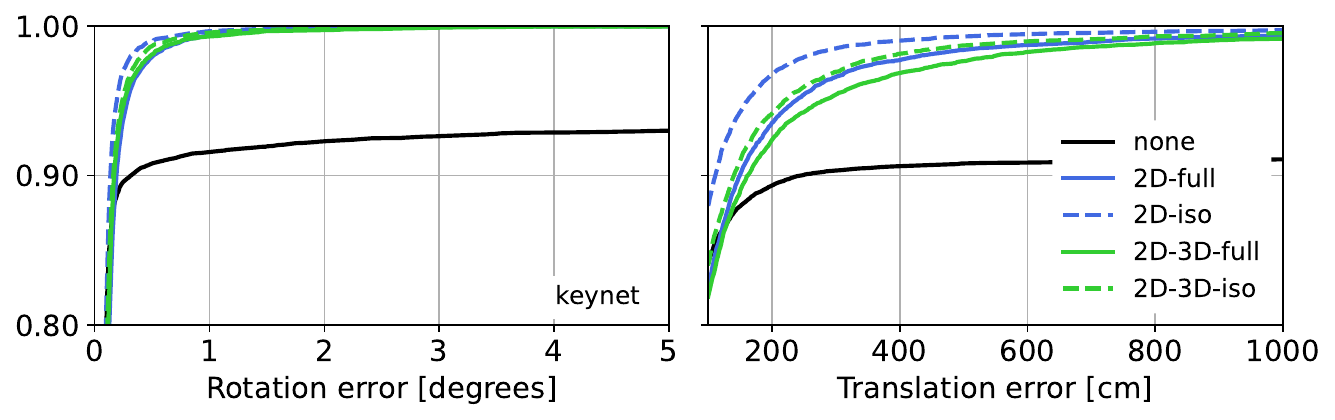} \\
        - & 
        \includegraphics[width=7.9cm, valign=c]{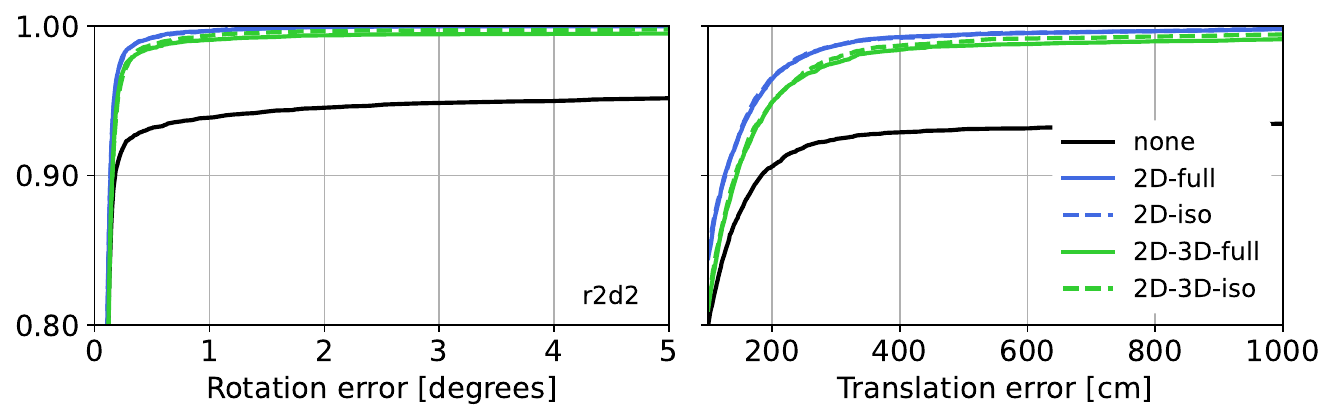} &
        \includegraphics[width=7.9cm, valign=c]{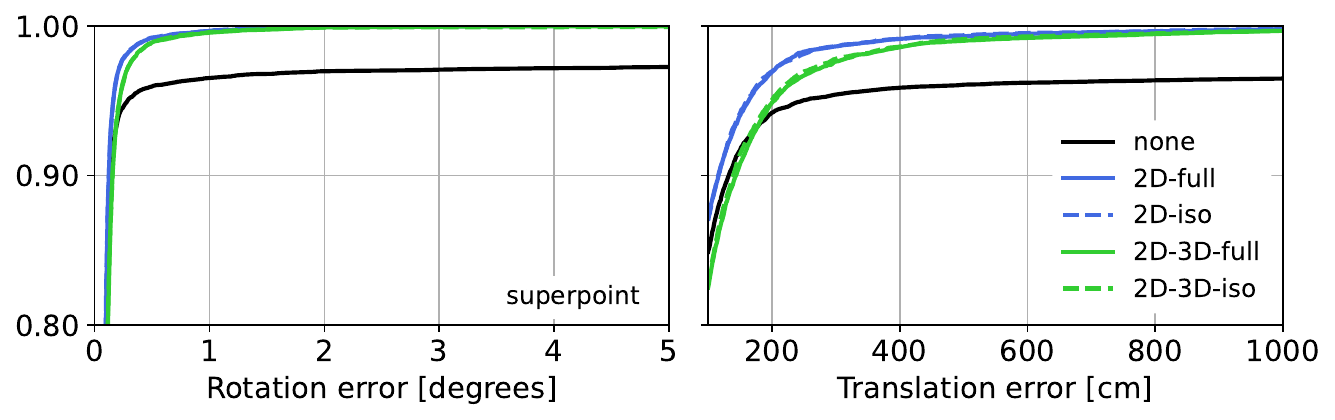}
    \end{tblr}
    \caption{\textbf{Evaluation in TUM-RGBD \cite{sturm2012benchmark} and KITTI \cite{geiger2013vision}.} We report the cumulative errors for camera pose rotation and translation. Practically all estimations converge to acceptable thresholds when using our \emph{2D full} and \emph{2D iso}tropic covariances. This is also apparent when using 3D covariances derived from our 2D ones. Without our covariances, a significant percentage of poses fail to localize.}
    \label{fig:tumkitti_recall}
\end{figure*}

\PAR{Evaluation protocol.}
We base our evaluation on the one proposed by \cite{dusmanu2019d2}. First, extraction of local features and, in our case, covariance matrices of their locations is performed for all images. For every sequence, pairwise matching is done between a reference image $r$ and each remaining image $i$ by using Mutual Nearest Neighbor search (MNN). We then compute the reprojection errors and their covariances with the homographies, $\bH_{i,r}$, provided by the dataset: 
\begin{align}
    \be_{i,r} &\ceqq \mathrm{cart}(\bxhat_i - \bH_{i,r}\bxhat_r)~, \\
    \bSigma_{\be_{i,r}} &\ceqq \bJ \bSigma_{\bx_r} \bJ^\top + \bSigma_{\bx_i}~,
\end{align}
where $\mathrm{cart}(\cdot)$ maps from homogeneous to Cartesian coordinates, and $\bJ \ceqq \pt \be_{i,r}/\pt \bx_r$, \ie we linearly propagate each covariance matrix $\bSigma_{\bx_r}$ of the reference locations.

To quantify the uncertainty of the match with a scalar, we use the biggest eigenvalue of the corresponding $\bSigma_{\be_{i,r}}$. Based on them, all matches gathered in the dataset are distributed in 10 ranges from lowest to highest uncertainty estimates, such that each range has the same number of matches. To quantify the accuracy in matching at each range, we choose the mean matching accuracy error (MMA), which represents the average percentage of matches with a corresponding value of $\lVert \be_{i,r} \rVert$ below a threshold. We use the same thresholds as in \cite{dusmanu2019d2}. Finally, we compute the mean of all the MMA values at each range. This process is repeated for all the evaluated detectors and with the proposed \emph{full} and \emph{isotropic} covariances.

\tumkittiAll[htb]

\paragraph{Results.} Figure \ref{fig:hpatches} shows the averaged mean matching accuracy, $\overline{\mathrm{MMA}}$, at each uncertainty range. Ranges are ordered from lowest (1) to highest (10) estimated uncertainty. As can be seen, it exists a direct relationship between  matching accuracy and both \emph{full} and \emph{isotropic} covariance estimates. With full covariances and for all evaluated ranges, lower uncertainty estimates imply higher accuracy in matching. However, this is not always the case when using isotropic covariances. As can be seen with R2D2, there is a certain increase in $\overline{\mathrm{MMA}}$ for more uncertain matches. Additionally, there is a higher sensitivity of $\overline{\mathrm{MMA}}$ to the uncertainty estimates stemming from full covariances on D2Net, Key.Net, and R2D2. This motivates the need for taking into account the learned local structure when quantifying the spatial uncertainty of the local feature, rather than basing it only on the regressed scalar estimate of the regressed score map.

\subsection{Geometry estimation}\label{sec:exp_geom}
To test the influence of our uncertainties in 3D-geometry estimation, we follow the evaluation proposed in \cite{vakhitov2021uncertainty}. It covers common stages in geometric estimation pipelines such as solving the perspective-n-point problem and motion-only bundle adjustment. The data used consists in the three sequences 00-02 of KITTI \cite{geiger2013vision} and the first three `freiburg\_1' monocular RGB sequences of TUM-RGBD \cite{sturm2012benchmark}. 

\PAR{Evaluation protocol.} KITTI is used with a temporal window of two left frames, while three are used in TUM RGB-D (each with a pose distance $>2.5$ cm). Features and our 2D covariance matrices are extracted with the evaluated detectors. Pairwise matching is done across frames with MNN. In TUM, since more than two images are used, we form feature tracks (set of 2D local features corresponding to the same 3D point) with the track separation algorithm of \cite{dusmanu2020multi}. Matched features are triangulated with ground-truth camera poses and DLT algorithm \cite{Hartley2004}, and refined with 2D-covariance-weighted Levenberg-Marquardt (LM), producing also covariance matrices for 3D point coordinates. The next frame is used for evaluation. After matching it to the reference images we obtain 2D-3D matches which are then processed with P3P LO-RANSAC \cite{chum2003locally} to filter potential outliers. Given the potential inliers, and when using no uncertainty, we choose EPnP~\cite{lepetit2009epnp} as the non-minimal PnP solver. Otherwise, when leveraging our proposed 2D covariances, and optionally, the 3D covariances from LM, we use our implementation (validation in Supp.) of \mbox{EPnPU}~\cite{vakhitov2021uncertainty}. Finally, the estimated camera pose is refined with a covariance-weighted motion-only bundle adjustment. In the Supplemental, we detail how the inclusion of our uncertainty estimates is done in the previous tasks. 

To quantify the accuracy in pose estimation at each sequence, we use the absolute rotation error in degrees: \mbox{$e_{\mathrm{rot}}=\arccos(0.5\operatorname{trace}(\bR^\top_{\mathrm{true}}\bR-1))$}, and the absolute translation error $e_{\mathrm{t}}=\lVert\bt_{\mathrm{true}}-\bt\rVert$ in cm., where $\bR_{\mathrm{true}}, \bt_{\mathrm{true}}$ is the ground-truth pose and $\bR, \bt$ is the estimated one.

\PAR{Results.} Following \cite{vakhitov2021uncertainty}, in Table \ref{tab:tumkitti_all}, we report the mean errors obtained across sequences of both datasets. As can be seen, taking into account the proposed uncertainties is a key aspect to converge pose estimations across sequences. Figures \ref{fig:qualitative_kitti} and \ref{fig:qualitative_tum} show qualitative examples of why uncertainties help geometric estimations by assigning more uncertainty to distant or less reliable keypoints. This behavior can be understood better by having a look at the cumulative error curves. In Figure \ref{fig:tumkitti_recall}, it is shown that practically all pose estimations obtained with methods leveraging the proposed covariances, fall under acceptable error thresholds, whereas the ones from the baseline do not.

\section{Limitations}
The proposed covariances modeling the spatial uncertainty of the learned local-features are up-to-scale. This is not an issue for common 3D geometric estimation algorithms, such as solving linear systems \cite{lepetit2009epnp, vakhitov2021uncertainty} or nonlinear least-squares optimizations \cite{triggs2000bundle}, as their solutions depend only on the relative weight imposed by the covariance matrices. However, this limitation hinders reasoning about the covariances in pixel units. For instance, extracting the absolute scale factor would facilitate the use of robust cost functions, pointing towards a potential direction for future work.

Additionally, while we achieved improvements in 3D geometric estimation tasks on the standard datasets TUM-RGBD \cite{sturm2012benchmark} and KITTI \cite{geiger2013vision}, their exposure to effects like illumination changes or different types of camera motions might be limited. These effects may pose challenges to our approach, as we believe it is subject to the equivariance of the noise of the learned score maps to such changes.
\section{Conclusions}

In this paper, we formulate, for the first time in the literature, detector-agnostic models for the spatial covariances of deep local features. Specifically, we proposed two methods based on their learned score maps: one using local-feature scores directly, and another theoretically-motivated method using local structure tensors. Our experiments on KITTI and TUM show that our covariances are well calibrated, significantly benefiting 3D-geometry estimation tasks.
\iftoggle{cvprfinal}{
{\footnotesize \PAR{Acknowledgements.} The authors thank Alejandro Fontán Villacampa for his thoughtful comments and help with the experiments. This work was supported by the Ministerio de Universidades Scholarship FPU21/04468. \par}
}{}

{
    \small
    \bibliographystyle{ieeenat_fullname}
    \bibliography{main}
}

\clearpage
\setcounter{page}{1}
\maketitlesupplementary

\appendix

\section{Accounting for uncertainty}

\subsection{Perspective-n-Point problem}\label{subsec:pnp}
\PAR{PnP.} Consider a set of $n$ 3D points $\{\bp_{w,i}\in\bb{R}^3\mid i\in 1\twodots n\}$ expressed in an absolute reference system $\{w\}$, along with their corresponding 2D points $\{\bx_{i}\in\bb{R}^2\mid i\in 1\twodots n\}$ in an image captured by a camera with calibration matrix $\bK\in\bb{R}^{3\times3}$. The Perspective-n-Point problem (PnP) involves finding the rotation $\bR_{cw}\in SO(3)$ and translation $\bt_{cw}\in\bb{R}^3$ that transform the 3D points to the camera's reference system $\{c\}$: $\bp_{c,i}\ceqq\bR_{cw}\bp_{w,i}+\bt_{cw}$.

\PAR{EPnP(U).} To leverage our proposed covariance matrices, we adopt the recent EPnPU \cite{vakhitov2021uncertainty} as our PnP solver. EPnPU extends EPnP \cite{lepetit2009epnp} to take into account the uncertainty of the observations. To understand how uncertainties are included, consider the algebraic residual $\br_i$ ($i\in 1\twodots n$):
\begin{equation}\label{eq:epnp_res0}
    \br_i~\ceqq~\homobx_i\sel{1,2} - \homobx_i\sel{3}\bx_i~,
\end{equation}
where $\homobx_i$ represents,  in homogeneous coordinates, the estimated 2D location corresponding to $\bp_{w,i}$:
\begin{equation}\label{eq:res_homog}
    \homobx_i \ceqq \bK(\bR_{c,w}\bp_{w,i} + \bt_{c,w})~,
\end{equation}
where $\homobx_i^{(1,2)}$ (resp. $\homobx_i^{(3)}$) represents its two first entries (resp. third entry). Given this, we seek to estimate the covariance matrix of each residual, $\bSigma_{\br_i}\ceqq\operatorname{Cov}(\br_i)$, that stems from the covariance matrices of the observations $\bSigma_{\bx_i}$ and $\bSigma_{\bp_{w,i}}$. 

Each residual can be linearized by instead considering a set of control points as the unknowns\footnote{To ease uncertainty propagation from $\bp_{w,i}$ to $\homobx_i$, we instead parameterize  $\homobx_i$ with the pose $(\bR_{cw}, \bt_{cw})$, as in \cref{eq:res_homog}. Because of this, we need an initial rough pose estimate. This pose estimate is obtained at no additional cost during the outlier filtering step with P3P LO-RANSAC.} of the problem, whose concatenation we denote here as $\bc\in\bb{R}^{12}$, and a matrix block $\bM_i\in\bb{R}^{2\times12}$ that depends only on the input data \cite{lepetit2009epnp}. Thus, the solution to $\bc$ is found in the null space of the matrix formed by the concatenation of each residual \cite[Eq. 7]{lepetit2009epnp}:
\begin{equation}
    \bM\bc = \bzero~,
    \quad\mathrm{with}\quad
    \bM \ceqq 
    \begin{bmatrix}
        \bM_{1}^\top & \cdots & \bM_{n}^\top
    \end{bmatrix}^\top~.
\end{equation}
Thanks to this, we can weigh the influence of the residuals according to their projections onto the directions of extreme uncertainty:
\begin{align}
    &\bM^\top\bSigma_{\br}^{-1}\bM\bc = \bzero~,\\
    \mathrm{with}\quad
    &\bSigma_{\br} \ceqq \operatorname{diag}(\bSigma_{\br_1}, \ldots, \bSigma_{\br_n})~.
\end{align}

\PAR{$\bSigma_{\br_i}$ derivation.} For the noise $\bnu\in\bb{R}^3$ affecting the coordinates of each 3D point, we assume a similar model to the one proposed for the 2D location $\bx_i$ of each local feature (\cref{eq:base_model,eq:unbiased_location}):
\begin{align}
    &\bp_{w,i} = \E{\bp_{w,i}} + \bnu~,\label{eq:rand_pw_i}\\
    &\E{\bnu} =\bzero~,\quad
    \bSigma_{\bp_{w,i}}\ceqq\E{\bnu\bnu^\top}~.
\end{align}
where, as we do in our case, the covariance matrix $\bSigma_{\bp_{w,i}}\in\bb{R}^{3\times3}$ can be estimated after the convergence of the nonlinear optimization of $\bp_{w,i}$ (\cref{sec:supp_nonlinear}). 

Plugging \cref{eq:rand_pw_i} in \cref{eq:res_homog} leads to
\begin{equation}\label{eq:vak_derhx}
    \homobx_i 
    = 
    \bK\bR\bp_{w,i} + \bK\bt 
    =
    \E{\homobx_i} + \bzeta~,
\end{equation}
where $\E{\bzeta}=\bzero$, because $\bzeta\ceqq\bK\bR\bnu$, and $\E{\bnu}=\bzero$ (by definition in \cref{eq:rand_pw_i}). According to \cref{eq:vak_derhx}, we can propagate the uncertainty of $\bp_{w,i}$ to $\homobx_i$ by:
\begin{equation}\label{eq:vak_covx}
    \bSigma_{\homobx_i} 
    \ceqq 
    \begin{bmatrix}
        \bLambda & \bw \\
        \bw^\top & \gamma
    \end{bmatrix} \ceqq 
    \bK\bR\bSigma_{\bp_{w,i}}\bR^\top\bK^\top~.
\end{equation}

Each residual can then be expressed as a random variable as follows:
\begin{align}\label{eq:vak_derr}
    \br_{i} 
    &= \homobx_i\sel{1,2} - \homobx_i\sel{3}\bx_i~,\\
    \begin{split}
        &= (\E{\homobx_i\sel{1,2}} + \bzeta\sel{1,2}) \\
        &\qquad - (\E{\homobx_i\sel{3}}+\bzeta\sel{3})(\E{\bx_i} + \bxi)~,
    \end{split} \\
    \begin{split}
        &= \E{\homobx_i\sel{1,2} - \homobx_i\sel{3}\bx_i}
        + \bzeta\sel{1, 2} 
        - \E{\homobx_i\sel{3}}\bxi \\
        &\qquad - \bzeta\sel{3}\E{\bx_i}
        - \bzeta\sel{3}\bxi~,
    \end{split}
\end{align}
where we have used the linearity of the expectation and assumed independence between $\bxi$ and $\bzeta$. Following \cref{eq:vak_derr} and the fact that $\E{\br_{i}} = \E{\homobx_i\sel{1,2} - \homobx_i\sel{3}\bx_i}~$ since $\E{\bxi}=\bzero_2,~\E{\bzeta}=\bzero_3~$, the derivation of $\bSigma_{\br_i}~$ follows as in \cref{tab:sigma_ri_der}.

\begin{table*}
\centering
\begin{minipage}{0.75\textwidth}
\begin{align}
    \bSigma_{\br_i} 
    \ceqq & \E{(\br_i - \E{\br_i})(\br_i - \E{\br_i})^\top}~, \\[1em]
    \begin{split}
        = & \E{
        (\bzeta\sel{1, 2} - \E{\homobx_i\sel{3}}\bxi - \bzeta\sel{3}\E{\bx_i} - \bzeta\sel{3}\bxi)\\%
        &~~~~~~~~~(\bzeta\sel{1, 2} - \E{\homobx_i\sel{3}}\bxi - \bzeta\sel{3}\E{\bx_i} - \bzeta\sel{3}\bxi)^\top}~,
    \end{split} \\[1em]
    \begin{split}
        = &%
        \underbrace{\E{\bzeta\sel{1,2}(\bzeta\sel{1,2})^\top}}_{\bLambda} -
        \underbrace{\cancel{\E{\E{\homobx_i\sel{3}}\bzeta\sel{1,2}\bxi^\top}}}_{\E{\bzeta\sel{1,2}}=\E{\bxi}=\bzero} -
        \underbrace{\E{\bzeta\sel{3}\bzeta\sel{1,2}\E{\bx_i}^\top}}_{\bw\E{\bx_i}^\top} - \\[1em] 
        &%
        \underbrace{\cancel{\E{\bzeta\sel{3}\bzeta\sel{1,2}\bxi^\top}}}_{\E{\bxi}=\bzero} -
        \underbrace{\cancel{\E{\E{\homobx_i\sel{3}}\bxi(\bzeta\sel{1,2})^\top}}}_{\E{\bxi}=\E{\bzeta\sel{1,2}}=\bzero} +
        \underbrace{\E{\E{\homobx_i\sel{3}}^2\bxi\bxi^\top}}_{\E{\homobx_i\sel{3}}^2\bSigma_\bx} +\\[1em]
        &%
        \underbrace{\cancel{\E{\E{\homobx_i\sel{3}}\bzeta\sel{3}\bxi\E{\bx_i}^\top}}}_{\E{\bzeta\sel{3}}=0,~\E{\bxi}=\bzero} +
        \underbrace{\cancel{\E{\E{\homobx_i\sel{3}}\bzeta\sel{3}\bxi\bxi^\top}}}_{\E{\bzeta\sel{3}}=0} -
        \underbrace{\E{\bzeta\sel{3}\E{\bx_i}(\bzeta\sel{1,2})^\top}}_{\E{\bx_i}\bw^\top} +\\[1em]
        &%
        \underbrace{\cancel{\E{\bzeta\sel{3}\E{\homobx_i\sel{3}}\E{\bx_i}\bxi^\top}}}_{\E{\bzeta\sel{3}}=0,~\E{\bxi}=\bzero} +
        \underbrace{\E{(\bzeta\sel{3})^2\E{\bx_i}\E{\bx_i}^\top}}_{\gamma\E{\bx_i}\E{\bx_i}^\top} +
        \underbrace{\cancel{\E{(\bzeta\sel{3})^2\E{\bx_i}\bxi^\top}}}_{\E{\bxi}=\bzero} -\\[1em]
        &%
        \underbrace{\cancel{\E{\bzeta\sel{3}\bxi(\bzeta\sel{1,2})^\top}}}_{\E{\bxi}=\bzero} +
        \underbrace{\cancel{\E{\bzeta\sel{3}\E{\homobx_i\sel{3}}\bxi\bxi^\top}}}_{\E{\bzeta\sel{3}}=0} +
        \underbrace{\cancel{\E{(\bzeta\sel{3})^2\bxi\E{\bx_i}^\top}}}_{\E{\bxi}=\bzero} +\\[1em]
        &%
        \underbrace{\E{(\bzeta\sel{3})^2\bxi\bxi^\top}}_{\E{(\bzeta\sel{3})^2}\E{\bxi\bxi^\top}=\gamma\bSigma_{\bx_i}}~,
    \end{split} \\[0.2em]
    =&~
    \bLambda - 
    \bw\E{\bx_i}^\top 
    + \E{\homobx_i\sel{3}}^2\bSigma_{\bx_i} 
    - \E{\bx_i}\bw^\top 
    + \gamma\E{\bx_i}\E{\bx_i}^\top + 
    \gamma\bSigma_{\bx_i}~.%
\end{align}
\end{minipage}
\caption{Derivation of $\bSigma_{\br_i}~$ (\cref{subsec:pnp}).}
\label{tab:sigma_ri_der}
\end{table*}

\subsection{Nonlinear optimizations}\label{sec:supp_nonlinear}
\PAR{Uncertainty-based geometric refinement.} Commonly, initial estimates of camera poses and 3D points (obtained, in our case, using EPnP(U) and multi-view DLT triangulation \cite{Hartley2004}, respectively) are not geometrically optimal\footnote{DLT and EPnP(U) minimize algebraic errors. Furthermore, EPnP(U) approximate the solution, as they do not guarantee equality between the control point distances when expressed in $\{c\}$ and $\{w\}$ \cite{lepetit2009epnp}.}. Therefore, they are typically optimized using iterative algorithms such as \emph{Gauss-Newton} or \emph{Levenberg-Marquardt} \cite{triggs2000bundle} by minimizing \emph{reprojection errors}. This approach is considered in the literature as the \emph{gold standard} \cite{Hartley2004}.

Consider a 3D point $j$ expressed in $\{w\}$: $\bp_{w, j}\in\bb{R}^3$, that is observed from a camera $i$ whose pose is: $\bR_{cw,i}\in SO(3),\bt_{cw,i}\in\bb{R}^3$, then we define the reprojection error as
\begin{equation}\label{eq:rep_err}
    \be_{ij} \ceqq \bx_{ij} - \operatorname{\pi}(\bR_{cw,i},\bt_{cw,i},\bp_{w, j})~,
\end{equation}
where $\bx_{ij}\in\bb{R}^2$ is the 2D location in the image of a local feature corresponding to $\bp_{w, j}$ and $\pi(\cdot)$ is the projection function. Thereby, the variables of interest (poses and/or 3D points), which we represent for convenience in vectorized form as $\by$, are refined with the following optimization
\begin{align}
    \by^*
    &\ceqq \arg
    \min_\by \frac{1}{2} \sum_{i,j}\be_{ij}(\by)^\top,\bW_{ij},\be_{ij}(\by)~,\label{eq:min_rep_err}\\
    &= \arg \min_\by\frac{1}{2} \be(\by)^\top\bW\be(\by)~,\\
    \text{with}~~
    \be(\by) &\ceqq 
    \begin{bmatrix}
        \ldots & \be_{ij}(\by)^\top & \ldots
    \end{bmatrix}^\top~,\\
    \bW &\ceqq \operatorname{diag}(\ldots,~\bW_{ij},~\ldots)~,
\end{align}
where $\bW_{ij}\ceqq\bSigma_{ij}^{-1}$ represents the inverse of the covariance matrix of $\be_{ij}$ (we use the identity matrix $\bW_{ij}\ceqq\bI_2$ in the no-uncertainty baseline), thus weighing each $\be_{ij}$ according to its projection onto the directions of extreme uncertainty.

In our experiments, we approximate the solution of \cref{eq:min_rep_err} with ten iterations of Levenberg-Marquardt (LM) \cite{triggs2000bundle} \ie we iteratively update $\by\leftarrow\by\oplus\Delta\by$ within the manifold of the unknowns $\by$ \cite{blancoclaraco2022tutorial}, by composing them with increments $\Delta\by$ computed by solving the following linear system of equations
\begin{equation}
    (\bJ^\top\bW\bJ + \lambda\operatorname{diag}(\bJ^\top\bW\bJ))\dby = -\bJ^\top\bW\be(\by)~,
\end{equation}
where $\bJ\ceqq\evalat{\frac{\pt\be}{\pt\dby}}$ is the Jacobian matrix of the residuals. As such, LM imposes a penalization of the magnitude of $\Delta\by$ that is controlled with the damping factor $\lambda\in\bb{R}$. We initialize $\lambda$ as $10^{-3}$ times the average of $\operatorname{diag}(\bJ^\top\bW\bJ)$, as recommended in \cite{Hartley2004}. We also follow the recommended protocol of \cite[App. A6.2]{Hartley2004} for updating $\lambda$ at each iteration.

\begin{figure*}[!htb]
    \centering
    \begin{tblr}{
            colspec={*{2}{c}},
            cell{1,2}{1}={c, cmd={\rotatebox[origin=c]{90}}}
        }
        2D noise &
        \includegraphics[width=0.93\textwidth, valign=c]{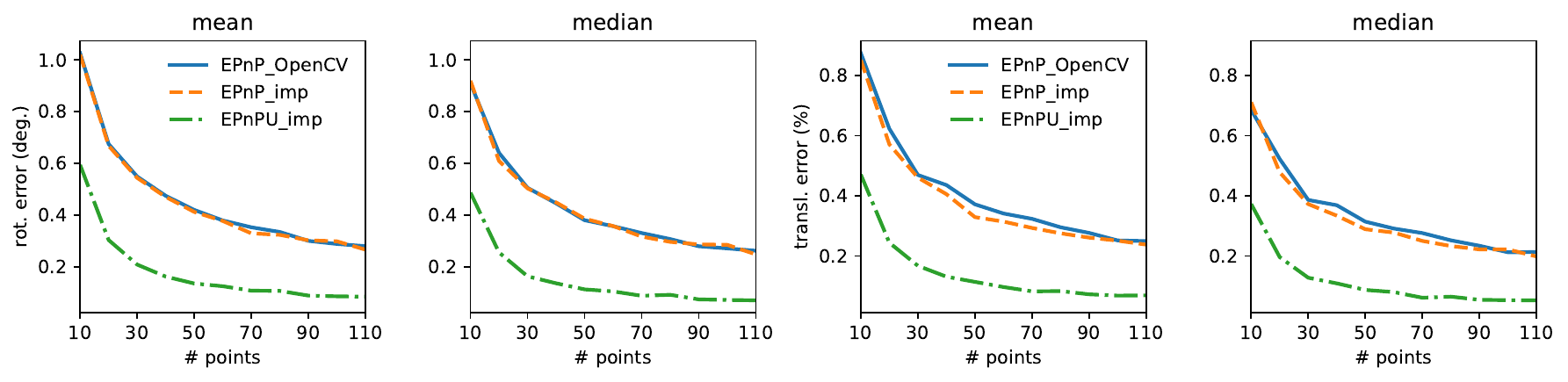} \\
        2D and 3D noise & 
        \includegraphics[width=0.93\textwidth, valign=c]{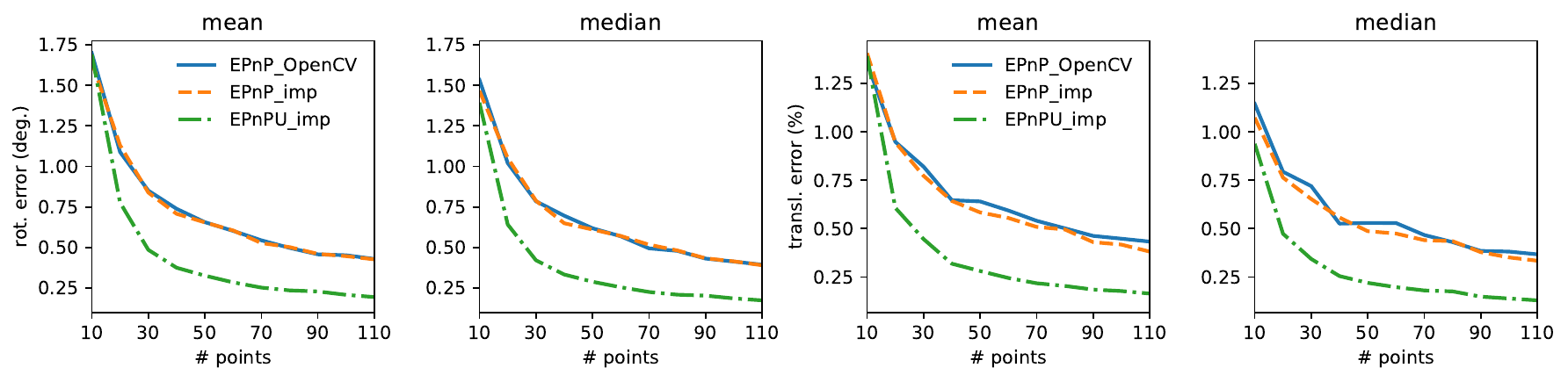}
    \end{tblr}
    \caption{\textbf{Validation of EPnPU \cite{vakhitov2021uncertainty} implementation with known synthetic noise.} Our implementation, labeled as \mbox{`EPnPU\_imp'} obtains pose errors akin to the ones of \cite{vakhitov2021uncertainty}, improving as it should over EPnP \cite{lepetit2009epnp}, labeled as `EPnP\_OpenCV', whose behavior is matched when using our implementation with identity covariance matrices, labeled as `EPnP\_imp'.}
    \label{fig:epnpu_val}
\end{figure*}

\PAR{Covariances for 3D-points refinement.} Since we use ground-truth camera poses for triangulation in our experiments, we directly use our proposed 2D covariances $\bSigma_{ij}\ceqq\bSigma_{\bx_{ij}}$ as the covariance matrix of each error $\be_{ij}$. After convergence, we estimate the covariance matrix $\bSigma_{\bp_{w,j}}$ of the 3D point as the inverse of the Hessian, following \cite{vakhitov2021uncertainty}.

\PAR{Covariances for motion-only bundle-adjustment.} 
The covariance matrix of each error $\be_{j}$\footnote{We drop here the subscript $i$ to avoid clutter since just one camera is considered in motion-only BA.}, when considering 3D noise, is estimated \emph{at each iteration} following \cite{vakhitov2021uncertainty}:
\begin{equation}
    \bSigma_{j} \ceqq 
    \bSigma_{\bx_j} + 
    \frac{\pt\pi(\bp_{c,j})}{\pt\bp_{c,j}}\,\,
    \bR_{cw}\bSigma_{\bp_{w,j}}\bR_{cw}^\top
    \left(\frac{\pt\pi(\bp_{c,j})}{\pt\bp_{c,j}}\right)^\top
\end{equation}
\ie linearly propagating $\bSigma_{\bp_{w,j}}$ and assuming independence between the distributions of $\bx_j$ and $\bp_{w,j}$. On the other hand, if only 2D noise is considered, we consider $\bSigma_{j} \ceqq \bSigma_{\bx_j}$.

\section{Validation of EPnPU implementation}

Since all learned detectors used in our paper \cite{detone2018superpoint, dusmanu2019d2, revaud2019r2d2, barroso2022key} are implemented in Python, but \mbox{EPnPU}~\cite{vakhitov2021uncertainty} is originally written in
MATLAB\footnote{\url{https://github.com/alexandervakhitov/uncertain-pnp}}, we reimplemented this last one in Python to ease the workflow. As validation, we followed the same synthetic experiments done in \cite{vakhitov2021uncertainty}, where the noise affecting the simulated observations (2D and 3D points) is known beforehand. Comparisons are done with OpenCV's implementation of EPnP \cite{lepetit2009epnp}. Since EPnPU is an extension of EPnP's algorithm to include uncertainties, results should improve accordingly when leveraging them. Additionally, we compare our implementation when using just identity covariance matrices, to verify that it matches the behavior of EPnP.

As shown in Figure~\ref{fig:epnpu_val}, our EPnPU implementation gives results very similar to the ones reported in \cite{vakhitov2021uncertainty}. In turn it improves the results of OpenCV's EPnP, whose behavior also matches our EPnPU implementation when just using identity covariances matrices.

\section{Interpretability}

\begin{figure}
    \centering
    \includegraphics[width=1.0\linewidth]{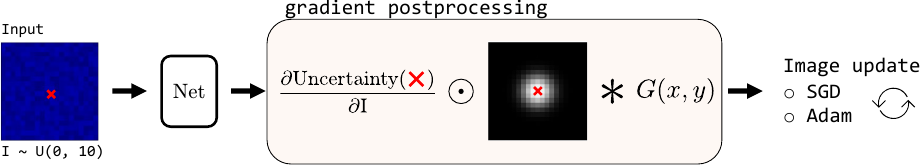}
    \caption{\textbf{Interpretability}. We update a \mbox{$20\times20$} input patch $\mathtt{I}$ by minimizing our uncertainty estimates (\emph{full} and \emph{isotropic}) at the center of the patch. Gradients are downweighted and smoothed, as in \cite{mordvintsev2015inceptionism}, to favor convergence.}
    \label{fig:interp_pipeline}
\end{figure}

\begin{figure}[!htb]
    \centering
    \includegraphics[width=1.0\linewidth]{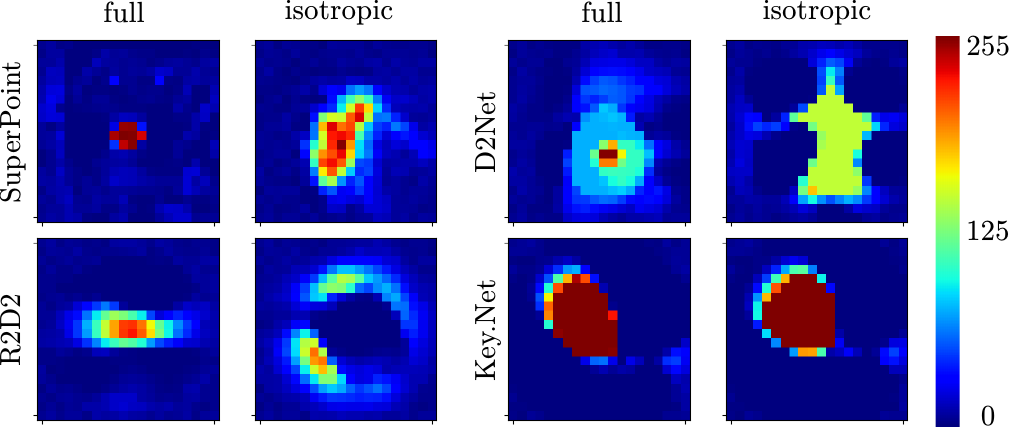}
    \caption{\textbf{Interpretability results}. Minimizing both uncertainty measures lead to distinctive (blob-like and corner-like) regions generated in the input patch. Differences obtained depending on learned detector highlights the detector-agnostic behavior of the proposals.}
    \label{fig:interp_results}
\end{figure}

Our proposed methods for quantifying the uncertainty of the locations are based on the learned score maps, independently of the detector that has learned it. Depending on its training, systems learn to focus on different input image patterns. For instance, SuperPoint~\cite{detone2018superpoint} is trained to detect corners, while Key.Net~\cite{barroso2022key}, D2Net~\cite{dusmanu2019d2} and R2D2~\cite{revaud2019r2d2} do not directly impose such constraint. Additionally, all of them use different learning objectives. 

To explore what kind of locations get assigned low uncertainty estimates, we set up a toy experiment inspired by DeepDream~\cite{mordvintsev2015inceptionism}. As depicted in Fig. \ref{fig:interp_pipeline} we update a \mbox{$20\times20$} synthetic input patch via gradient descent, such that we minimize the biggest eigenvalue of the covariance matrix (computed by using the regressed score map) at the center pixel. We downweight the gradients located at the extremes of the receptive field and, as in \cite{mordvintsev2015inceptionism}, we smooth the gradients with a Gaussian filter.

Results after convergence are shown in Fig.~\ref{fig:interp_results}. We obtain distinctive blob/corner-like regions by minimizing both uncertainty estimates. This highlights the detector-agnostic behavior of our two methods. Interestingly, excepting Key.Net, the generated patterns are slightly different depending on the method. We attribute this to the the fact that the \emph{full} approach takes into account the surrounding learned patterns, which in this case increases the saliency of the generated input image pattern.

\end{document}